\documentclass[10pt,journal,compsoc]{IEEEtran}
\ifCLASSOPTIONcompsoc
  \usepackage[nocompress]{cite}
\else
  \usepackage{cite}
\fi
\ifCLASSINFOpdf
\else
\fi
\usepackage{amsfonts}
\usepackage{array}
\usepackage[caption=false,font=normalsize,labelfont=sf,textfont=sf]{subfig}
\usepackage{textcomp}
\usepackage{stfloats}
\usepackage{url}
\usepackage{verbatim}
\usepackage{graphicx}
\usepackage{cite}
\usepackage{times}
\usepackage{colortbl}
\usepackage{epsfig}
\usepackage{graphicx}
\usepackage{amsmath}
\usepackage{amssymb}
\usepackage{makecell} 

\usepackage{enumitem} 
\usepackage[linesnumbered,ruled,vlined,algo2e]{algorithm2e}
\usepackage{algorithm, algorithmic}
\usepackage{multirow}
\usepackage{threeparttable}
\usepackage{stfloats}
\usepackage{booktabs}
\usepackage{arydshln}
\usepackage{bm}
\usepackage{makecell}
\usepackage{pifont}
\usepackage{lipsum}
\usepackage{ulem}
\usepackage[table,dvipsnames]{xcolor} 
\usepackage[pagebackref,breaklinks,colorlinks]{hyperref}

\usepackage[capitalize]{cleveref}
\crefname{section}{Sec.}{Secs.}
\Crefname{section}{Section}{Sections}
\Crefname{table}{Table}{Tables}
\crefname{table}{Tab.}{Tabs.}

\colorlet{colorFst}{Green!25}       
\colorlet{colorSnd}{Green!10} 
\colorlet{colorTrd}{Yellow!30}      
\colorlet{colorLow}{darkgray!30}    
\definecolor{darkpastelgreen}{rgb}{0.01, 0.75, 0.24}
\definecolor{darkgreen}{rgb}{0.00, 0.8, 0.2}
\definecolor{darkyellow}{rgb}{0.96, 0.75, 0.00}
\definecolor{badcolor}{rgb}{0.82,0.25,0.12}
\newcommand{\fs}{\cellcolor{colorFst}\bf}   
\newcommand{\nd}{\cellcolor{colorSnd}}      
\newcommand{\rd}{\cellcolor{colorTrd}}      

\newcommand{\impressive}{\color{black}}
\newcommand{\good}{\color{black}}
\newcommand{\goodd}{\color{darkpastelgreen}}
\newcommand{\zero}{\color{black}}
\newcommand{\bad}{\color{badcolor}}
\newcommand{\broma}{\color{RawSienna}}

\newcommand{\ours}{\cellcolor[rgb]{.95,.95,.95}} 
\newcommand{\dashlineours}[2]{ \noalign{\vskip 0.25ex} \cdashline{#1-#2} \noalign{\vskip 0.5ex}}

\def\eg{\emph{e.g.}\xspace} 
\def\ie{\emph{i.e.}\xspace}

\newcommand{\boldparagraph}[1]{\vspace{0.1em}{\bf #1}}

\let\titleold\title
\renewcommand{\title}[1]{\titleold{#1}\newcommand{\thetitle}{#1}}

\begin{document}
%
\title{MESA: Effective Matching Redundancy Reduction by Semantic Area Segmentation}
%
%
%

\author{Yesheng Zhang, Shuhan Shen,~\IEEEmembership{Senior Member,~IEEE}, Xu Zhao,~\IEEEmembership{Member,~IEEE}
\thanks{Yesheng Zhang and Xu Zhao are with Department of Automation, Shanghai Jiao Tong University. (e-mail: {\tt\small \{preacher, zhaoxu\}@sjtu.edu.cn})
	\\Shuhan Shen is with the Institute of Automation, Chinese Academy of Sciences. (e-mail: {\tt\small shuhan.shen@ia.ac.cn})
	\\Corresponding author: Xu Zhao}
}

%
%

\markboth{Journal of \LaTeX\ Class Files,~Vol.~14, No.~8, August~2015}%
{Shell \MakeLowercase{\textit{et al.}}: Bare Demo of IEEEtran.cls for Computer Society Journals}
%



\IEEEtitleabstractindextext{%
\begin{abstract}
Matching redundancy, which refers to fine-grained feature comparison between irrelevant image areas, is a prevalent limitation in current feature matching approaches. It leads to unnecessary and error-prone computations, ultimately diminishing matching accuracy.
To reduce matching redundancy, we propose MESA and DMESA, both leveraging advanced image understanding of \textit{Segment Anything Model} (SAM) to establish semantic area matches prior to point matching.
These informative area matches, then, can undergo effective internal feature comparison, facilitating precise inside-area point matching. 
Specifically, MESA adopts a sparse matching framework, while DMESA applies a dense one.
Both of them first obtain candidate areas from SAM results through a novel Area Graph (AG). 
In MESA, matching the candidates is formulated as a graph energy minimization and solved by graphical models derived from AG. 
In contrast, DMESA performs area matching by generating dense matching distributions on the entire image, aiming at enhancing efficiency.
The distributions are produced from off-the-shelf patch matching, modeled as the Gaussian Mixture Model, and refined via the Expectation Maximization. 
With less repetitive computation, DMESA showcases a speed improvement of nearly five times compared to MESA, while maintaining competitive accuracy. 
Our methods are extensively evaluated on \textbf{five} datasets encompassing both indoor and outdoor scenes. The results illustrate consistent and prominent performance improvements from our methods for \textbf{six} point matching baselines across all datasets. Furthermore, our methods exhibit promise generalization and improved robustness against image resolution.
Our code is publicly available at \href{https://github.com/Easonyesheng/A2PM-MESA}{\texttt{github.com/Easonyesheng/A2PM-MESA}}.
\end{abstract}

}

\maketitle

\IEEEdisplaynontitleabstractindextext

%
\IEEEpeerreviewmaketitle

\IEEEraisesectionheading{\section{Introduction}\label{sec:introduction}}

\IEEEPARstart{F}{eature} matching aims at establishing correspondences between images, which is vital in a broad range of applications, such as SLAM~\cite{orbslam}, SfM~\cite{colmap} and visual localization~\cite{loc}.
However, achieving exact point matches is still a challenge due to the presence of {matching noises}~\cite{matchsurvey}, including scale variations, viewpoint and illumination changes, repetitive patterns, and poor texturing.

Recent years have witnessed significant advancements in learning-based feature matching. Classical sparse matching methods have been revolutionized by learning detectors~\cite{superpoint}, descriptors~\cite{r2d2} and matchers~\cite{superglue,lightglue}.
Learning-based semi-dense~\cite{loftr,aspanformer} and dense~\cite{dkm,roma} methods further obtain an impressive precision gap over their sparse counterparts, by dense feature comparison across entire images. 
Nevertheless, all these matching methods encounter a common obstacle: \textbf{matching redundancy}, which involves detailed comparisons of learning features in irrelevant regions between images. 
These unnecessary computations are prone to the matching noises, limiting the matching precision.

\begin{figure}[!t]
\centering
\includegraphics[width=\linewidth]{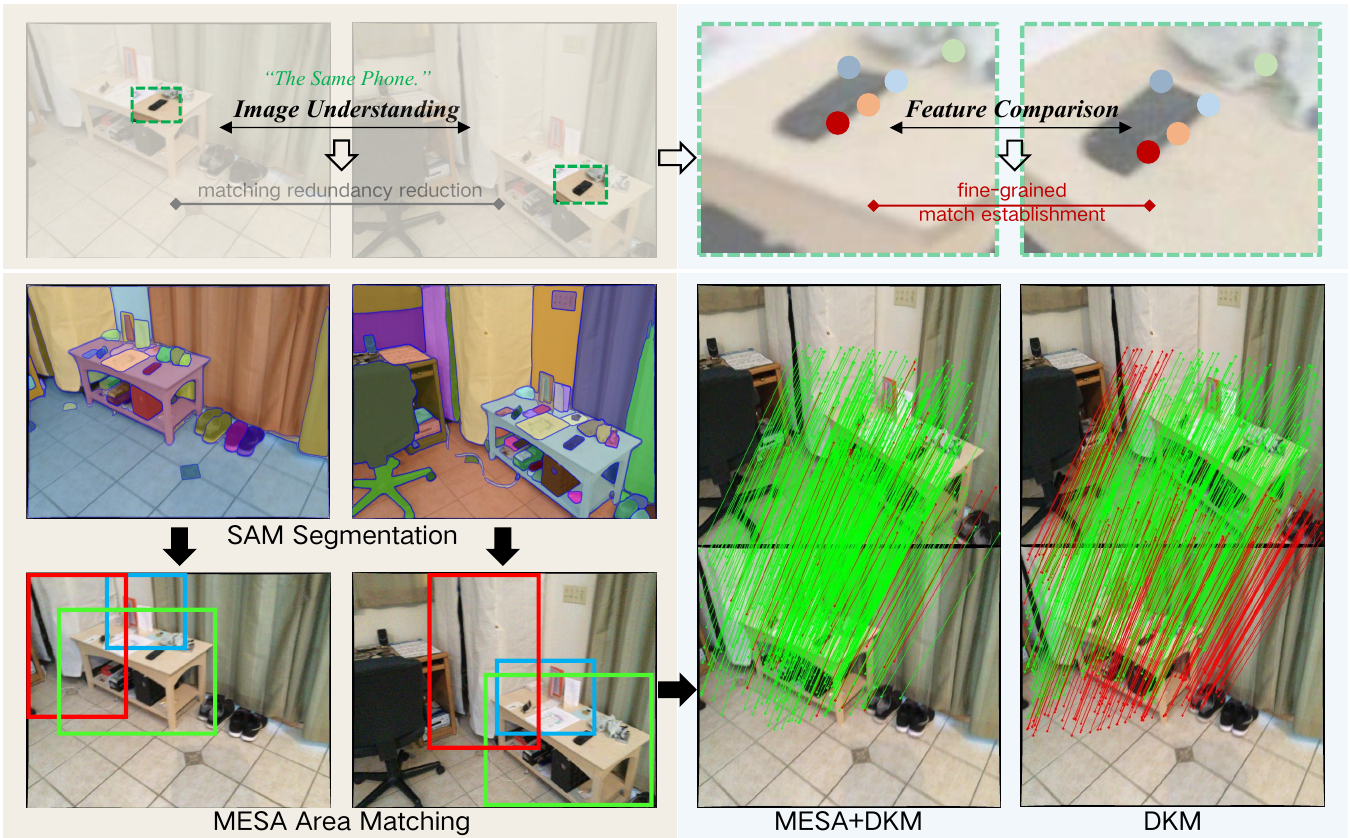}
\caption{\textbf{The matching redundancy reduction of our methods.} High-level \colorbox[rgb]{0.95,0.93,0.90}{image understanding} enables efficient matching redundancy reduction, allowing for precise point matching by local dense \colorbox[rgb]{0.95,0.97,0.99}{feature comparison}. Therefore, the proposed MESA effectively reduces the matching redundancy by area matching based on SAM~\cite{sam} segmentation, significantly improving the accuracy of DKM~\cite{dkm}. }
\vspace{-1.6em}
\label{fig:first}
\end{figure}

Intuitively, most of the matching redundancy can be effectively identified through high-level image understanding, and only strongly correlated local \textbf{areas} (or \textbf{regions}) need dense feature comparison to determine precise matches (cf. \cref{fig:first}).
Therefore, recent methods~\cite{topicfm,prism} perform learning-based redundancy pruning. However, \textit{implicit} learning leads to generalization challenges and lack of interpretability. 
To address these issues, some works turn to explicit image context~\cite{adamatcher,OETR,sgam}.
As the redundancy is evident in non-overlapping areas, overlapping segmentation is proposed~\cite{adamatcher,OETR}. However, the co-visible area is still rough, and the redundancy persists in it during subsequent matching. 
SGAM~\cite{sgam} provides a fine-grained way to reduce matching redundancy, named \textit{Area to Point Matching} (A2PM) framework.
Specifically, it establishes explicit semantic area matches before point matching, where the matching redundancy is largely removed according to semantic.
However, SGAM heavily relies on semantic segmentation.
Its performance, thus, decreases when encountering inexact semantic labeling and semantic ambiguity~\cite{sgam}.
Also, SGAM cannot be applied to general scenes due to the close-set semantic labels.
Hence, reducing matching redundancy by explicit semantic suffers from impracticality.

Recently, Segment Anything Model (SAM)~\cite{sam} has gained notable attention from the research community due to its exceptional performance and versatility, which can be the basic front-end of many tasks~\cite{midSAM,inpSAM}. This suggests that the foundation model can accurately comprehend image contents across various domains. Drawing inspiration from this, we realize that the image understanding of SAM can be leveraged to reduce matching redundancy.
Thus, we propose to establish area matches based on SAM segmentation to overcome the limitations of SGAM~\cite{sgam}.
Similar to the semantic segmentation, the SAM segmentation also provides multiple areas in images, but without semantic labels attached to these areas.
However, the general object perception of SAM ensures that its segmentation results inherently contain implicit semantic information. In other words, a complete semantic entity is always segmented as an independent area by SAM.
Hence, matching these \textit{implicit}-semantic areas also effectively reduces matching redundancy and promotes accurate point matching inside areas~\cite{sgam}.
Furthermore, the absence of explicit semantics alleviates the issues of inaccurate area matching caused by erroneous labeling. The limitation of generalization due to semantic granularity is also overcome.
Nevertheless, area matching cannot be simply achieved by semantic labels but requires other approaches under this situation.

\begin{figure*}[!t]
\centering
\includegraphics[width=\linewidth]{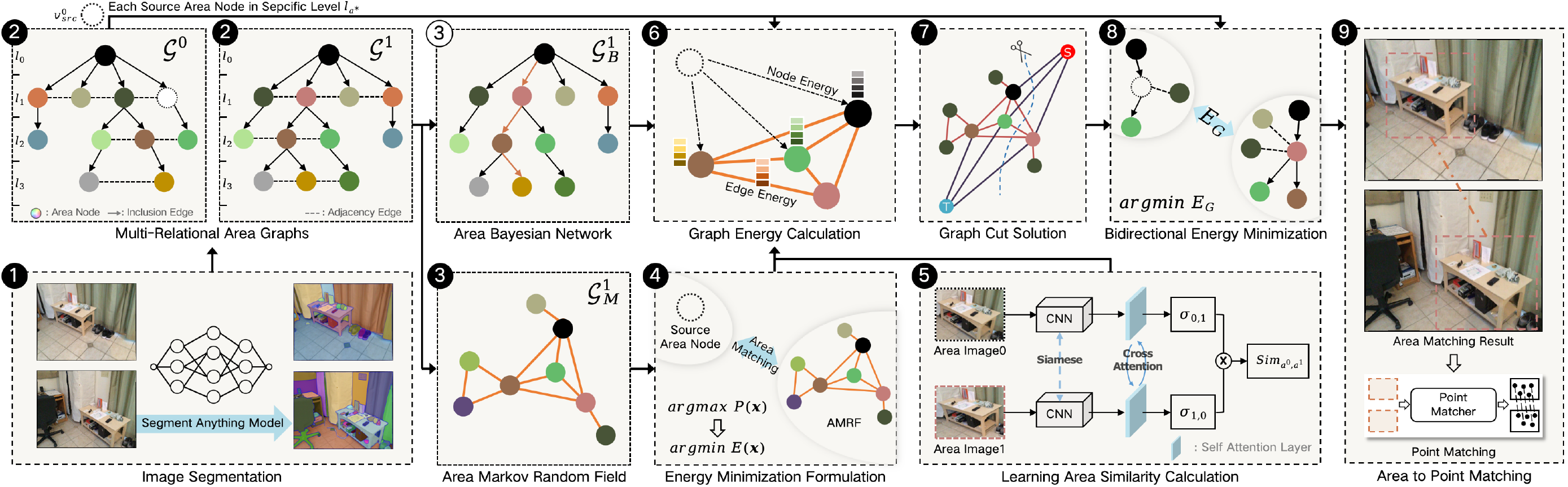}
\caption{\textbf{Overview of MESA.} Based on \ding{182}~\textit{SAM segmentation}, we first construct \ding{183}~\textit{Area Graph}s. Then the graph is turned to two graphical models based on its two different edges. Through \ding{184}~\textit{Area Markov Random Field}, area matching is formulated as an \ding{185}~\textit{Energy Minimization}. Meanwhile, leveraging \ding{174}~\textit{Area Bayesian Network} and our \ding{186}~\textit{Learning Area Similarity Calculation}, \ding{187}~\textit{Graph Energy} can be efficiently calculated. Therefore, \ding{188}~\textit{Graph Cut} is utilized to obtain putative area matches. Finally, \ding{189}~\textit{Bidirectional Energy Minimization} determines the best area match, which serves as the input of subsequent point matcher for precise feature matching, following the \ding{190}~\textit{Area to Point Matching} (A2PM) framework~\cite{sgam}.}
\vspace{-1.2em}
\label{fig_main}
\end{figure*}

In this work, we propose {Matching Everything by Segmenting Anything} (MESA, \cref{fig_main}), a method for precise area matching from SAM segmentation. 
MESA focuses on \textit{two} main aspects: \textbf{{area relation modeling}} and \textbf{area matching based on the relation}.
To be specific, since individual SAM areas provide only local information, matching them independently can lead to inaccurate results, especially in scenes with repetitiveness.
To address this, we construct a novel graph structure, named \textit{Area Graph} (AG), to model the global context of the areas as a basis for subsequent precise matching.
AG takes areas as nodes and connects them with two types of edges: undirected edges for adjacency and directed edges for inclusion.
Both edges capture global information, and the latter enables the construction of hierarchy structures similar to~\cite{quadtree} for efficient matching. 
After the \textbf{area relation modeling}, MESA performs \textbf{area matching} by deriving two graphical models from AG: \textit{Area Markov Random Field} (AMRF) and \textit{Area Bayesian Network} (ABN).
The AMRF involves all global-informative edges, thus allowing global-consistent area matching through energy minimization on the graph.
Specifically, the energy is determined based on the learning area similarity and spatial area relation, and this energy minimization is effectively solvable through \textit{Graph Cut}~\cite{GCE}.
The ABN, furthermore, is proposed to facilitate the graph energy calculation, leveraging the hierarchy structure.
Finally, we propose a global matching energy to tackle the issue of multiple solutions in \textit{Graph Cut}, ultimately leading to effective redundancy reduction from precise area matching.

Although MESA holds promise for high accuracy, its intricate process diminishes its efficiency.
To probe the root of this efficiency issue, we deeply review the matching procedure of MESA. Similar to the \textit{sparse} framework in point matching, MESA essentially operates as a \textit{sparse} area matching framework. It starts by extracting area candidates from images and subsequently conducts dense similarity computations between the candidate sets from the image pair. However, unlike points, determining area similarities among the sets involves repetitive computation due to area overlaps. Consequently, the efficiency drawback of MESA predominantly emerges from the costly computation of area similarities in the \textit{sparse} matching framework.

\begin{figure}[!t]
\centering
\includegraphics[width=\linewidth]{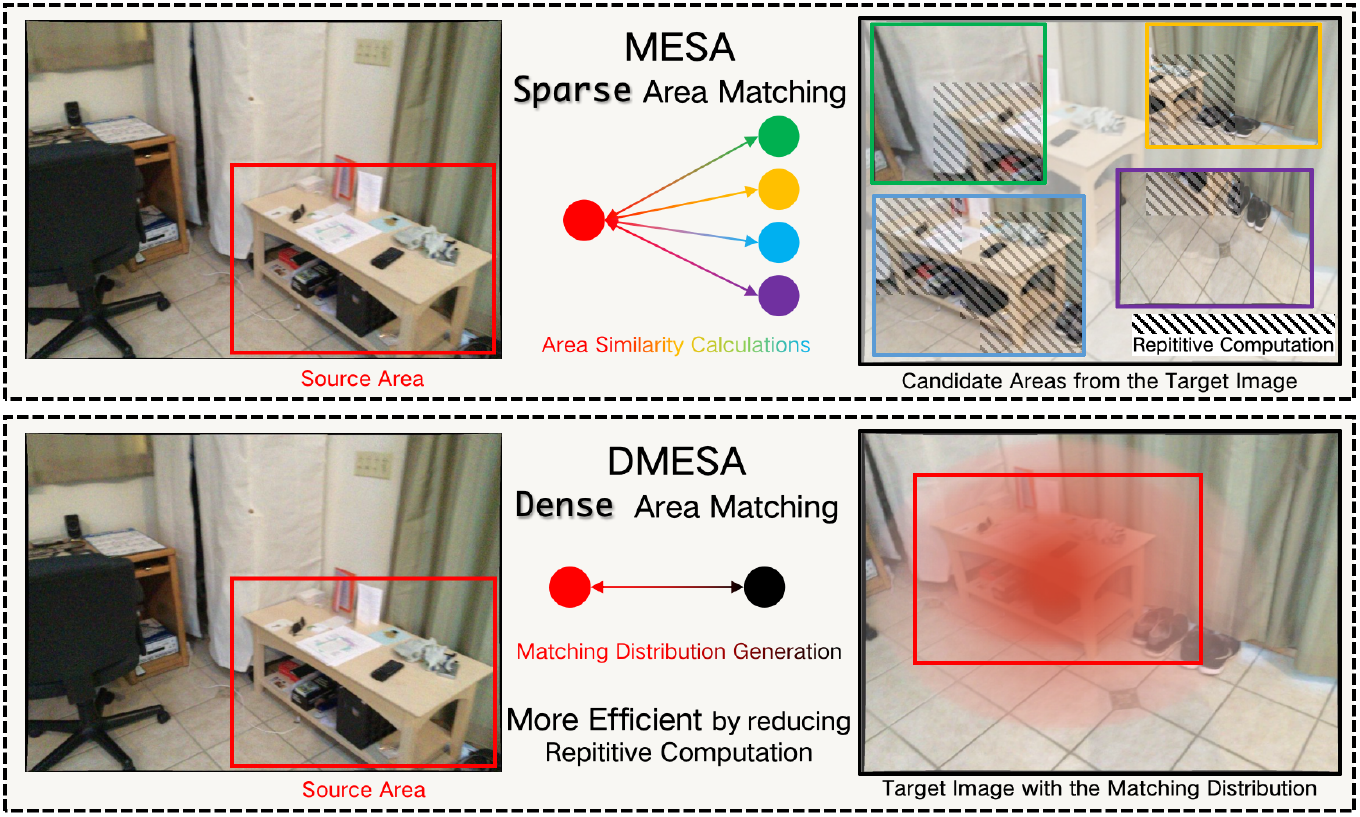}
\caption{\textbf{MESA vs. DMESA.} The sparse area matching framework of MESA involves repetitive computation in area similarity calculations, leading to an efficiency issue of MESA. To address this issue, DMESA leverages a dense matching distribution to guide the area matching, reducing repetitive computation.}
\vspace{-1.5em}
\label{fig:DvsMESA}
\end{figure}

To mitigate this concern, we take inspiration from the dense framework employed in point matching~\cite{loftr,dkm} and proposed a \textit{Dense} counterpart of MESA, named \textit{DMESA}, to conduct a dense area matching framework.
In contrast to common beliefs in point matching, a dense area matching framework is more efficient than a sparse one. 
This difference arises from the overlaps between the basic units. In point matching, there are no overlaps among the basic units (points). Thus, a dense framework that finds correspondences from the entire image incurs significantly higher computational costs than a sparse framework that only considers keypoints. 
However, in area matching, the basic units (areas) exhibit considerable overlap and often encompass the entire image (cf. \cref{fig:DvsMESA} top). Thus, repetitive computations are remarkable in the area similarities calculation of the sparse framework. Conversely, in a dense framework, these repetitive computations can be avoided by directly generating dense matching distributions on the entire image. Moreover, we notice that the matching distributions can be derived through patch matching, mirroring the coarse matching stage of current semi-dense point matchers~\cite{aspanformer}.

Consequently, DMESA focuses on utilizing patch matching to achieve area matching (cf. \cref{fig:dmesa}). Specifically, it first establishes patch matches between a source area and the target image, leveraging the \textit{off-the-shelf} coarse matching of~\cite{aspanformer}. Then, it models the joint matching distribution of these patch matches by the Gaussian Mixture Model (GMM), which can be viewed as a dense area matching distribution. Considering the accuracy concern of coarse matching, DMESA introduces the cycle consistency~\cite{silk} to optimize the distribution, adopting a finite-step Expectation Maximization (EM) algorithm. Afterwards, precise area matching can be obtained from the refined distribution. 

This work is an extension version of MESA~\cite{mesa} presented at CVPR'24. Here, we introduce the following technical enhancements and experimental contributions: \textbf{1)} After investigating the efficiency issue of MESA, we propose its \textit{dense} counterpart, named DMESA, applying a \textit{dense area matching framework}. DMESA enables area matching derived from off-the-shelf patch matching. In experiments, it can establish area matches with competitive accuracy at a speed nearly $5$ times faster than MESA, offering a better precision/efficiency trade-off; 
\textbf{2)} We observe a substantial impact from image resolution on feature matching in experiments. Thus, we conduct an in-depth analysis of this impact, resulting in an improved resolution configuration of A2PM. Moreover, we thoroughly examine image resolution in experiments, providing a more comprehensive evaluation of our methods;
\textbf{3)} We add two point matching baselines in the experiments, to prove our methods can benefit all existing types of point matchers. By employing a more reproducible experimental setup, we present new results for previous experiments and conduct experiments on two additional indoor and outdoor datasets. Furthermore, experiments about cross-domain generalization, model fine-tuning and using SAM2~\cite{sam2} segmentation are conducted in this version.

Our work makes several contributions. 
\textbf{1)} To effectively reduce matching redundancy, we propose utilizing the high-level image comprehension capability of SAM. To this end, we present two methods, \textit{MESA} and \textit{DMESA}, for implicit semantic area matching from SAM segmentation, and ultimately improving matching accuracy. 
\textbf{2)} Applying the sparse matching framework, \textit{MESA} first extracts semantic areas from SAM results by a novel graph, termed AG, which models the global area relations. Based on graphical models derived from AG, precise area matching is achieved for accurate inside-area point matching.
\textbf{3)} To improve the efficiency of MESA, we further introduce \textit{DMESA}, which employs a dense framework. It conducts area matching by generating dense matching distributions on the entire image. DMESA offers greater flexibility and speed, striking a superior balance between accuracy and efficiency.
\textbf{4)} In extensive experiments on five diverse datasets, our methods consistently yield substantial performance improvements for six point matchers spanning sparse, semi-dense, and dense matching categories, showing their versatility. Moreover, our methods exhibit prominent generalization across various datasets and superior robustness against the input image resolution.

\section{Related Work}
\subsection{Sparse, Semi-Dense and Dense Matching}
There are three types of feature matching methods: sparse, semi-dense and dense.
Classical feature matching methods~\cite{sift,orb} belongs to the sparse framework, which involves keypoint detection and description in images and matching among keypoint sets. 
The learning counterpart of this framework utilizes neural networks to perform feature detection~\cite{keyNet,superpoint}, description~\cite{d2net,r2d2,alike} or matching~\cite{superglue,OANet}.
To avoid the detection failure in sparse methods, semi-dense methods~\cite{loftr,aspanformer,quadtree,pats} are proposed, also known as the detector-free methods.
These methods perform dense feature comparison over the entire image and then select confident patch matches, which are used to refine precise point matches.
Dense matching methods~\cite{dkm,roma} output a dense warp with confidence map for the image pair. 
Recent DKM~\cite{dkm} gradually refines the dense warp from small to large resolution, which can also be viewed as patch matching from coarse to fine, and achieves \textit{state-of-the-art} performance.
However, MESA and DMESA focus on reducing redundancy in feature matching through area matching. Thus, they can be seamlessly combined with all kinds of point matching methods described above through the A2PM framework, to increase matching precision.

\subsection{Implicit Matching Redundancy Reduction}
In general, the establishment of keypoints sets (sparse matching) or patch matching (dense/semi-dense matching) can also be viewed as reducing matching redundancy. However, their implementations rely on feature computation spanning the entire image, inherently containing a significant amount of redundancy. Particularly, feature interactions in irrelevant areas lead to decreased matching accuracy. 
Thus, current methods for removing matching redundancy concentrate on eliminating these error-prone feature computations.
Following learning-based matching methods, TopicFM~\cite{topicfm} infers topics for learning features and then confines feature calculation to the same topic to avoid redundant computation. Similarly, PRISM~\cite{prism} generates feature masks based on mutual information to restrict interactions between features with low similarity.
However, both the feature topics and mutual information masks are attained by implicit feature learning, lacking a clear connection to the image context. Consequently, these methods encounter challenges in generalization.
Our methods utilize explicit semantic area matching to diminish matching redundancy, enhancing both interpretability and generalization.

\subsection{Explicit Matching Redundancy Reduction}
A well-defined intermediate search space leads to explicit matching redundancy reduction.
Initially, as the presence of redundancy is apparent in non-overlapping regions between images, several studies focus on extracting covisible areas. They predict overlaps between images by iterative matching~\cite{cotr} or overlap segmentation~\cite{OETR,oamatcher,mkpc}. However, within the overlapping regions, matching redundancy persists, especially in the context of detailed local point matching.
In contrast, SGAM~\cite{sgam} establishes semantic area matches between images to achieve refined reduction of matching redundancy.
Then, point matches are obtained by dense feature comparison inside the matched areas. This A2PM framework is simple yet effective. Our methods further build on its advantages, but leverage the advanced segmentation method, \ie SAM, to overcome the issues from explicit semantic.

\section{Preliminaries: A2PM Framework}
In this section, we introduce the task of feature matching and the A2PM framework.
The ultimate goal of feature matching is to establish point matches ($\mathcal{P}$) between images ($I_0, I_1$), also known as point matching (PM).
\begin{equation}
	 PM(I_0, I_1) = \mathcal{P}~\text{with}~\mathcal{P}:=\{(p_0^i,p_1^i)\}_i.
\end{equation}
Here, $(p_0^i \in I_0, p_1^i \in I_1)$ denotes the 2D projections in two images of the same 3D point, \ie, a point match. To reduce the redundancy in above matching, the Area to Point Matching (A2PM) framework~\cite{sgam} is proposed. Both Our MESA and DMESA adhere to this framework and focus on its area matching phase.

Firstly, we revisit the core idea of the A2PM framework in \cref{sec:a2pm-o}. Then, due to the pivotal role of image resolution in the A2PM framework, we analyze its effect on matching accuracy (\cref{sec:resolution}). Based on this, we further offer a detailed resolution configuration of the framework and discuss its motivation (\cref{sec:a2pm-c}).

\subsection{Overview of A2PM Framework}\label{sec:a2pm-o}
The initial motivation of~\cite{sgam} is to improve the search space for feature matching by semantic.
To this end, a semantic-friendly search space was proposed, known as the \textit{semantic area matches}.  
By matching these areas, matching redundancy between images could be effectively eliminated by semantic. Thus these area pairs could provide more local details, benefiting inside point matching. 
Then, the A2PM framework is introduced, which focuses on the coupling of area matching ({AM}) and point matching. In this framework, the semantic area matches ($\{a_0^i, a_1^i\}_i$) are first established by AM between images, followed by PM within the area pairs cropped from original images. Finally, fusing these inside-area point matches yields the ultimate matching results $\mathcal{P}$.
\begin{equation}
 AM(I_0,I_1) \xrightarrow{crop} \{PM(a_0^i,a_1^i)\}_i \xrightarrow{fuse} \mathcal{P}.
\end{equation}
This flexible combination enables independent development of AM techniques to improve the matching precision of various PM methods. Meanwhile, it is evident that accurate AM is the basis of A2PM. Our MESA and DMESA thus focus on achieving precise AM, and showcase consistent improvement for sparse, semi-dense and dense point matchers in extensive experiments.

\subsection{Image Resolution Impact on Feature Matching}\label{sec:resolution}
The impact of resolution on matching is rarely explored in current literature, as common point matchers typically resize the raw images directly to a default resolution (\eg, the training resolution). 

However, with the increasing computational demands of advanced matching methods (\eg, quadratic to input resolution for some semi-dense/dense methods~\cite{dkm,cotr}), the choice of resolution becomes a practical trade-off between accuracy and efficiency. Especially in scenarios with limited computational resources, point matchers may opt for reduced image sizes below the default, raising concerns about the impact of resolution variation on matching precision.

Moreover, in the A2PM framework, one more resolution is taken into account, that is the \textit{area image resolution}. It potentially leads to conflicts with the default PM resolution and raises additional concerns regarding resolution and matching accuracy.

This motivates us to discuss the influence of image resolution on matching performance here and experimentally investigate it in \cref{sec:exp}. 
Generally, resolution impacts feature matching in three primary ways.
\textbf{1)} Resolution essentially reflects the level of \textit{detail} preserved in images and higher resolution ideally enhances matching accuracy.
\textbf{2)} Changes in image resolution can lead to changes in \textit{aspect ratio}, causing \textit{distortion} in image content that can reduce matching accuracy. 
\textbf{3)} Learning-based methods may demonstrate \textit{overfitting to the training resolution}. Especially for semi-dense matchers, resolution variation may lead to significant performance declines (experimentally investigated in \cref{sec:roi} of the appendix), probably due to their Transformer-based structure~\cite{NaViT}. 

\subsection{Image Resolution Configuration in A2PM}\label{sec:a2pm-c}
Building on the findings in the last part, we offer a detailed resolution setting for the A2PM framework. Specifically, the original image resolution is set to the resolution of the raw images in the dataset. The resolution of area images and PM input share the same aspect ratio, which is set as $1$. Therefore, in A2PM, the specific cropping operation \textbf{first expands} the shorter side of areas to form squares, \textbf{then crops} them from the high-resolution raw images, and finally scale them to the required and \textit{square} input resolution. This setting is experimentally confirmed in \cref{sec:ab-ic}.

The reasons for the setting are as follows. 
\textbf{1)} Due to the accurate AM of our methods, matching redundancy is sufficiently reduced in area images. Thus, most of the inside-area pixels containing useful details for PM. Cropping areas from the high-resolution original image preserves these details as much as possible, thereby benefiting the matching accuracy.
\textbf{2)} The same aspect ratio between area images and PM input avoid distortion during resize. 
\textbf{3)} The aspect ratio constraint of $1$ arises from the uncertainty in sizes of semantic areas. In other words, the sizes of semantic areas are determined by specific semantics, whose aspect ratios vary across different images. However, point matchers require input image pairs to have the same dimensions. Therefore, we set a uniform input resolution with aspect ratio of $1$, which leads to minimum changes in area size in a statistical sense, thus reducing redundancy introduction in area size adjustment.

Next, we discuss the impact of learning resolution overfitting on the A2PM framework.
This overfitting issue mainly arises from the aspect ratio conflict between the training resolution of PM and its input resolution in A2PM framework.
When the training resolution has an aspect ratio of 1, this overfitting is harmless (cf. \cref{tab:YFCC+MD}). 
Conversely, when the training aspect ratio deviates from 1, the performance of A2PM methods using square inputs may be inferior to the original PM using training size, due to the overfitting issue. 
Meanwhile, using the training resolution as the input PM size in A2PM framework may also lead to a decrease in matching precision from excessive area size adjustments (cf. \cref{tab:SN}).
However, in experiments, we \textbf{only} observe the drop in performance specifically with \textit{Transformer-based methods} on the \textit{in-domain indoor dataset}, corresponding to their resolution overfitting issue (cf. \cref{sec:roi}). To alleviate this issue, we can fine-tune the models on a square resolution (cf. \cref{sec:ab-irmf}). However, this technique, while effective, entails additional training expenses. The optimal approach should focus on addressing the overfitting issue of Transformer in PM, similar to~\cite{NaViT}. We hope that this will inspire additional progress in PM within the community.

\section{Sparse Area Matching}\label{sec:mesa}
In this section, we introduce a sparse AM approach, called \textit{MESA}, which leverages SAM to effectively reduce matching redundancy. It initially identifies candidate semantic areas in images and subsequently matches these candidates, akin to sparse PM methods. 
There are two main components of MESA: the \textit{Area Graph} (AG, \cref{ssec:ag}) and the \textit{Graphical Area Matching} (\cref{ssec:gam}). The former is a novel graph that describes inter-area relations and serves as a basis for AM. The later is responsible for finding area matches utilizing both the inter-area relations and intra-area features.

\subsection{Area Graph}\label{ssec:ag}
The main motivation to propose AG is that direct AM on SAM results is inaccurate, as global information is ignored in independent areas.
Fixed area sizes from SAM also hinder robust PM under scale changes.
Hence, AG is designed to capture the global structure of these areas and construct scale hierarchy for them, by modeling inter-area relations.

Subsequently, we first introduce the definition of AG in \cref{sssec:agd} and then explain how to construct AG from SAM results in \cref{sssec:gc}.

\subsubsection{Area Graph Definition}\label{sssec:agd}
\begin{figure}[!t]
\centering
\includegraphics[width=\linewidth]{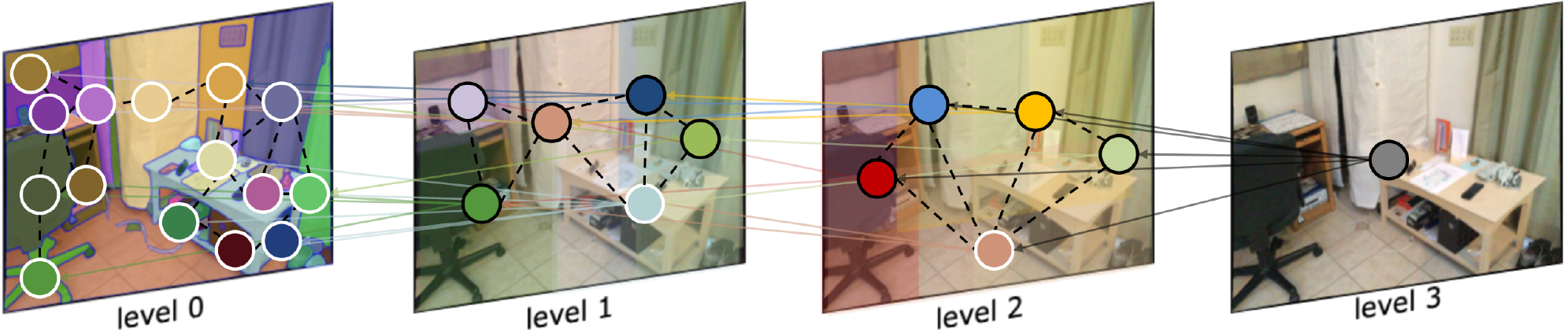}
\caption{\textbf{Area Graph.} The graph nodes (circles with masks representing rectangle areas) includes both areas from SAM (white boundaries) and our graph completion algorithm (black boundaries). They are divided into various levels according to their sizes. The adjacency edges (dashed lines) and inclusion edges (arrows) connect these nodes. Only adjacency edges within the same level are shown for better view.}
\vspace{-1.2em}
\label{fig:ag}
\end{figure}

AG ($\mathcal{G}\!=\!\langle \mathcal{V}, \mathcal{E}\rangle$) takes image areas as nodes and contains two edges to model inter-area relations (\cref{fig:ag}), thus making it a \textit{multi-relational graph}~\cite{mrg}.
The graph nodes include both areas provided by SAM and additional areas generated for scale hierarchy (cf.~\cref{sssec:gc}). They are divided into different levels according to their sizes.
On the other hand, the graph edges ($\mathcal{E}\! = \!\mathcal{E}_{in} \bigcup \mathcal{E}_{adj}$) represent two relations between areas, \ie, inclusion ($\mathcal{E}_{in}$) and adjacency ($\mathcal{E}_{adj}$).
The inclusion edge $e_{in}\!\in\!\mathcal{E}_{in}$ is directed, pointing from an area to one of its containing areas.
It forms a tree-like connection between graph nodes, enabling robust and efficient AM under scale changes.
The adjacency edge $e_{adj}\!\in\!\mathcal{E}_{adj}$ is undirected, indicating the areas it connects share common parts but without the larger one including the smaller one.
This edge captures the spatial relations among areas, beneficial to accurate AM.
By the above two edges, AG models the global structure of image areas, playing a fundamental role in our AM.

\subsubsection{Area Graph Construction}\label{sssec:gc}
The construction of AG includes collecting areas as nodes and connecting them by proper edges.
Notably, not all SAM areas can function as nodes, since some are too small or have extreme aspect ratios, rendering them unsuitable for inside PM. 
Thus, \textbf{Area Pre-processing} is performed first to obtain initial graph nodes.
We then approach the edge construction as a \textbf{Graph Link Prediction} problem~\cite{lpSurvey}.
Afterwards, the preliminary AG is formed, but it still lacks matching efficiency and scale robustness.
Thus, we propose the \textbf{Graph Completion} algorithm, which generates additional nodes and edges to construct the scale hierarchy. 

\boldparagraph{Area Pre-processing:}
To filter unsuitable areas, we set two criteria: the acceptable minimal area size ($T_s$) and maximum area aspect ratio ($T_r$).
Any area that has smaller size than $T_s$ or larger aspect ratio than $T_r$, gets screened out.
The remaining areas are added into the candidate set.
For each filtered area, we fuse it with its nearest neighbor area in the candidate set.
The filtering and fusion operations are repeated until no areas get screened out.
Then, we assign a level $l_a$ to each candidate area $a$ based on its size, by setting $L$ size thresholds ($\{TL_i~\big|~i\in[0,L-1]\}$):
\begin{equation}
    l_a = i \; \big| \; TL_{i} \leq W_a \times H_a < TL_{i+1}.
\end{equation}
These size levels are the basis of scale hierarchy in AG.

\boldparagraph{Graph Link Prediction:}
The edge construction is treated as a link prediction problem.
Given two area nodes ($v_i,v_j$), the edge between them ($e_{ij}$) can be predicted according to the spatial relation of their corresponding areas ($a_i, a_j$).
This approach adopts the ratio ($\delta$) of \textit{the overlap size} ($O_{ij}$) to \textit{the minimum size} between two areas ($\delta = {O_{ij}}/{\min (W_i \times H_i,W_j \times H_j )}$) as the score function:
\begin{equation}\label{eq:lp}
e_{ij} \in \left\{
	\begin{aligned}
	&\mathcal{E}_{in} &,\; &\delta >= \delta_h \\
	&\mathcal{E}_{adj} &,\;&\delta_l < \delta < \delta_h \\
	 &\varnothing  &,\; &\delta \leq \delta_l\\
	\end{aligned}
	\right
	.
    ,
\end{equation}
where $\delta_l, \delta_h$ are predefined thresholds.

\boldparagraph{Graph Completion:}
Initial AG is achieved by connecting all the processed nodes using different edges.
However, since SAM inherently produces areas containing complete entity, there are few inclusion relations among areas.
Consequently, initial AG lacks the scale hierarchy, which reduces its robustness at scale variations and makes accessing nodes inefficient.
To address the issue, we propose the \textit{Graph Completion} algorithm. It generates additional nodes and edges to ultimately construct a complete tree structure in the original graph.
The core of this algorithm is to generate parent nodes for each \textit{orphan} node, which has no parent node in the next higher level.
The algorithm begins at the smallest level, collects all orphan nodes, and clusters them based on their center locations.
Nodes in the same cluster have their corresponding areas fused with each other. It is noteworthy that the generated areas containing multiple objects preserve the internal implicit semantic. 
Based on proper level thresholds, the resulting areas correspond to new higher level nodes.
If a node remains single after clustering, we increase its area size to the next level to allow for potential parent nodes.
We repeat the above operations on the next level and connect generated nodes to others by suitable edges, until the highest level is reached.
More details can be found in the \cref{sec:d-agc} of the appendix. 

\subsection{Graphical Area Matching}\label{ssec:gam}
In this part, we describe the AM process in MESA, which is formulated on the graph, based on two graphical models derived from AG.
Given two AGs ($\mathcal{G}^0, \mathcal{G}^1$) of the input image pair ($I_0, I_1$) and one area ($a^0_{src}\!\in\!I_0$) corresponding to the node $v^0_{src}\!\in\!\mathcal{G}^0 $ (termed as the \textit{source node}), AM involves finding the node $v^1_j\!\in\!\mathcal{G}^1 $ with the highest probability of matching its area $a^1_j$ to the source node area $a^0_{src}$.
However, treating this problem as an independent node matching is inadequate, as which disregards global structure of areas.
Since the global structure is modeled in AG by its edges, considering the graph edges, thus, is essential for accurately matching these areas. 
Meanwhile, the two edges of AG respectively derive two graphical models, \ie Markov Random Fields (undirected edges) and Bayesian Network (directed edges).
These observations motivate us to formulate the AM task inside the framework of graphical model.

In the following, we first introduce the undirected graph converted from AG, named Area Markov Random Field (AMRF, \cref{sec:amrf}), which is leveraged to formulate the AM into an energy minimization task. To calculate the local matching energy between areas, we propose a learning model in \cref{sec:las} to achieve area similarities based on intra-area features. 
Then, the directed graph converted from AG, termed as Area Bayesian Network (ABN, \cref{sec:abn}) is presented to reduce redundant computation in the energy calculation. Finally, to achieve the best area match, an energy-based refinement is proposed (\cref{Sec:refine}), through considering the graph structures of both input images.

\subsubsection{Area Markov Random Field}\label{sec:amrf}
By considering the general adjacency relation, which includes the inclusion relations as adjacency too, the $\mathcal{G}^1$ is transformed into an undirected graph.
Then, random variables ($\bm{x}$) are introduced for all nodes to indicate their matching status with the source node.
The binary variable $x_i\in \bm{x}$ is equal to $1$ when $v^1_i$ matches $v^0_{src}$ and $0$ otherwise.
Therefore, the AMRF ($\mathcal{G}^1_M = \langle \mathcal{V}, \mathcal{E}_{adj} \rangle$) is obtained.
As these undirected edges imply the global consistency of matching, AM can be performed by maximizing the joint probability distribution over the AMRF:
\begin{equation}
    \arg\max_{\bm{x}} P(\bm{x}).
\end{equation}
Based on~\cite{markov}, the probability distribution defined by AMRF belongs to the \textit{Boltzmann distribution}, which is an exponential of negative energy function ($P(\bm{x})\!=\!\exp(-E(\bm{x}))$). 
Therefore, the AM can further be formulated as an energy minimization.
\begin{equation}\label{eq:em}
    \arg\min_{\bm{x}} E(\bm{x}). 
\end{equation}
The energy can be divided into two parts, including the energy of nodes ($E_{\mathcal{V}}$) and edges ($E_{\mathcal{E}}$), based on the graph structure.
\begin{equation}\label{eq:e}
    E(\bm{x})=\sum_iE_{\mathcal{V}}(x_i)+\lambda\sum_{(i,j)\in\mathcal{N}}E_{\mathcal{E}}(x_i,x_j),
\end{equation}
where $\lambda$ is a parameter balancing the terms and $\mathcal{N}$ is the set of all pairs of neighboring nodes.
For a graph node $v^1_i$, its energy is expected to be low when its matching probability is high, which can be reflected by the apparent similarity ($S_{a^0_{src}a^1_i}$) between $a^0_{src}$ and $a^1_i$. This energy term corresponds to the intra-area features.
\begin{equation}
    E_{\mathcal{V}}(x_i)=|x_i-S_{a^0_{src}a^1_i}|.
\end{equation}
The edge energy aims to penalize all neighbors with different labels, and the Potts model~\cite{potts} ($T$) would be a justifiable choice.
To better reflect the spatial relation, the Potts interactions are specified by $IoU$~\cite{IoU} of neighboring areas. This energy term corresponds to the inter-area relations.
\begin{equation}
    E_{\mathcal{E}}(x_i,x_j)=IoU(a^1_i,a^1_j) \cdot T(x_i\not=x_j).
\end{equation}
Function $T(\cdot)$ is $1$ if the argument is true and $0$ otherwise.
Finally, the AM is formulated as an binary labeling energy minimization.
By carefully defining the energy function, the energy minimization problem in \cref{eq:em} is efficiently solvable via the \textit{Graph Cut} algorithm~\cite{GC,GCE}.
The obtained minimum cut of the graph $\mathcal{G}^1_M$ is the matched node set ($\{v^1_{h}\big | h \in {\mathcal{H}}\}$).
Although the set may contain more than one area node, the best matching result can be achieved from this set by our refinement algorithm in \cref{Sec:refine}.

\begin{figure}[!t]
\centering
\includegraphics[width=\linewidth]{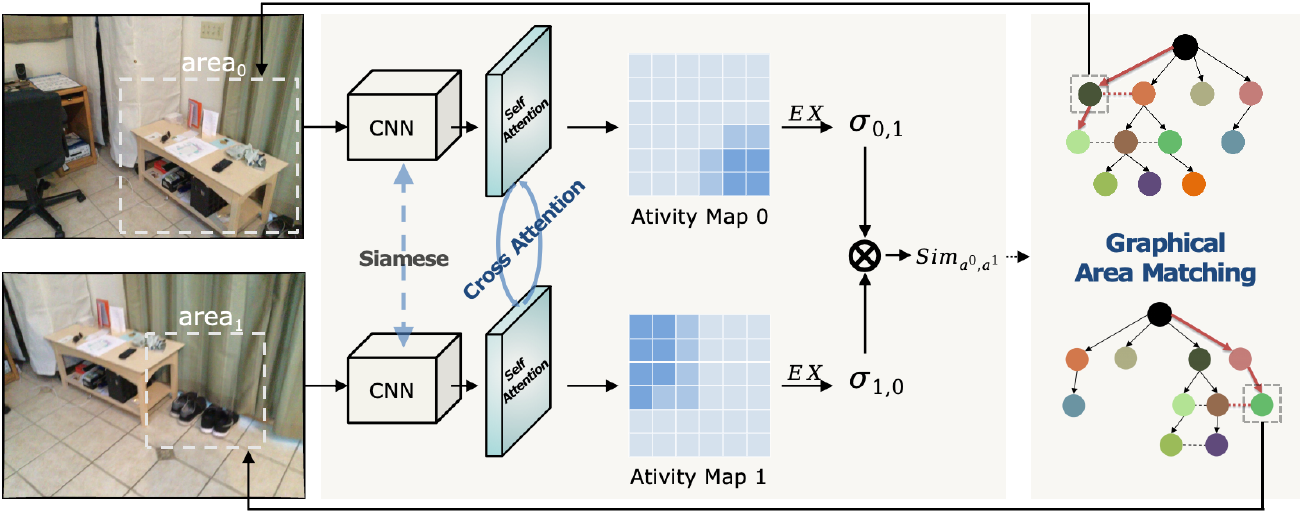}
\caption{\textbf{Learning area similarity.} The area similarity calculation is formed as the patch-level classification. We predict the probability of each patch in one area appearing on the other to construct activity maps. The similarity is obtained by the product of activity expectations, contributing to our exact AM.}
\vspace{-1.2em}
\label{fig:las}
\end{figure}

\subsubsection{Learning Area Similarity}\label{sec:las}
The proposed \textit{Graph Cut} solution relies on energy calculations for graph nodes and edges. 
Unlike easily available \textit{IoU} of areas for $E_{\mathcal{E}}$, determining the area apparent similarity for $E_{\mathcal{V}}$ is not straightforward.
Thus, we turn to the learning-based framework, inspired by recent successes of learning models in PM~\cite{aspanformer}.
A straightforward idea is to calculate the correlation of learning descriptors of two areas~\cite{pats} as the area similarity.
However, the descriptor correlation is too rough for the accurate AM and lacks fine-grained interpretability.
To overcome these issues, we decompose the area similarity calculation into two parallel patch-level classification problems as shown in \cref{fig:las}.

Specifically, for each image in the area image pair $\{a_j\big|j\in\{0,1\}\}$ reshaped to the same size, we perform binary classification for each $8\times 8$ image patch $p^j_i$ (\textit{where $i$ is the index of patch and $j$ is the index of area image}) in it, computing the probability of $p^j_i$ appearing on the other area image, termed as the patch activity $\sigma^j_i$.
To accomplish the classification, we first extract patch-wise features from each area image using a Siamese CNN~\cite{siamese}.
Then we update these patch features via self and cross-attention with normalization~\cite{attention}, resulting in patch activities.
Utilizing these patch activities, we construct an activity map ($\sigma^j_m = \{\sigma^j_i\big | j \in \{0,1\}\}_i$) for each area image.
When two areas are ideally matched, the corresponding 3D point of every pixel in one area is projected onto the other area.
Hence, all the patch activities of both areas should be closed to $1$, revealing the area similarity can be achieved by the product of expectations (EX) of two activity maps.
\begin{equation}
    Sim_{a^0,a^1} = EX(\sigma_m^0) \times EX(\sigma_m^1) := \sigma_{0,1} \times \sigma_{1,0}.
\end{equation}
Through this approach, the calculation of area similarity is transformed into the patch-level classification, which enhances the interpretability and accuracy of AM.

\subsubsection{Area Bayesian Network}\label{sec:abn}
Although the \textit{Graph Cut} can be accomplished in polynomial time~\cite{GC}, the dense energy calculation over $\mathcal{G}_M^1$ is time consuming.
Furthermore, due to the scale hierarchy in AG, this dense calculation is highly redundant.
In particular, if the source area $a^0_{src}$ is not matched to $a^1_j$, it won't be matched to any children area of $a^1_j$.
This observation reveals the conditional independence in the similarity calculation, which involves inclusion edges in $\mathcal{G}^1$, thus turning $\mathcal{G}^1$ into a Bayesian Network ($\mathcal{G}^1_B$)~\cite{prml}.
Based on $\mathcal{G}^1_B$, the redundancy in the similarity calculation can be reduced.
In practice, we calculate the dense similarities by constructing a similarity matrix $M_S\!\in\!\mathbb{R}^{|\mathcal{V}^0| \times |\mathcal{V}^1|}$. Note \textit{not} all similarities in $M_S$ need calculation, but any similarity can be accessed in $M_S$. We first calculate similarities directly related to all source nodes. Subsequent calculations are saved in $M_S$ as well.
For $M_S[i,j]$ that has not been acquired, we calculate it by our learning model:
\begin{equation}
    M_S[i,j] = Sim_{a^0_{i}, a^1_j}.
\end{equation}
If $M_S[i,j] < T_{as}$, all children nodes $\{v^0_{h}\big | h\in ch^0(i)\}$ and $\{v^1_{c}\big | c\in ch^1(j)\}$ of $v^0_{i}$ and $v^1_j$ are found from $\mathcal{G}_B^0$ and $\mathcal{G}_B^1$, where $ch^j(i)$ is the index set of children indices of node $v_i^j$ from $\mathcal{G}^j_B$.
Based on the conditional independence, we have:
\begin{equation}\label{eq:abn}
    M_S[h,k] = 0,~\forall (h,k) \in ch^0(i) \times ch^1(j).
\end{equation}
This operation effectively reduce the redundancy in similarity calculation, leading to more efficient AM.

\subsubsection{Bidirectional Matching Energy Minimization}\label{Sec:refine}
The minimum cut $\{v^1_{h} \big | h\! \in \! {\mathcal{H}}\}$ achieved through the \textit{Graph Cut} may contain more than one area node, indicating further refinement is necessary to obtain the best area match.
Moreover, the aforementioned \textit{graphical area matching}, \ie finding the corresponding area node in $\mathcal{G}^1$ for $v^0_{src}\!  \in\!  \mathcal{G}^0 $, only considers the structure information of $\mathcal{G}^1$ and ignores the structure of $\mathcal{G}^0$.
To overcome this limitation, we propose a bidirectional matching energy $E_G$ for each candidate node $v^1_{h}$, consisting of four parts:
\begin{equation}\label{eq:eg}
    \begin{aligned}
       E_G(v^1_{{h}}) & = \frac{1}{Z}( \mu \cdot E_{self}(v^1_{{h}})+\alpha  \cdot  E_{parent}(v^1_{{h}}) \\ 
         & +\beta \cdot E_{children}(v^1_{{h}})+\gamma \cdot E_{neighbor}(v^1_{{h}}))
    \end{aligned},
\end{equation}
where $\mu, \alpha, \beta$ and $\gamma$ are weights to balance the terms; $Z$ is the partition function.
The $E_{self}(v^1_{{h}})$ is the energy related to matching probability between $v^0_{src}$ and $v^1_{{h}}$:
\begin{equation}\label{eq:e_self}
    E_{self}(v^1_{{h}}) = |1 - Sim_{a^0_{src},a^1_h}|.
\end{equation}
The $E_{parent}(v^1_{{h}})$ is the energy related to matching probability between the parent node pairs of $v^0_{src}$ and $v^1_{{h}}$:
\begin{equation}
\centering
\begin{gathered}
         E_{parent}(v^1_{{h}}) = \\ 
         \min \{|1 - Sim_{a^0_{u},a^1_{r}}| \big|  u \in p^0(src), r \in p^1(h)\},
\end{gathered}
\end{equation}
where $p^i(j)$ is the index set of parent nodes of $v^i_j$ in $\mathcal{G}_i$.
This energy is the minimum matching energy among all parent node pairs of $v^1_{{h}}$ and $v^0_{src}$.
Same as the $E_{parent}$, the $E_{children}$ and $E_{neighbor}$ are the energy terms of children and neighbor node pairs.
Afterwards, the best area match $v^1_{h^*}$ in the set can be found by minimizing $E_G$:
\begin{equation}
     h^* = \arg\min_{h \in {\mathcal{H}}} E_G(v^1_{h}).
\end{equation}
If the $ E_G(v^1_{h^*}) > T_{E_{max}}$ (a threshold parameter), the source area node $v^0_{src}$ is considered to have no matches.
To further improve the accuracy of final match, we set an energy range threshold $T_{Er}$ to collect all the candidates within a certain energy range.
\begin{equation}\label{eq:collectarea}
    \{v^1_{\bar{h}}\big||E_G(v^1_{\bar{h}}) -  E_G(v^1_{h^*})| \leq T_{Er}, \bar{h} \in {\mathcal{H}} \}.
\end{equation}
Then the final area match is achieved by \textit{fusing} $v^1_{h^*}$ and all candidates $\{v^1_{\bar{h}}\}_{\bar{h}}$, using $E_G$ as weights, utilizing the method proposed in \cite{pats}.
This refinement completely considers the structure information of both $\mathcal{G}^0$ and $\mathcal{G}^1$ and achieves exact area matches.


\begin{figure}[!t]
\centering
\includegraphics[width=\linewidth]{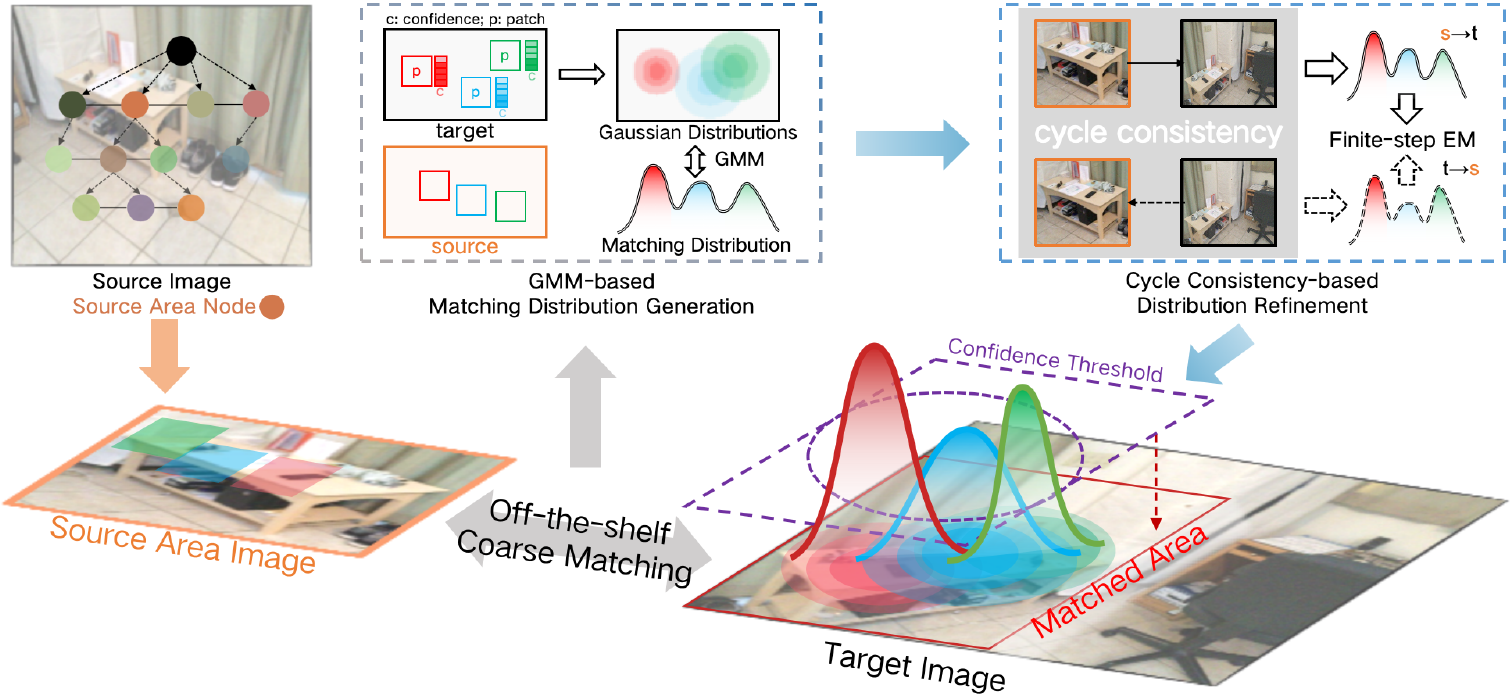}
\caption{\textbf{Overview of DMESA.} DMESA derives area matches from dense patch matches between the source area image and the target image, which are obtained through an off-the-shelf coarse matching. We model the patch matches utilizing GMM to generate a matching distribution in the target image, where the cycle-consistency can be introduced by a finite-step EM algorithm for accuracy refinement. Then, the area matches can be efficiently attained by applying a confidence threshold.}
\label{fig:dmesa}
\end{figure}

\section{Dense Area Matching}\label{sec:dmesa}
The proposed MESA is robust and precise. Nonetheless, its primary drawback is efficiency-related, stemming from the numerous repetitive similarity calculations caused by its sparse nature.
To address the limitation, we propose a more flexible and faster AM method, named {DMESA} (\cref{fig:dmesa}), which matches semantic areas from SAM in a dense manner and requires no training. 
The core of DMESA involves deriving area matches from patch correspondences, which can be established through a coarse-level matching of an off-the-shelf point matcher~\cite{aspanformer}. 
Particularly, after source area image is obtained from AG, DMESA first establishes patch matches between the area and the target image.
Then a dense matching distribution is generated on the target image (\cref{sec:pcr}) to guide AM, by formulating the patch matches as a Gaussian Mixture Model (GMM). To further migrate the coarse accuracy issue, the cycle consistency-based~\cite{silk} refinement is proposed in \cref{sec:ccr}. This refinement employs finite-step Expectation Maximization (EM) algorithm to ultimately enhance the AM precision.

\subsection{Matching Distribution Generation}\label{sec:pcr}
The key of DMESA is to generate a dense matching distribution to guide AM, leveraging patch matches.
To this end, we first achieve patch matches ($\{\mathsf{p}_k\}_k^K$) between the source area and the target image, along with their \textit{confidences}. 
This can be easily accomplished by utilizing a coarse matching stage of an off-the-shelf point matcher~\cite{aspanformer}.
Then, the patch matches can be treated as mulitiple Gaussian distributions in the target image; the patch centers ($\{u_k,v_k\}_k^K$) are the means ($\{{\bm\mu}_k\}_k^K$) and the match confidence ($\{c_k\}_k^K$) can be used to generate variance ($\{{\bm \Sigma}_k\}_k^K$):
\begin{equation}
    {\bm \mu}_k = (u_k,v_k),~{\bm \Sigma}_k = 
    \begin{bmatrix}
        \frac{w_{\mathsf{p}_k}}{c_k} & 0  \\ 0 & \frac{h_{\mathsf{p}_k}}{c_k}
    \end{bmatrix},
\end{equation}
where $w_{\mathsf{p}_k}=h_{\mathsf{p}_k}=8$ is the size of the patch $\mathsf{p}_k$.
Afterwards, these Gaussian distributions can represent the matching probabilities of 2D locations ($\{{\bm x}_k\!\in\!\mathsf{p}_k\}_k$) inside the patches:
\begin{equation}
\begin{aligned}
        p(\bm{x}_k) & = \frac{1}{2\pi |{\bm \Sigma}_k|^{\frac{1}{2}}} \exp{-\frac{1}{2}(\bm{x}_k-\mu_k)^T{\bm\Sigma}_k^{-1}(\bm{x}_k-\mu_k)} \\
        & =\mathcal{N}_k(\bm{x}_k|\bm{\mu}_k,{\bm \Sigma}_k).
\end{aligned}
\end{equation}
Therefore, we can model the joint matching distribution in the target image as a GMM, by introducing an one-hot $K$-dimensional latent variable ${\bm z}$ and $p(\bm{z})=\Pi_{k}^{K}\pi_k^{z_k}$, where $z_k$ represents the $k$-th entry of the vector $\bm{z}$ and $\pi_k$ is the mixing coefficients~\cite{prml}. 
\begin{equation}\label{eq:patch-prob}
    p(\bm{x})=\sum\limits_{k}^K \pi_{k}~\mathcal{N}(\bm{x}_k|\bm{\mu}_k,{\bm \Sigma}_k).
\end{equation}
This matching distribution can guide the following AM. By setting a specific confidence threshold $T_c$, the potential boundary points of the area ($p(\bm{x})=T_c$) can be obtained from the distribution. The matched area in the target image, thus, can be acquired as the bounding box of these boundary points.

\subsection{Cycle Consistency Refinement}\label{sec:ccr}
Utilizing patch matching is an economical way to obtain area matches, but the inherently coarse nature of patch matches limits their accuracy, subsequently restricting the precision of the resulting area matches. This motivates us to further refine the matching distribution. In particular, we improve the precision of matching distribution by introducing the cycle consistency prior. Cycle consistency, a common constraint in matching~\cite{silk,disk}, asserts that correct matches remain unaffected by the matching direction, which refers to the choice of \textbf{source} and \textbf{target} images. The coarse matching method~\cite{aspanformer} employed in DMESA operates asymmetrically on the input image pairs. It only searches for correspondences in the target image for patches in the source image. Thus, the cycle-consistency prior can be introduced by exchanging the source and target images in this coarse matching. Specially, following the probability form in Sec.~\ref{sec:pcr}, coarse matching achieves the joint distribution $p(\bm{x}^{s\rightarrow t}; \{{\bm\mu}^{ s\rightarrow t}_k\}_k^K,\{{\bm \Sigma}^{s\rightarrow t}_k\}_k^K)$ (Eq.~\ref{eq:patch-prob}) for AM. After modifying the matching direction, another distribution can be obtained: $p(\bm{x}^{t\rightarrow s};\{{\bm\mu}^{t\rightarrow s}_k\}_k^K, \{{\bm \Sigma}^{t\rightarrow s}_k\}_k^K)$. Since the key insight is the consistency between two matching directions, we can enforce the fusion of above two distributions to enhance the consistent matching results. Therefore, we propose a finite-step EM algorithm~\cite{prml} to fuse the two distributions. 

There are two primary elements in the EM algorithm applied to the GMM: the observed data and the initial parameters. In this case, we can sample data from one distribution $p(\bm{x}^{s\rightarrow t})$ as the observation ${\mathsf{\textbf{x}}}^{s\rightarrow t}$, and set parameters from the other distribution ($\{{\bm\mu}^{t\rightarrow s}_k\}_k^K, \{{\bm \Sigma}^{t\rightarrow s}_k\}_k^K$) as the initial parameters $\theta^{t\rightarrow s}_0$. Then, we can use the EM algorithm to update the parameters:
\begin{equation}
    \theta^{t\rightarrow s}_{t+1} = \arg \max_{\theta} \int \log p({\mathsf{\textbf{x}}}^{s\rightarrow t},{\bm z}|\theta^{t\rightarrow s})p({\bm z}|{\mathsf{\textbf{x}}}^{s\rightarrow t},\theta^{t\rightarrow s}_{t})dx,
\end{equation}
where $p(\bm{z})=\Pi_{k}^{K}\pi_k^{z_k}$ and $\{\pi_k\}_k^K$ are initialized as $\pi_k = \frac{1}{K}$. 
After a finite number of update steps ($S_{EM}$), we use the updated parameters $\theta_{S_{EM}}^{t\rightarrow s}$ to generate the refined matching distribution. This distribution has improved consistency in two matching directions, ultimately increase the AM precision. Note the matched patches are established in the source area as well. Thus, we can refine the source area by the above techniques at the same time.

\section{Experiments}\label{sec:exp}
In this section, we comprehensively evaluate our methods on feature matching and its downstream tasks.
Firstly, the implementation details of our methods are presented in \cref{sec:id}.
Then, experiment results on the area and point matching tasks are reported respectively in \cref{sec:exp-am} and \cref{sec:exp-pm}. 
Extensive pose estimation experiments (\cref{sec:exp-pe}) are also conducted on various datasets to prove the efficacy of our methods. 
Additionally, we perform visual odometry (\cref{sec:exp-vo}) to showcase the performance of our methods in driving scenes. 
The impact of input resolution and PM model fine-tuning is experimentally investigated in \cref{sec:ab-irmf}.
Further ablation studies about MESA, DMESA and SAM2~\cite{sam2}
 are provided in \cref{sec:exp-ab}.
Cross-domain experimental results of our methods are reported in \cref{sec:ab-cd} of the appendix as well.

\begin{table}[t]
\caption{\textbf{Area matching results.} We compare the area matching performance between SGAM, MESA and DMESA, combined with GAM~\cite{sgam} under various $\phi$ settings. Results of each series are highlighted as \colorbox{colorFst}{\bf best}, \colorbox{colorSnd}{second} and \colorbox{colorTrd}{third} respectively.}\label{tab:AM}
\resizebox{\linewidth}{!}{
\begin{threeparttable}
\begin{tabular}{llcccc}
\toprule
\multicolumn{2}{c}{Method} & AOR $\uparrow$    & AMP@$0.6$ $\uparrow$ & ACR $\uparrow$ & Pose AUC@$5$ $\uparrow$
 $\uparrow$ \\ \cmidrule(r){1-2} \cmidrule(r){3-4}\cmidrule(r){5-5}\cmidrule(r){6-6}
\multicolumn{2}{c}{SGAM\tnote{$\dagger$}~\cite{sgam}+SP+SG~\cite{superglue}} &  50.37 &  46.76  &  \fs 80.45 & 19.15    \\  	 
\multicolumn{2}{c}{\small w/ \textit{GAM} ($\phi=0.5$)}  &  \rd 54.87 &  \rd 50.22  &  67.54 & \rd 19.32  \\  
\multicolumn{2}{c}{\small w/ \textit{GAM} ($\phi=1.0$)}  &  \fs 60.36 &  \fs 53.47  &  \rd 71.31 & \fs 20.54  \\  
\multicolumn{2}{c}{\small w/ \textit{GAM} ($\phi=3.5$)}  &  \nd 59.74 &  \nd 52.31  &  \nd 73.42 & \nd 20.27  \\   \hdashline \noalign{\vskip 1pt}
\multicolumn{2}{c}{\textbf{MESA+SP+SG}}                   & \rd  65.12 &  \rd 77.89  & \fs 94.93 & \rd 20.33  \\  
\multicolumn{2}{c}{\small w/ \textit{GAM} ($\phi=0.5$)}  &  62.34 &  75.56  &  71.65 & 19.97  \\  
\multicolumn{2}{c}{\small w/ \textit{GAM} ($\phi=1.0$)}  &  \nd 67.45 &  \nd 80.24  &  \rd 85.22 & \nd 21.22  \\ 
\multicolumn{2}{c}{\small w/ \textit{GAM} ($\phi=3.5$)}  &  \fs 68.44 &  \fs 83.25  &  \nd 94.57 & \fs 22.72  \\    \hdashline \noalign{\vskip 1pt}
\multicolumn{2}{c}{\textbf{DMESA+SP+SG}}                 &  71.36 &   82.56  &  \fs 85.52 & 19.97  \\  
\multicolumn{2}{c}{\small w/ \textit{GAM} ($\phi=0.5$)}  & \rd 72.46 &  \rd 84.33  &  64.50 & \rd 20.14  \\  
\multicolumn{2}{c}{\small w/ \textit{GAM} ($\phi=1.0$)}  &  \nd 75.33 &  \nd 85.46  &  \rd 69.08 & \nd 21.23  \\  
\multicolumn{2}{c}{\small w/ \textit{GAM} ($\phi=3.5$)}  & \fs 78.13 &  \fs 86.45  &  \nd 79.44 & \fs 22.19  \\
 \bottomrule
\end{tabular}
\begin{tablenotes}
     \item[$\dagger$] {\small SGAM with only semantic area matching activated.}
\end{tablenotes}
\end{threeparttable}
}
\end{table}

\subsection{Implementation Details}\label{sec:id}
This section describes the implementation details of our method in three aspects, corresponding to three key elements of the A2PM framework: AM, PM and the integration of these two parts.

\subsubsection{AM Details}\label{sec:am-d}
The AM phase contains the proposed MESA and DMESA. We offer the \textbf{parameter settings} of both methods here, along with the \textbf{training details} of the learning area similarity model in MESA.

\boldparagraph{Parameter settings:} 
\textbf{For MESA}, the common parameters for different scenes are set as follows.
In AG construction, the input images are resized to $640\! \times \! 480$.
The aspect ratio threshold $T_r\!=\!4$ and minimal size threshold is $T_s\!=\!80^2$.
The number of area size threshold is $4$ and specific $TL_i$s are $\mathsf{80^2, 130^2, 256^2, 390^2, 560^2}$.
The $\delta_l$ is $0.1$ and $\delta_h$ is $0.8$.
In graphical area matching, the $\lambda$ in \cref{eq:e} is $0.1$.
The area similarity threshold $T_{as}\!=\!0.05$.
The energy balance weights ($\mu, \alpha, \beta, \gamma$) in \cref{eq:eg} are $\mathsf{4,2,2,2}$.  The specific area level $l_{a^*}$ for point matching is $1$. The $T_{Er}$ in \cref{eq:collectarea} is $0.1$.
Other parameters specified for different scenes are described in experiment sections. Ablation study about the parameter settings of MESA can be found in \cref{sec:ab-ep} of the appendix.
\textbf{For DMESA}, the confidence threshold is empirically set as $T_c\!=\!{e^{-1}}/{2\pi}$ and the step number of EM is $S_{EM}\!=\!1$.

\boldparagraph{Training details:} 
We propose the learning model for area similarity calculation in MESA, whose training protocol is described as follows.
Due to the classification formulation, we use the binary cross entropy~\cite{ce} of each patch classification to form the loss function of area similarity calculation. Regular area images are generated using AG from both indoor and outdoor datasets~\cite{scannet,megadepth} as the training data. We train the indoor and outdoor models respectively on 2 NVIDIA RTX 4090 GPUs using AdamW~\cite{adamw}.

\subsubsection{PM Details}\label{sec:pm-d}
As described in \cref{sec:a2pm-c}, model fine-tuning is able to increase matching accuracy in specific input sizes. However, this process incurs additional training costs. Therefore, to demonstrate the efficacy of our methods in a practical manner, we utilize \textbf{the original models} of point matchers provided by their authors in the following experiments, unless explicitly stated otherwise. 

\subsubsection{A2PM Details}\label{sec:a2pm-d}
The A2PM framework is responsible for the integration between AM and PM, which includes the \textbf{area image cropping} (from AM results to PM input) and \textbf{match fusion} (from inside-area PM results to final matching results).

\boldparagraph{Area image cropping:} 
As described in \cref{sec:a2pm-c}, we crop areas with a specified aspect ratio ($r_a\!:=\!W/H$, usually set as $1$) by area expansion. We adjust the area size to possess the required aspect ratio, while trying to keep the area center unchanged. Specifically, if the original area respect ratio ($W_a/H_a$) is larger than $r_a$, we fix the width $W_a$ of area image and expand the height $H_a$ to $W_a/r_a$. Otherwise, we fix the $H_a$ and expand the $W_a$ to $H_a\times r_a$. If the expanded area exceeds the original image, we will move its center to keep it inside the image. This cropping operation is experimentally confirmed in \cref{sec:ab-ic} of the appendix.

\boldparagraph{Match fusion:} 
The final matches can be obtained by merging the inside-area point matches.
Instead of naive fusion, we adopt \textit{Geometric Area Matching} (GAM)~\cite{sgam} to enhance the matching precision utilizing geometry consistency. Additionally, we also adopt the \textit{Global Match Collection}~\cite{sgam} and set the occupancy ratio as $0.6$, which achieves global point matches, guided by inside-area matches, to avoid matching aggregation issue.

\begin{table*}[!t]
\caption{\textbf{Point Matching on ScanNet1500.} Relative gains are highlighted as subscripts. The \colorbox{colorFst}{\bf best}, \colorbox{colorSnd}{second} and \colorbox{colorTrd}{third} results are highlighted.}\label{tab:SNMMA}
\resizebox{\linewidth}{!}{
\begin{threeparttable}
\begin{tabular}{llllllllllll}
\toprule
 \multicolumn{3}{c}{\multirow{2}{*}{Image Matching}} & \multicolumn{3}{c}{$640 \times 640$} & \multicolumn{3}{c}{$640 \times 480$\tnote{$\dagger$}} & \multicolumn{3}{c}{$480 \times 480$}  \\ \cmidrule(l){4-6} \cmidrule(l){7-9} \cmidrule(l){10-12} 
\multicolumn{3}{c}{}  & MMA@3$\uparrow$       & MMA@5$\uparrow$      & MMA@7$\uparrow$      & MMA@3$\uparrow$       & MMA@5$\uparrow$      & MMA@7$\uparrow$  & MMA@3$\uparrow$       & MMA@5$\uparrow$      & MMA@7$\uparrow$       \\ \midrule
\multirow{4}{*}{\rotatebox{90}{Sparse}} & \multicolumn{2}{l}{SP~\cite{superpoint}+SG~\cite{superglue}}                    & 20.50 & 38.22 & 51.84 & 20.61 & 38.46 & 51.82 & 20.53 & 38.25 & 51.56   \\
 &\multicolumn{2}{l}{SGAM~\cite{sgam}+SP+SG}               & \rd21.37{\good$_{+4.24\%}$} & \rd 40.85{\good$_{+6.88\%}$} & \rd53.61{\good$_{+3.41\%}$} & \rd21.75{\good$_{+5.53\%}$} & \rd40.23{\good$_{+4.60\%}$} & \rd52.81{\good$_{+1.91\%}$} & \rd22.71{\good$_{+10.62\%}$} & \rd40.45{\good$_{+5.75\%}$} & \rd52.21{\good$_{+1.26\%}$}  \\ \dashlineours{3}{12}
 &\multicolumn{2}{l}{\ours MESA+SP+SG}               & \fs24.62{\good$_{+20.10\%}$} & \fs43.18{\good$_{+12.98\%}$} & \fs56.29{\good$_{+8.58\%}$} & \fs25.79{\good$_{+25.13\%}$} & \fs44.86{\good$_{+16.64\%}$} & \fs57.81{\good$_{+11.56\%}$} & \fs25.34{\good$_{+23.43\%}$} & \fs44.02{\good$_{+15.08\%}$} & \fs56.87{\good$_{+10.30\%}$}  \\
 &  \multicolumn{2}{l}{\ours DMESA+SP+SG}              & \nd22.89{\good$_{+11.66\%}$} & \nd41.12{\good$_{+7.59\%}$} & \nd54.29{\good$_{+4.73\%}$} & \nd22.89{\good$_{+11.06\%}$} & \nd41.13{\good$_{+6.94\%}$} & \nd54.17{\good$_{+4.53\%}$} & \nd23.46{\good$_{+14.27\%}$} & \nd41.96{\good$_{+9.70\%}$} & \nd55.03{\good$_{+6.73\%}$}  \\ \cmidrule(l){1-12} 
\multirow{12}{*}{\rotatebox{90}{Semi-Dense}} &\multicolumn{2}{l}{ASpan~\cite{aspanformer}}                   & 25.13 & 47.02 & 62.34 & \fs 27.50 & \fs 49.13 & \fs 63.65 & 18.97 & 37.80 & 52.94   \\ 
 &  \multicolumn{2}{l}{SGAM+ASpan}  & \rd25.59{\good$_{+1.83\%}$} & \rd47.64{\good$_{+1.32\%}$} & \rd62.75{\good$_{+0.66\%}$} & 24.51{\bad$_{-10.87\%}$} & 45.95{\bad$_{-6.47\%}$} & \rd62.27{\bad$_{-2.17\%}$} & \rd20.97{\good$_{+10.54\%}$} & \rd38.18{\good$_{+1.01\%}$} & \rd53.19{\good$_{+0.47\%}$}  \\ \dashlineours{3}{12}
 &  \multicolumn{2}{l}{\ours MESA+ASpan}                & \nd26.20{\good$_{+4.26\%}$} & \nd48.94{\good$_{+4.08\%}$} & \nd63.88{\good$_{+2.47\%}$} & \rd25.60{\bad$_{-6.91\%}$} & \rd46.82{\bad$_{-4.70\%}$} & 61.63{\bad$_{-3.17\%}$} & \nd22.19{\good$_{+16.97\%}$} & \nd42.17{\good$_{+11.56\%}$} & \nd57.14{\good$_{+7.93\%}$}   \\
 &  \multicolumn{2}{l}{\ours DMESA+ASpan}              & \fs28.78{\good$_{+14.52\%}$} & \fs51.06{\good$_{+8.59\%}$} & \fs65.45{\good$_{+4.99\%}$} & \nd26.65{\bad$_{-3.09\%}$} & \nd48.47{\bad$_{-1.34\%}$} & \nd62.99{\bad$_{-1.04\%}$} & \fs25.76{\good$_{+35.79\%}$} & \fs46.71{\good$_{+23.57\%}$} & \fs60.97{\good$_{+15.17\%}$}  \\ \cmidrule(l){3-12}
 & \multicolumn{2}{l}{QT~\cite{quadtree}}                      & 22.85 & 41.78 & 53.43 & 29.87 & \rd 52.78 & \rd 67.64 & 24.56 & 45.91 & 61.22   \\
& \multicolumn{2}{l}{SGAM+QT}  & \rd23.35{\good$_{+2.19\%}$} & \rd42.13{\good$_{+0.84\%}$} & \rd55.32{\good$_{+3.54\%}$} & \rd30.14{\good$_{+0.90\%}$} & 52.41{\bad$_{-0.70\%}$} & 66.38{\bad$_{-1.86\%}$} & \rd25.54{\good$_{+3.99\%}$} & \rd46.23{\good$_{+0.70\%}$} & \rd62.45{\good$_{+2.01\%}$}  \\ \dashlineours{3}{12}
&  \multicolumn{2}{l}{\ours MESA+QT}                   & \fs29.32{\good$_{+28.32\%}$} & \fs48.41{\good$_{+15.87\%}$} & \fs60.34{\good$_{+12.93\%}$} & \fs31.25{\good$_{+4.62\%}$} & \fs54.73{\good$_{+3.69\%}$} & \fs69.15{\good$_{+2.23\%}$} & \nd26.93{\good$_{+9.65\%}$} & \nd48.56{\good$_{+5.77\%}$} & \nd63.79{\good$_{+4.20\%}$}  \\
 &  \multicolumn{2}{l}{\ours DMESA+QT}                & \nd24.47{\good$_{+7.09\%}$} & \nd43.72{\good$_{+4.64\%}$} & \nd55.44{\good$_{+3.76\%}$} & \nd30.39{\good$_{+1.74\%}$} & \nd53.47{\good$_{+1.31\%}$} & \nd67.94{\good$_{+0.44\%}$} & \fs28.72{\good$_{+16.94\%}$} & \fs50.70{\good$_{+10.43\%}$} & \fs65.20{\good$_{+6.50\%}$}  \\ \cmidrule(l){3-12}
 & \multicolumn{2}{l}{LoFTR~\cite{loftr}}                   & 26.47 & 48.99 & 63.75 & \nd 28.18 & \nd 50.68 & \nd 65.43 & 20.08 & 40.22 & 55.86   \\
 &  \multicolumn{2}{l}{SGAM+LoFTR}   & \rd27.15{\good$_{+2.57\%}$} & \rd49.53{\good$_{+1.10\%}$} & \rd65.52{\good$_{+2.78\%}$} & 26.22{\bad$_{-6.96\%}$} & 49.13{\bad$_{-3.06\%}$} & 64.73{\bad$_{-1.07\%}$} & \rd21.41{\good$_{+6.62\%}$} & \rd42.03{\good$_{+4.50\%}$} & \rd56.73{\good$_{+1.56\%}$}  \\ \dashlineours{3}{12}
 &  \multicolumn{2}{l}{\ours MESA+LoFTR}                & \fs29.97{\good$_{+13.22\%}$} & \fs52.13{\good$_{+6.41\%}$} & \fs66.64{\good$_{+4.53\%}$} & \rd27.12{\bad$_{-3.76\%}$} & \rd49.63{\bad$_{-2.07\%}$} & \rd64.99{\bad$_{-0.67\%}$} & \nd22.55{\good$_{+12.30\%}$} & \nd43.43{\good$_{+7.98\%}$} & \nd58.64{\good$_{+4.98\%}$}  \\
 &  \multicolumn{2}{l}{\ours DMESA+LoFTR}               & \nd29.86{\good$_{+12.81\%}$} & \nd51.94{\good$_{+6.02\%}$} & \nd65.77{\good$_{+3.17\%}$} & \fs30.29{\good$_{+7.49\%}$} & \fs52.75{\good$_{+4.08\%}$} & \fs66.67{\good$_{+1.90\%}$} & \fs27.07{\good$_{+34.81\%}$} & \fs48.48{\good$_{+20.54\%}$} & \fs62.63{\good$_{+12.12\%}$}  \\ \cmidrule(l){1-12}
\multirow{4}{*}{\rotatebox{90}{Dense}} & \multicolumn{2}{l}{DKM~\cite{dkm}}                    & 26.15 & 45.92 & 59.12 & 26.70 & 46.82 & 60.16 & 26.28 & 46.31 & 59.61   \\
&  \multicolumn{2}{l}{SGAM+DKM} & \rd27.65{\good$_{+5.74\%}$} & \rd46.58{\good$_{+1.44\%}$} & \rd60.88{\good$_{+2.98\%}$} & \rd27.12{\good$_{+1.57\%}$} & \rd47.11{\good$_{+0.62\%}$} & \rd62.21{\good$_{+3.41\%}$} & \rd27.25{\good$_{+3.69\%}$} & \rd47.62{\good$_{+2.83\%}$} & \rd60.34{\good$_{+1.22\%}$}  \\ \dashlineours{3}{12}
&  \multicolumn{2}{l}{\ours MESA+DKM}                & \fs30.15{\good$_{+15.30\%}$} & \fs50.21{\good$_{+9.34\%}$} & \fs64.42{\good$_{+8.96\%}$} & \fs29.67{\good$_{+11.12\%}$} & \fs50.69{\good$_{+8.27\%}$} & \fs64.01{\good$_{+6.40\%}$} & \nd27.87{\good$_{+6.05\%}$} & \nd47.85{\good$_{+3.33\%}$} & \nd60.42{\good$_{+1.36\%}$}  \\
&  \multicolumn{2}{l}{\ours DMESA+DKM}               & \nd28.30{\good$_{+8.22\%}$} & \nd48.81{\good$_{+6.29\%}$} & \nd62.08{\good$_{+5.01\%}$} & \nd28.51{\good$_{+6.78\%}$} & \nd49.26{\good$_{+5.21\%}$} & \nd62.77{\good$_{+4.34\%}$} & \fs28.66{\good$_{+9.06\%}$} & \fs49.52{\good$_{+6.93\%}$} &\fs63.04{\good$_{+5.75\%}$}  \\  \bottomrule
\end{tabular}
\begin{tablenotes}
     \item[$\dagger$] {\small The training size.}
\end{tablenotes}
\end{threeparttable}
}

 \vspace{-1.2em}
\end{table*}

\subsection{Area Matching}\label{sec:exp-am}
Since accurate area matching is the prerequisite for the precise feature matching, we first evaluate our methods for this task on ScanNet1500~\cite{scannet} benchmark.

\subsubsection{Experimental setup} 
We compare the area matching precision between our methods (\ie MESA and DMESA) and semantic-based SGAM~\cite{sgam}. The $T_{E_{max}}$ in MESA is $0.35$. The area size is $480\times 480$. The settings of GAM parameter $\phi$ in all methods are also investigated, which reflects the strictness of outlier rejection. We employ the \textit{Area Overlap Ratio} (AOR, \%) and \textit{Area Matching Precision} (AMP@$0.6$, \%)~\cite{sgam} as metrics. 
The AOR utilizes re-projected Intersection of Union (IoU) of matched areas to measure the AM accuracy. The AMP$0.6$ is the proportion of correct area matches, taking $AOR>0.6$ as the threshold of correct AM.
 Moreover, we propose the \textit{Area Cover Ratio} (ACR, \%), which is the coverage of the all matched areas on the entire image, to measure the completeness of AM. To reveal the relation between AM and the subsequent geometry task, we combine the sparse point matcher SP+SG~\cite{superglue} with the above AM methods, to evaluate the pose estimation accuracy using Pose AUC@$5$~\cite{loftr}, which is the area under the cumulative error curve (AUC) of the pose error at the threshold of $5^{\circ}$. 

\subsubsection{Results} 
The results in \cref{tab:AM} show that our methods outperform SGAM by a large gap, \eg, $60.36$ AOR for SGAM vs. $68.44$ for MESA and $78.33$ for DMESA. DMESA achieves better area matching precision compared to MESA. However, MESA exhibits a higher ACR ($94.93$ vs. $85.52$), possibly due to its dense area comparison, which, although resource-intensive, leads to a greater number of area matches. The improved coverage of area matches enhances the validity of point matches, crucial for subsequent geometric tasks, ultimately resulting in MESA achieving the highest pose estimation accuracy. Nevertheless, the precision of DMESA is also comparable. Considering its faster speed and flexibility, it offers a better efficiency/accuracy trade-off. Additionally, GAM settings impact the precision of both area and point matches. Notably, in both MESA and DMESA, optimal performance is attained with the most relaxed geometric constraint ($\phi\!=\!3.5$), indicating the high accuracy of most area matches obtained by our methods.

\subsection{Point Matching}\label{sec:exp-pm}
Point matching accuracy is a direct reflection of feature matching performance, which is evaluated on ScanNet1500 as well.

\subsubsection{Experimental setup} 
To showcase the versatility and effectiveness of our methods, we incorporate five PM baselines, containing all existing three matching categories, into the A2PM framework as the PM module. These include a widely-used sparse matcher: SP~\cite{superpoint}+SG~\cite{superglue}; three SOTA semi-dense matchers: ASpan~\cite{aspanformer}, QT~\cite{quadtree}, LoFTR~\cite{loftr}; and a leading dense matcher: DKM~\cite{dkm}. For the AM part, we compare MESA, DMESA, and SGAM~\cite{sgam}. The Mean Matching Accuracy (MMA@$3/5/7$)~\cite{d2net} is used to measure the precision. 

As described in \cref{sec:a2pm-c}, we adopt three PM input resolutions for impact investigation, including a small size ($480\times 480$), a middle size ($640\times 480$, also the training size of baselines) and a large size ($640\times 640$). The smaller one leads to less computational cost, while the larger one possesses more details, ideally resulting in better performance. Note the aspect ratio conflict exists in this dataset (the training aspect ratio $\neq\!1$). Thus we also evaluate the performance under the training size of $640\times 480$. The choice of PM input resolution influences both computational requirements and matching accuracy, striking a balance between efficiency and precision in practice. By considering these resolutions, we aim to comprehensively assess the practical value of our approaches.

\begin{table*}[!t]
\caption{\textbf{Pose Estimation on ScanNet1500.} Relative gains are represented as subscripts. The \colorbox{colorFst}{\bf best}, \colorbox{colorSnd}{second} and \colorbox{colorTrd}{third} results are highlighted.}\label{tab:SN}
\resizebox{\linewidth}{!}{
\begin{threeparttable}
\begin{tabular}{llllllllllll}
\toprule
 \multicolumn{3}{c}{\multirow{2}{*}{Pose estimation AUC}} & \multicolumn{3}{c}{$640 \times 640$} & \multicolumn{3}{c}{$640 \times 480$\tnote{$\dagger$}} & \multicolumn{3}{c}{$480 \times 480$}  \\ \cmidrule(l){4-6} \cmidrule(l){7-9} \cmidrule(l){10-12} 
\multicolumn{3}{c}{}  & AUC@5$\uparrow$       & AUC@10$\uparrow$      & AUC@20$\uparrow$      & AUC@5$\uparrow$       & AUC@10$\uparrow$      & AUC@20$\uparrow$  & AUC@5$\uparrow$       & AUC@10$\uparrow$      & AUC@20$\uparrow$       \\ \midrule
\multirow{4}{*}{\rotatebox{90}{Sparse}} & \multicolumn{2}{l}{SP~\cite{superpoint}+SG~\cite{superglue}}                    & 20.22 & 39.62 & 57.80 & 20.20 & 38.87 & 56.86 & 19.27 & 38.06 & 56.26   \\
 &\multicolumn{2}{l}{SGAM~\cite{sgam}+SP+SG}               & \rd21.42{\good$_{+5.93\%}$} & \rd40.61{\good$_{+2.50\%}$} & \rd58.34{\good$_{+0.93\%}$} & \nd21.97{\good$_{+8.76\%}$} & \rd39.94{\good$_{+2.75\%}$} & \rd57.91{\good$_{+1.85\%}$} & \rd20.54{\good$_{+6.59\%}$} & \rd38.87{\good$_{+2.13\%}$} & \rd57.48{\good$_{+2.17\%}$}  \\ \dashlineours{3}{12}
 &\multicolumn{2}{l}{\ours MESA+SP+SG}               & \fs23.42{\good$_{+15.83\%}$} & \fs42.79{\good$_{+8.00\%}$} & \fs61.49{\good$_{+6.38\%}$} & \fs23.24{\good$_{+15.05\%}$} & \fs42.35{\good$_{+8.95\%}$} & \fs60.04{\good$_{+5.59\%}$} & \fs22.72{\good$_{+17.90\%}$} & \fs42.25{\good$_{+11.01\%}$} & \fs59.51{\good$_{+5.78\%}$}  \\
 &  \multicolumn{2}{l}{\ours DMESA+SP+SG}              & \nd22.60{\good$_{+11.77\%}$} & \nd41.31{\good$_{+4.27\%}$} & \nd59.07{\good$_{+2.20\%}$} & \nd21.97{\good$_{+8.76\%}$} & \nd40.88{\good$_{+5.17\%}$} & \nd58.71{\good$_{+3.25\%}$} & \nd22.19{\good$_{+15.15\%}$} & \nd41.25{\good$_{+8.38\%}$} & \nd58.79{\good$_{+4.50\%}$}  \\ \cmidrule(l){1-12} 
\multirow{12}{*}{\rotatebox{90}{Semi-Dense}} &\multicolumn{2}{l}{ASpan~\cite{aspanformer}}                    & 24.48 & 43.64 & 60.38 & \fs 28.37 & \nd 49.24 & \nd 66.44 & 22.43 & 41.67 & 60.26    \\
 &  \multicolumn{2}{l}{SGAM+ASpan}  & \rd25.13{\good$_{+2.66\%}$} & \rd44.27{\good$_{+1.44\%}$} & \rd60.98{\good$_{+0.99\%}$} & 26.14{\bad$_{-7.86\%}$} & \rd46.85{\bad$_{-4.85\%}$} & 62.72{\bad$_{-5.60\%}$} & \rd23.78{\good$_{+6.02\%}$} & \rd42.25{\good$_{+1.39\%}$} & \rd60.93{\good$_{+1.11\%}$} \\ \dashlineours{3}{12}
 &  \multicolumn{2}{l}{\ours MESA+ASpan}               & \fs25.87{\good$_{+5.68\%}$} & \fs46.43{\good$_{+6.39\%}$} & \fs62.47{\good$_{+3.46\%}$} & \nd 28.23{\bad$_{-0.49\%}$} & \fs49.33{\good$_{+0.18\%}$} & \fs67.04{\good$_{+0.90\%}$} & \fs24.56{\good$_{+9.50\%}$} & \fs44.37{\good$_{+6.48\%}$} & \nd 61.29{\good$_{+1.71\%}$}  \\
 &  \multicolumn{2}{l}{\ours DMESA+ASpan}              & \nd25.75{\good$_{+5.19\%}$} & \nd45.19{\good$_{+3.55\%}$} & \nd62.18{\good$_{+2.98\%}$} & \rd26.36{\bad$_{-7.08\%}$} & 46.60{\bad$_{-5.36\%}$} & \rd63.92{\bad$_{-3.79\%}$} & \nd24.25{\good$_{+8.11\%}$} & \nd44.07{\good$_{+5.76\%}$} & \fs 62.20{\good$_{+3.22\%}$}   \\ \cmidrule(l){2-12} 
 & \multicolumn{2}{l}{QT~\cite{quadtree}}                       & 22.40 & 40.10 & 56.90 & \nd 28.56 & \fs 49.30 & \nd 65.78 & 21.56 & 40.95 & 57.93  \\
& \multicolumn{2}{l}{SGAM+QT}  & \nd23.71{\good$_{+5.85\%}$} & \rd41.55{\good$_{+3.62\%}$} & \rd56.13{\bad$_{-1.35\%}$} &  26.25{\bad$_{-8.09\%}$} &  44.63{\bad$_{-9.47\%}$} &  62.73{\bad$_{-4.64\%}$} & \rd22.79{\good$_{+5.71\%}$} & \rd42.04{\good$_{+2.66\%}$} & \rd58.20{\good$_{+0.47\%}$} \\ \dashlineours{3}{12}
&  \multicolumn{2}{l}{\ours MESA+QT}                  & \fs24.12{\good$_{+7.68\%}$} & \fs43.03{\good$_{+7.31\%}$} & \fs60.13{\good$_{+5.68\%}$} & \fs28.74{\good$_{+0.63\%}$} & \nd49.12{\bad$_{-0.37\%}$} & \fs66.03{\good $_{+0.38\%}$} & \fs24.72{\good$_{+14.66\%}$} & \fs43.57{\good$_{+6.40\%}$} & \fs60.41{\good$_{+4.28\%}$}   \\
 &  \multicolumn{2}{l}{\ours DMESA+QT}                 & \rd23.41{\good$_{+4.51\%}$} & \nd41.70{\good$_{+3.99\%}$} & \nd59.14{\good$_{+3.94\%}$} & \rd26.51{\bad$_{-7.18\%}$} & \rd46.71{\bad$_{-5.25\%}$} & \rd63.41{\bad$_{-3.60\%}$} & \nd23.57{\good$_{+9.32\%}$} & \nd43.00{\good$_{+5.01\%}$} & \nd59.98{\good$_{+3.54\%}$}  \\ \cmidrule(l){2-12} 
 & \multicolumn{2}{l}{LoFTR~\cite{loftr}}                    & 21.61 & 40.03 & 55.82 & \nd 25.68 & \nd 45.86 & \nd 62.60 & 20.36 & 39.44 & 57.16  \\
 &  \multicolumn{2}{l}{SGAM+LoFTR}  & \rd22.05{\good$_{+2.04\%}$} & \rd40.11{\good$_{+0.20\%}$} & \nd56.65{\good$_{+1.49\%}$} & 23.82{\bad$_{-7.24\%}$} & 44.19{\bad$_{-3.64\%}$} & \rd61.51{\bad$_{-1.74\%}$} & \rd21.94{\good$_{+7.76\%}$} & \rd40.42{\good$_{+2.48\%}$} & \rd57.42{\good$_{+0.45\%}$} \\ \dashlineours{3}{12}
 &  \multicolumn{2}{l}{\ours MESA+LoFTR}               & \fs23.41{\good$_{+8.33\%}$} & \fs42.68{\good$_{+6.62\%}$} & \fs57.68{\good$_{+3.33\%}$} & \fs26.23{\good$_{+2.14\%}$} & \fs46.06{\good$_{+0.44\%}$} & \fs62.90{\good$_{+0.48\%}$} & \fs22.35{\good$_{+9.77\%}$} & \fs42.04{\good$_{+6.59\%}$} & \fs58.34{\good$_{+2.06\%}$}   \\
 &  \multicolumn{2}{l}{\ours DMESA+LoFTR}              & \nd22.57{\good $_{+4.44\%}$} & \nd40.67{\good $_{+1.60\%}$} & \rd56.22{\good $_{+0.72\%}$} & \rd24.37{\bad $_{-5.10\%}$} & \rd44.42{\bad $_{-3.14\%}$} &  61.34{\bad $_{-2.01\%}$} & \nd21.99{\good $_{+8.01\%}$} & \nd40.53{\good $_{+2.76\%}$} & \nd57.52{\good $_{+0.63\%}$}    \\ \cmidrule(l){1-12} 
\multirow{4}{*}{\rotatebox{90}{Dense}} & \multicolumn{2}{l}{DKM~\cite{dkm}}                    & 29.20 & 50.96 & 68.55 & 29.76 & 51.65 & 69.39 & 28.55 & 49.97 & 67.82  \\
&  \multicolumn{2}{l}{SGAM+DKM} & \rd29.45{\good$_{+0.86\%}$} & \nd51.74{\good$_{+1.53\%}$} & \rd69.91{\good$_{+1.98\%}$} & \rd30.33{\good$_{+1.92\%}$} & \rd51.96{\good$_{+0.60\%}$} & \nd69.54{\good$_{+0.22\%}$} & \rd29.57{\good$_{+3.57\%}$} & \rd50.86{\good$_{+1.78\%}$} & \rd68.39{\good$_{+0.84\%}$} \\ \dashlineours{3}{12}
&  \multicolumn{2}{l}{\ours MESA+DKM}               & \fs31.84{\good $_{+9.04\%}$} & \fs53.07{\good $_{+4.14\%}$} & \fs70.12{\good $_{+2.29\%}$} & \fs32.14{\good $_{+8.00\%}$} & \fs53.97{\good $_{+4.49\%}$} & \fs71.02{\good $_{+2.35\%}$} & \fs30.12{\good $_{+5.50\%}$} & \nd51.03{\good $_{+2.12\%}$} & \nd68.71{\good $_{+1.31\%}$}   \\
&  \multicolumn{2}{l}{\ours DMESA+DKM}              & \nd29.59{\good $_{+1.34\%}$} & \rd51.21{\good $_{+0.49\%}$} & \nd70.02{\good $_{+2.14\%}$} & \nd30.77{\good $_{+3.39\%}$} & \nd52.19{\good $_{+1.05\%}$} & \rd69.52{\good $_{+0.19\%}$} & \nd29.94{\good $_{+4.87\%}$} & \fs51.35{\good $_{+2.76\%}$} & \fs68.82{\good $_{+1.47\%}$}  \\  \bottomrule
\end{tabular}
\begin{tablenotes}
     \item[$\dagger$] {\small The training size.}
\end{tablenotes}
\end{threeparttable}
}

 \vspace{-1.2em}
\end{table*}

\subsubsection{Results} 
The point matching results are summarized in \cref{tab:SNMMA}. These results are analyzed with the categories of PM as the primary focus.

\textit{For the sparse matcher}, SP+SG, we observe consistent and substantial accuracy improvements achieved by our methods across all input sizes. Our methods surpass SGAM by a large margin. MESA exhibits the best overall performance, with DMESA also delivering impressive results. Particularly, MESA/DMESA+SP+SG gains better results with the small size of $480\!\times\!480$ than the large size of $640\! \times\! 640$ (MESA on MMA@7: $56.87$ vs. $56.29$), achieving higher accuracy with less computational cost, proving its resolution robustness.

\textit{For three semi-dense matchers}, the resolution overfitting is noticeable (cf. \cref{sec:a2pm-c}), as the results with the training size remarkably surpass others. In two square resolutions, our methods gain prominent improvements for all three matchers. In the training size, however, declines in precision are observed for MESA/DMESA+ASpan and MESA+LoFTR. This can partly attribute to that this non-square size leads to excessive area size adjustment, which can introduce matching redundancy into area matches.
However, our methods still improve the results of QT and surpass SGAM. 
DMESA achieves better results than MESA here, due to its higher area matching accuracy (cf. \cref{sec:exp-am}). 

\textit{For the dense matcher}, the overfitting issue is relatively minor, thanks to the robustness against resolution of DKM. Our methods consistently improve the performance across all three input sizes, notably outperforming SGAM. Additionally, DMESA demonstrates superior performance in the small input size.

\subsubsection{Discussion}\label{sec:pm-diss}
\begin{figure*}[!t]
\centering
\includegraphics[width=\linewidth]{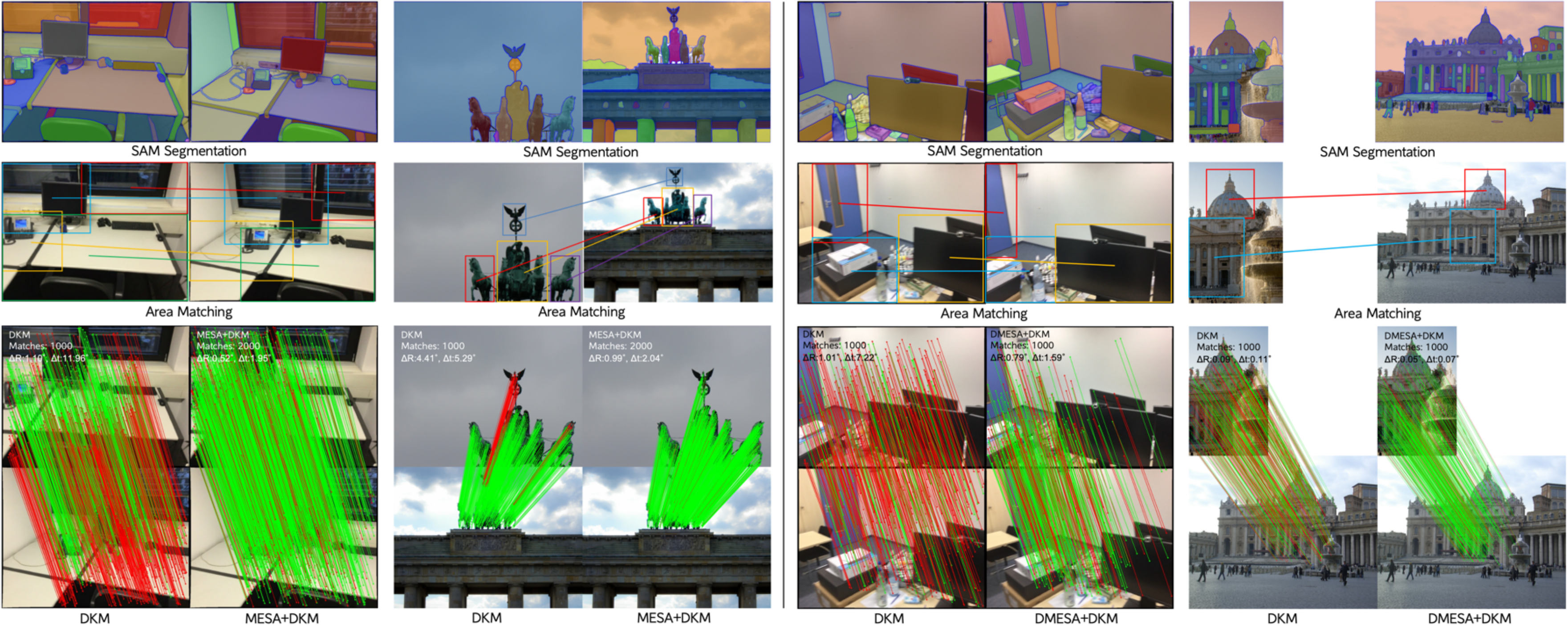}
\caption{\textbf{The qualitative results of our methods.} We provide qualitative results of MESA and DMESA on ScanNet1500 and MegaDepth1500. Our methods significantly improve the point matching and pose estimation performance of DKM, by attaining precise area matches.}
 \vspace{-1.2em}
\label{fig:qbmd}
\end{figure*}

When comparing between \cref{tab:AM} and \cref{tab:SNMMA}, counterintuitive results can be observed.
MESA generally achieves better PM precision than DMESA, but DMESA obtains better AOR in \cref{tab:AM}. 
This conflict can be attributed to that MESA can obtain more areas with finer granularity (better ACR of MESA and see \cref{fig:qbmd}) by its multi-level area fusion (cf. \cref{Sec:refine}) in this dataset. This leads to more precise PM from more detailed feature comparison within areas.
On the other hand, the AM accuracy advantage of DMESA is revealed and discussed in \cref{sec:pe-diss}.


Comparing results across the baselines, our methods combined with DKM achieves the best results. 
The sparse matcher enhanced by our methods achieves precision comparable to its semi-dense counterparts, at the small resolution of $480\!\times\!480$ (\eg, MESA+SP+SG on MMA@7: $56.87$ vs. LoFTR: $55.86$).
This underscores the practical significance of our methods, particularly with \textit{constrained computational resources}, as the sparse method is much more efficient than the semi-dense one.

\begin{table}[!t]
\caption{\textbf{Pose Estimation on ETH3D.} Relative gains are represented as subscripts. The \colorbox{colorFst}{\bf best}, \colorbox{colorSnd}{second} and \colorbox{colorTrd}{third} results are highlighted.}\label{tab:ETH3D}
\resizebox{\linewidth}{!}{
\begin{tabular}{llllllll}
\toprule
\multicolumn{4}{c}{{Pose estimation AUC}} &  AUC@5$\uparrow$       & AUC@10$\uparrow$      & AUC@20$\uparrow$          \\ \midrule
\multirow{20}{*}{\rotatebox{90}{$640 \times 640$}} & \multirow{4}{*}{\rotatebox{90}{Sparse}} & \multicolumn{2}{l}{SP~\cite{superpoint}+SG~\cite{superglue}}                    & 19.09 & 32.89 & 46.59    \\		
\multicolumn{2}{c}{} &\multicolumn{2}{l}{SGAM~\cite{sgam}+SP+SG}               & \rd 19.73{\good $_{+3.35\%}$} & \rd 33.14{\good $_{+0.76\%}$} & \nd 47.58{\good $_{+2.12\%}$}
 \\ \dashlineours{3}{8}
\multicolumn{2}{c}{} &\multicolumn{2}{l}{\ours MESA+SP+SG}               & \fs 22.79{\good $_{+19.38\%}$} & \fs 36.43{\good $_{+10.76\%}$} & \fs 48.51{\good $_{+4.12\%}$} 	
 \\
\multicolumn{2}{c}{} &  \multicolumn{2}{l}{\ours DMESA+SP+SG}              & \nd 20.30{\good $_{+6.34\%}$} & \nd 33.46{\good $_{+1.73\%}$} & \rd 46.56{\bad $_{-0.06\%}$} &    \\ \cmidrule(l){2-8} 	
 & \multirow{12}{*}{\rotatebox{90}{Semi-Dense}} &\multicolumn{2}{l}{ASpan~\cite{aspanformer}}                    & 16.92 & 31.06 & 45.5             \\
 \multicolumn{2}{c}{} &\multicolumn{2}{l}{SGAM+ASpan}               & \rd 17.35{\good $_{+2.54\%}$} & \rd 31.88{\good $_{+2.64\%}$} & \rd 46.12{\good $_{+1.36\%}$}
 \\ \dashlineours{3}{8}
\multicolumn{2}{c}{} & \multicolumn{2}{l}{\ours MESA+ASpan}               & \fs 22.31{\good $_{+31.86\%}$} & \fs 35.71{\good $_{+14.97\%}$} & \fs 50.07{\good $_{+10.04\%}$} \\
\multicolumn{2}{c}{} &  \multicolumn{2}{l}{\ours DMESA+ASpan}              & \nd 18.73{\good $_{+10.70\%}$} & \nd 33.24{\good $_{+7.02\%}$} & \nd 47.83{\good $_{+5.12\%}$}   \\ \cmidrule(l){3-8} 
\multicolumn{2}{c}{} &\multicolumn{2}{l}{QT~\cite{quadtree}}                    & 17.41 & 32.35 & 47.39           \\
 \multicolumn{2}{c}{} &\multicolumn{2}{l}{SGAM+QT}               & \rd 17.68{\good $_{+1.55\%}$} & \rd 32.93{\good $_{+1.79\%}$} & \rd 47.86{\good $_{+0.99\%}$}
 \\ 	\dashlineours{3}{8}
\multicolumn{2}{c}{} &  \multicolumn{2}{l}{\ours MESA+QT}               & \fs 21.72{\good $_{+24.76\%}$} & \fs 36.77{\good $_{+13.66\%}$} & \fs 50.23{\good $_{+5.99\%}$}   \\
\multicolumn{2}{c}{} &  \multicolumn{2}{l}{\ours DMESA+QT}              & \nd 19.29{\good $_{+10.80\%}$} & \nd 34.41{\good $_{+6.37\%}$} & \nd 49.10{\good $_{+3.61\%}$} \\ \cmidrule(l){3-8}
\multicolumn{2}{c}{}  &\multicolumn{2}{l}{LoFTR~\cite{loftr}}                    & 15.27 & 28.70 & 42.40        \\
 \multicolumn{2}{c}{} &\multicolumn{2}{l}{SGAM+LoFTR}               & \rd 15.84{\good $_{+3.73\%}$} & \rd 29.34{\good $_{+2.23\%}$} & \rd 42.85{\good $_{+1.06\%}$}
 \\ \dashlineours{3}{8}
\multicolumn{2}{c}{} &  \multicolumn{2}{l}{\ours MESA+LoFTR}               & \fs 19.37{\good $_{+26.85\%}$} & \fs 32.82{\good $_{+14.36\%}$} & \fs 46.71{\good $_{+10.17\%}$}  & 
 \\
\multicolumn{2}{c}{} &  \multicolumn{2}{l}{\ours DMESA+LoFTR}              & \nd 15.99{\good $_{+4.72\%}$} & \nd 29.40{\good $_{+2.44\%}$} & \nd 43.00{\good $_{+1.42\%}$}  \\ \cmidrule(l){2-8}
& \multirow{4}{*}{\rotatebox{90}{Dense}} & \multicolumn{2}{l}{DKM}                    & 38.51 & 52.06 & 63.53    \\
 \multicolumn{2}{c}{} &\multicolumn{2}{l}{SGAM+DKM}               & \rd 37.42{\bad $_{-2.83\%}$} & \rd 51.53{\bad $_{-1.02\%}$} & \rd 63.02{\bad $_{-0.80\%}$}
 \\ 	\dashlineours{3}{8}	
\multicolumn{2}{c}{} &\multicolumn{2}{l}{\ours MESA+DKM}               & \fs 43.47{\good $_{+12.88\%}$} & \fs 55.32{\good $_{+6.26\%}$} & \fs 66.15{\good $_{+4.12\%}$}  \\
\multicolumn{2}{c}{} &  \multicolumn{2}{l}{\ours DMESA+DKM}              & \nd 38.27{\bad $_{-0.62\%}$} & \nd 51.89{\bad $_{-0.33\%}$} & \nd 63.31{\bad $_{-0.35\%}$}   \\ 
\midrule
\multirow{20}{*}{\rotatebox{90}{$480 \times 480$}} & \multirow{4}{*}{\rotatebox{90}{Sparse}} & \multicolumn{2}{l}{SP+SG}                    & 16.59 & 30.41 & 44.21    \\
\multicolumn{2}{c}{} &\multicolumn{2}{l}{SGAM+SP+SG}               & \rd 17.31{\good $_{+4.34\%}$} & \rd 31.33{\good $_{+3.03\%}$} & \rd 44.78{\good $_{+1.29\%}$}
 \\ \dashlineours{3}{8}
\multicolumn{2}{c}{} &\multicolumn{2}{l}{\ours MESA+SP+SG}               & \fs 22.45{\good $_{+35.32\%}$} & \fs 35.68{\good $_{+17.33\%}$} & \fs 48.85{\good $_{+10.50\%}$}
 \\
\multicolumn{2}{c}{} &  \multicolumn{2}{l}{\ours DMESA+SP+SG}              & \nd 18.60{\good $_{+12.12\%}$} & \nd 32.39{\good $_{+6.51\%}$} & \nd 45.94{\good $_{+3.91\%}$}   \\ \cmidrule(l){2-8} 
 & \multirow{12}{*}{\rotatebox{90}{Semi-Dense}} &\multicolumn{2}{l}{ASpan}                    & 8.61 & 18.94 & 32.49  \\
 \multicolumn{2}{c}{} &\multicolumn{2}{l}{SGAM+ASpan}               & \rd 10.13{\good $_{+17.65\%}$} & \rd 19.35{\good $_{+2.16\%}$} & \rd 33.11{\good $_{+1.91\%}$}
 \\ \dashlineours{3}{8} 	 	 
\multicolumn{2}{c}{} &  \multicolumn{2}{l}{\ours MESA+ASpan}               & \fs 15.57{\good $_{+80.84\%}$} & \fs 28.66{\good $_{+51.32\%}$} & \fs 42.74{\good $_{+31.55\%}$} 	 \\
\multicolumn{2}{c}{} &  \multicolumn{2}{l}{\ours DMESA+ASpan}              & \nd 13.91{\good $_{+61.56\%}$} & \nd 26.62{\good $_{+40.55\%}$} & \nd 40.82{\good $_{+25.64\%}$}  \\ \cmidrule(l){3-8} 
\multicolumn{2}{c}{} &\multicolumn{2}{l}{QT}                    & 11.29 & 23.86 & 38.13  \\
 \multicolumn{2}{c}{} & \multicolumn{2}{l}{SGAM+QT}               & \rd 11.35{\good $_{+0.53\%}$} & \rd 24.24{\good $_{+1.59\%}$} & \rd 39.05{\good $_{+2.41\%}$}
 \\ 	\dashlineours{3}{8} 		
\multicolumn{2}{c}{} &  \multicolumn{2}{l}{\ours MESA+QT}               & \fs 19.63{\impressive $_{+73.87\%}$} & \fs 32.33{\impressive $_{+35.50\%}$} & \fs 46.75{\impressive $_{+22.61\%}$} 	  \\
\multicolumn{2}{c}{} &  \multicolumn{2}{l}{\ours DMESA+QT}              & \nd 16.96{\good $_{+50.22\%}$} & \nd 30.97{\good $_{+29.80\%}$} & \nd 45.37{\good $_{+18.99\%}$}  \\ \cmidrule(l){3-8}
\multicolumn{2}{c}{}  &\multicolumn{2}{l}{LoFTR}                    & 9.12 & 19.34 & 32.79   \\
 \multicolumn{2}{c}{} &\multicolumn{2}{l}{SGAM+LoFTR}               & \rd 10.26{\good $_{+12.50\%}$} & \rd 19.97{\good $_{+3.26\%}$} & \rd 33.54{\good $_{+2.29\%}$}
 \\  \dashlineours{3}{8}	 	 	 
\multicolumn{2}{c}{} &  \multicolumn{2}{l}{\ours MESA+LoFTR}               & \fs 15.19{\good $_{+66.56\%}$} & \fs 27.43{\good $_{+41.83\%}$} & \fs 40.22{\good $_{+22.66\%}$} 	
 \\
\multicolumn{2}{c}{} &   \multicolumn{2}{l}{\ours DMESA+LoFTR}              & \nd 12.26{\good $_{+34.43\%}$} & \nd 23.93{\good $_{+23.73\%}$} & \nd 37.51{\good $_{+14.39\%}$}   \\ \cmidrule(l){2-8}
& \multirow{4}{*}{\rotatebox{90}{Dense}} & \multicolumn{2}{l}{DKM}                    & 36.25 & 49.45 & 60.34     \\
 \multicolumn{2}{c}{} &\multicolumn{2}{l}{SGAM+DKM}               & \nd 37.13{\good $_{+2.43\%}$} & \rd 49.85{\good $_{+0.81\%}$} & \rd 60.57{\good $_{+0.38\%}$}
 \\ \dashlineours{3}{8}
\multicolumn{2}{c}{} & \multicolumn{2}{l}{\ours MESA+DKM}               & \fs 39.98{\good $_{+10.29\%}$} & \fs 52.27{\good $_{+5.70\%}$} & \fs 62.74{\good $_{+3.98\%}$} 	 \\
\multicolumn{2}{c}{} &  \multicolumn{2}{l}{\ours DMESA+DKM}              & \rd 36.31{\good $_{+0.17\%}$} & \nd 50.07{\good $_{+1.25\%}$} & \nd 61.57{\good $_{+2.04\%}$}   \\
\bottomrule

\end{tabular}
}
 \vspace{-1.2em}
\end{table}

\subsection{Pose estimation}\label{sec:exp-pe}
Pose estimation between images is a crucial subsequent task of PM and a basic of many applications~\cite{matchsurvey}. Thus, we extensively evaluate the pose estimation precision of our methods here.

\subsubsection{Experimental setup} 
In order to showcase the versatility of our methods, we conduct extensive experiments across four datasets encompassing both indoor and outdoor scenes. Specifically, we utilize two indoor datasets, ScanNet1500 and ETH3D~\cite{eth3d}, as well as two outdoor datasets, MegaDepth1500~\cite{megadepth} and YFCC~\cite{YFCC}. ScanNet1500 and MegaDepth1500 are both in-domain datasets, each comprising 1500 image pairs~\cite{aspanformer}. For ETH3D, we use the first 10 sequences to conduct experiments, with 3K image pairs sampled from them at a rate of $6$, following~\cite{ecotr}. Evaluations on YFCC have been conducted following~\cite{pats}, including 4K outdoor image pairs. For each dataset, we choose a large size and a small size for complete evaluation. Moreover, for the in-domain datasets (ScanNet1500 and MegaDepth1500), we further evaluate on the training resolution to investigate the overfitting issue.

Consistent with our prior experiments, we first select five point matchers as the baselines. 
Additionally, another cutting-edge dense point matcher, RoMa~\cite{roma}, is combined with our methods and evaluated on the ScanNet1500 and MegaDepth1500. Due to its unique training resolution and large computational cost, we only investigate its performance in the training size of $560\times560$.

In the indoor scenes, we include SGAM as a comparison method. In the outdoor scenes, we contrast our methods with another overlap matching technique, OETR~\cite{OETR}. The parameter $T_{E_{max}}$ of MESA is set as $0.35$ for indoor scenes and $0.3$ for outdoor scenes. Other parameters are fixed (cf. \cref{sec:am-d}).

Typically, RANSAC~\cite{ransac} is employed to filter outliers and derive camera poses. However, adjusting RANSAC parameters can be cumbersome across diverse datasets. Therefore, in our experiments, we employ \textit{MAGSAC++}~\cite{magsac++} instead, eliminating the parameter adjustment and enhancing the reproducibility.

To facilitate straightforward result comparisons across various datasets, we adopt a unified evaluation metric, specifically the standard pose estimation AUC~\cite{superglue}, throughout all the experiments.
This metric represents the AUC of the pose error at the thresholds (AUC@$5/10/20$), where the pose error is defined as the maximum of angular error in rotation and translation.

\begin{table*}[!t]
\caption{\textbf{Pose Estimation on YFCC and MegaDepth1500.} The subscripts are relative gains. The \colorbox{colorFst}{\bf best}, \colorbox{colorSnd}{second} and \colorbox{colorTrd}{third} results are highlighted.}\label{tab:YFCC+MD}
\resizebox{\linewidth}{!}{
\begin{threeparttable}
\begin{tabular}{lllllllllllllll}
\toprule
 \multicolumn{3}{c}{\multirow{3}{*}{Pose estimation AUC}} &
 \multicolumn{6}{c}{MegaDepth1500} & \multicolumn{6}{c}{YFCC} \\ \cmidrule(l){4-9} \cmidrule(l){10-15}
 \multicolumn{3}{c}{} & \multicolumn{3}{c}{$832\times832$\tnote{$\dagger$}} & \multicolumn{3}{c}{$480\times 480$} & \multicolumn{3}{c}{$640\times640$} & \multicolumn{3}{c}{$480\times 480$} \\ \cmidrule(l){4-6} \cmidrule(l){7-9} \cmidrule(l){10-12} \cmidrule(l){13-15} 
\multicolumn{3}{c}{}  & AUC@5$\uparrow$      & AUC@10$\uparrow$      & AUC@20$\uparrow$      & AUC@5$\uparrow$     & AUC@10$\uparrow$
& AUC@20$\uparrow$ & AUC@5$\uparrow$      & AUC@10$\uparrow$       & AUC@20$\uparrow$  & AUC@5$\uparrow$      & AUC@10$\uparrow$       & AUC@20$\uparrow$    \\ \midrule
\multirow{4}{*}{\rotatebox{90}{Sparse}} & \multicolumn{2}{l}{SP~\cite{superpoint}+SG~\cite{superglue}}

&51.27 & 67.29 & 79.65 &

48.14 & 63.71 & 76.40  &

42.18 & 62.17 & 77.26 &

36.20 & 56.41 & 72.74
\\

&\multicolumn{2}{l}{OETR~\cite{OETR}+SP+SG} 
& 
\rd 52.47{\good $_{+2.34\%}$} & \rd68.35{\good $_{+1.58\%}$} & \rd80.24{\good $_{+0.74\%}$} &

\rd 51.22{\good $_{+6.40\%}$} & \rd 66.04{\good $_{+3.66\%}$} & \rd 78.43 {\good $_{+2.66\%}$}

& 
\nd 43.76{\good $_{+3.75\%}$} & \nd 64.11{\good $_{+3.12\%}$} & \nd 78.13{\good $_{+1.13\%}$} & 

\rd 36.77{\good $_{+1.57\%}$} & \rd 57.31{\good $_{+1.60\%}$} & \rd 73.52{\good $_{+1.07\%}$}

\\ \dashlineours{2}{15}

&\multicolumn{2}{l}{\ours MESA+SP+SG} 

& 
\nd 56.27{\good $_{+9.75\%}$} & \nd71.35{\good $_{+6.03\%}$} & \nd82.11{\good $_{+3.09\%}$} & 
\nd 55.56{\good $_{+15.41\%}$} & \nd 70.89{\good $_{+11.27\%}$} & \nd 81.78{\good $_{+7.04\%}$}

& 

\fs 45.25{\good $_{+7.28\%}$} & \fs 65.37{\good $_{+5.15\%}$} & \fs 79.78{\good $_{+3.26\%}$} &
\fs 40.22{\good $_{+11.10\%}$} & \fs 59.79{\good $_{+5.99\%}$} & \fs 75.37{\good $_{+3.62\%}$}

\\
&\multicolumn{2}{l}{\ours DMESA+SP+SG} 
& 
\fs 57.69{\good $_{+12.52\%}$} & \fs 72.34{\good $_{+7.50\%}$} & \fs 83.10{\good $_{+4.33\%}$} 

& \fs 58.02{\good $_{+20.52\%}$} & \fs 71.77{\good $_{+12.65\%}$} & \fs 82.49{\good $_{+7.97\%}$}

& 
\rd 42.57{\good $_{+0.92\%}$} & \rd 62.46{\good $_{+0.47\%}$} & \rd 77.47{\good $_{+0.27\%}$} & 

\nd 38.25{\good $_{+5.66\%}$} & \nd 58.12{\good $_{+3.03\%}$} & \nd 74.15{\good $_{+1.94\%}$}

\\ \midrule
\multirow{12}{*}{\rotatebox{90}{Semi-Dense}} &\multicolumn{2}{l}{ASpan~\cite{aspanformer}} 
& 
62.45 & 75.96 & 85.45

&48.66 & 64.47 & 76.89 

&
46.67 & 65.96 & 79.65&

41.65 & 61.39 & 76.48
 
\\

& \multicolumn{2}{l}{OETR+ASpan}  

& 
\rd 63.38{\good $_{+1.49\%}$} & \rd 76.89{\good $_{+1.22\%}$} & \rd 86.30{\good $_{+0.99\%}$} & 

\rd 53.24{\good $_{+9.41\%}$} & \rd 66.37{\good $_{+2.95\%}$} & \rd 77.94 {\good $_{+1.37\%}$} 

& 
\nd 47.37{\good $_{+1.50\%}$} & \nd 66.42{\good $_{+0.70\%}$} & \rd 79.97{\good $_{+0.40\%}$}& 

\rd 42.53{\good $_{+2.11\%}$} & \nd 62.27{\good $_{+1.43\%}$} & \rd 77.15{\good $_{+0.88\%}$}

\\	\dashlineours{2}{15}

& \multicolumn{2}{l}{\ours MESA+ASpan} 
& 
\nd 64.56{\good $_{+3.38\%}$} & \fs 78.23{\good $_{+2.99\%}$} & \fs 87.11{\good $_{+1.94\%}$} & 

\nd 57.32{\good $_{+17.80\%}$} & \nd 70.96{\good $_{+10.07\%}$} & \nd 81.53{\good $_{+6.03\%}$} 

& 
\fs 48.78{\good $_{+4.52\%}$} & \fs 67.32{\good $_{+2.06\%}$} & \fs 81.55{\good $_{+2.39\%}$} &

\fs 45.52{\good $_{+9.29\%}$} & \rd 61.97{\good $_{+0.94\%}$} & \fs 79.11{\good $_{+3.44\%}$}

\\
& \multicolumn{2}{l}{\ours DMESA+ASpan} 
& 
\fs 64.78{\good $_{+3.73\%}$} & \nd 78.00{\good $_{+2.69\%}$} & \nd 86.94{\good $_{+1.74\%}$}  & 

\fs 61.76{\good $_{+26.92\%}$} & \fs 75.15{\good $_{+16.57\%}$} & \fs 84.37{\good $_{+9.73\%}$} 

& 
\rd 46.95{\good $_{+0.60\%}$} & \rd 66.08{\good $_{+0.18\%}$} & \nd 80.00{\good $_{+0.44\%}$} & 

\nd 43.32{\good $_{+4.01\%}$} & \fs 62.93{\good $_{+2.51\%}$} & \nd77.85{\good $_{+1.79\%}$}

\\ \cmidrule(l){2-15} 

& \multicolumn{2}{l}{QT~\cite{quadtree}} 

&
62.46 & 76.18 & 85.75

&52.65 & 68.24 & 79.53 

&
47.62 & 66.48 & 80.07

&43.90 & 63.46 & 78.16

\\

& \multicolumn{2}{l}{OETR+QT}

& \rd 63.17{\good $_{+1.14\%}$} & \rd 77.32{\good $_{+1.50\%}$} & \rd 86.24{\good $_{+0.57\%}$} 

& \rd 53.72{\good $_{+2.03\%}$} & \rd 69.47{\good $_{+1.80\%}$} & \rd 80.69 {\good $_{+1.46\%}$} 

& \nd 48.32{\good $_{+1.47\%}$} & \nd 67.17{\good $_{+1.04\%}$} & \nd 80.69{\good $_{+0.77\%}$} 

& \rd 44.62{\good $_{+1.64\%}$} & \rd 64.17{\good $_{+1.12\%}$} & \rd 78.88{\good $_{+0.92\%}$}

\\ \dashlineours{2}{15}

& \multicolumn{2}{l}{\ours MESA+QT} 

& \fs 65.11{\good $_{+4.24\%}$} & \fs 78.32{\good $_{+2.81\%}$} & \fs 87.29{\good $_{+1.80\%}$} 

& \nd 57.28{\good $_{+8.79\%}$} & \nd 71.94{\good $_{+5.42\%}$} & \nd 83.16{\good $_{+4.56\%}$}

& \fs 50.32{\good $_{+5.67\%}$} & \fs 68.31{\good $_{+2.75\%}$} & \fs 81.24{\good $_{+1.46\%}$}

& \fs 47.21{\good $_{+7.54\%}$} & \fs 66.35{\good $_{+4.55\%}$} & \fs 80.13{\good $_{+2.52\%}$}

\\

& \multicolumn{2}{l}{\ours DMESA+QT}

& \nd 65.01{\good $_{+4.08\%}$} & \nd 77.91{\good $_{+2.27\%}$} & \nd 86.74{\good $_{+1.15\%}$}

& \fs 62.05{\good $_{+17.85\%}$} & \fs 75.69{\good $_{+10.92\%}$} & \fs 84.76{\good $_{+6.58\%}$} 

& \nd 48.32{\good $_{+1.47\%}$} & \rd 67.04{\good $_{+0.84\%}$} & \rd 80.45{\good $_{+0.47\%}$}

& \nd 45.62{\good $_{+3.92\%}$} & \nd 64.84{\good $_{+2.17\%}$} & \nd 79.08{\good $_{+1.18\%}$}

\\\cmidrule(l){2-15} 
 & \multicolumn{2}{l}{LoFTR~\cite{loftr}}
&59.00 & 73.30 & 83.37

&49.02 & 65.14 & 77.24 

&46.19 & 65.33 & 79.36

&43.14 & 63.06 & 77.82

\\
& \multicolumn{2}{l}{OETR+LoFTR}

& \rd 60.47{\good $_{+2.46\%}$} & \rd 74.22{\good $_{+1.26\%}$} & \rd 84.28{\good $_{+1.09\%}$}  

& \rd 51.33{\good $_{+4.71\%}$} & \rd 66.71{\good $_{+2.41\%}$} & \rd 78.56 {\good $_{+1.71\%}$}

& \nd 46.67{\good $_{+1.04\%}$} & \nd 65.84{\good $_{+0.78\%}$} & \nd 79.93{\good $_{+0.72\%}$}

& \rd 44.15{\good $_{+2.34\%}$} & \rd 64.17{\good $_{+1.76\%}$} & \nd 79.53{\good $_{+2.20\%}$}

\\ \dashlineours{2}{15}

& \multicolumn{2}{l}{\ours MESA+LoFTR}

& \nd 62.50{\good $_{+5.90\%}$} & \nd 75.33{\good $_{+2.77\%}$} & \nd 85.01{\good $_{+1.97\%}$}

& \nd 55.17{\good $_{+12.55\%}$} & \nd 70.34{\good $_{+7.98\%}$} & \nd 81.18{\good $_{+5.10\%}$} 

& \fs 48.52{\good $_{+5.04\%}$} & \fs 67.31{\good $_{+3.03\%}$} & \fs 81.29{\good $_{+2.43\%}$}

& \fs 46.23{\good $_{+7.16\%}$} & \fs 65.94{\good $_{+4.57\%}$} & \fs 80.21{\good $_{+3.07\%}$}
\\

& \multicolumn{2}{l}{\ours DMESA+LoFTR}

& \fs 62.54{\good $_{+5.96\%}$} & \fs 76.06{\good $_{+3.77\%}$} & \fs 85.50{\good $_{+2.55\%}$} 

& \fs 60.23{\impressive $_{+22.87\%}$} & \fs 74.22{\impressive $_{+13.94\%}$} & \fs 83.82{\impressive $_{+8.52\%}$}

& \rd 46.63{\good $_{+0.95\%}$} & \rd 65.56{\good $_{+0.35\%}$} & \rd 79.48{\good $_{+0.15\%}$}

& \nd 44.60{\good $_{+3.38\%}$} & \nd 65.06{\good $_{+3.17\%}$} & \nd 78.74{\good $_{+1.18\%}$}  
\\ \midrule

\multirow{4}{*}{\rotatebox{90}{Dense}} & \multicolumn{2}{l}{DKM~\cite{dkm}} 
&62.37 & 75.80 & 85.14

&61.70 & 75.24 & 84.49  

&51.07 & 68.75 & 81.24

& 50.95 & 68.73 & 81.33

\\

&  \multicolumn{2}{l}{OETR+DKM}

& \rd 63.35{\good $_{+1.57\%}$} & \rd 76.23{\good $_{+0.57\%}$} & \rd 85.57{\good $_{+0.51\%}$} 

& \rd 63.18{\good $_{+2.40\%}$} & \rd 76.71{\good $_{+1.95\%}$} & \rd 85.12 {\good $_{+0.75\%}$}  

& \rd 51.26{\good $_{+0.37\%}$} & \nd 69.31{\good $_{+0.81\%}$} & \rd 81.43{\good $_{+0.24\%}$}

& \rd 50.97{\good $_{+0.04\%}$} & \nd 68.79{\good $_{+0.09\%}$} & \rd 81.24{\bad $_{-0.11\%}$}

\\ 

\dashlineours{2}{15}
&  \multicolumn{2}{l}{\ours MESA+DKM}

& \nd 65.21{\good $_{+4.55\%}$} & \nd 77.94{\good $_{+2.82\%}$} & \fs 86.93{\good $_{+2.10\%}$} 

& \nd 64.22{\impressive $_{+4.08\%}$} & \nd 77.52{\impressive $_{+3.03\%}$} & \nd 86.34{\impressive $_{+2.19\%}$} 

& \fs 52.13{\good $_{+2.08\%}$} & \fs 69.84{\good $_{+1.59\%}$} & \fs 81.95{\good $_{+0.87\%}$}

& \fs 51.33{\good $_{+0.75\%}$} & \fs 68.97{\good $_{+0.35\%}$} & \fs 81.74{\good $_{+0.50\%}$}

\\ 
&  \multicolumn{2}{l}{\ours DMESA+DKM}

& \fs 65.48{\good $_{+4.99\%}$} & \fs 78.11{\good $_{+3.05\%}$} & \nd 86.63{\good $_{+1.75\%}$} 

& \fs 66.29{\impressive $_{+7.44\%}$} & \fs 78.50{\impressive $_{+4.33\%}$} & \fs 86.82{\impressive $_{+2.76\%}$} 

& \nd 51.60{\good $_{+1.04\%}$} & \rd 69.08{\good $_{+0.48\%}$} & \nd 81.50{\good $_{+0.32\%}$}

& \nd 51.18{\good $_{+0.45\%}$} & \nd 68.79{\good $_{+0.09\%}$} & \nd 81.44{\good $_{+0.14\%}$}

\\ \bottomrule

\end{tabular}
\begin{tablenotes}
     \item[$\dagger$] {\small The training size.}
\end{tablenotes}
\end{threeparttable}
}

 \vspace{-1.2em}
\end{table*}

\begin{table}[!t]
\centering
\caption{\textbf{Pose Estimation Results of RoMa.} We perform pose estimation experiments for leading dense point matcher RoMa~\cite{roma} on two standard benchmarks. Its training size is $560\times 560$ which is different from other baselines. Our methods significantly {\broma improve} its precision. } 
\label{tab:roma}
\resizebox{\linewidth}{!}{
\begin{tabular}{cllllll}
\toprule
\multirow{2}{*}{$560\times 560$} & \multicolumn{3}{c}{ScanNet1500} & \multicolumn{3}{c}{MegaDepth1500} \\ \cmidrule(l){2-4} \cmidrule(l){5-7}
 & AUC@5~$\uparrow$ & AUC@10~$\uparrow$ & AUC@20~$\uparrow$ & AUC@5~$\uparrow$ & AUC@10~$\uparrow$ & AUC@20~$\uparrow$ \\ \midrule
RoMa~\cite{roma} & 32.37 & 54.52 & 71.74 & 65.49 & 78.25 & 86.66 \\ \dashlineours{1}{7}
MESA+RoMa & 33.74{\broma$_{+4.23\%}$} & 56.33{\broma$_{+3.32\%}$} & 73.59{\broma$_{+2.58\%}$} & 67.91{\broma$_{+3.70\%}$} & 80.17{\broma$_{+2.45\%}$} & 88.47{\broma$_{+2.09\%}$} \\
DMESA+RoMa & 33.32{\broma$_{+2.93\%}$} & 55.26{\broma$_{+1.36\%}$} & 72.48{\broma$_{+1.03\%}$} & 68.62{\broma$_{+4.78\%}$} & 80.43{\broma$_{+2.79\%}$} & 88.41{\broma$_{+2.02\%}$} \\ \bottomrule
\end{tabular}}
\end{table}

\subsubsection{Results on ScanNet1500}
The pose estimation results on {ScanNet1500} are reported in \cref{tab:SN}. Same as the point matching experiments, we evaluate the pose estimation accuracy across three PM input resolutions.

\textit{For the sparse matcher}, our methods are able to enhance pose precision consistently and significantly across all resolutions. Notably, the precision gap across resolutions of SP+SG is minor, indicating its robustness. Thus our methods offer stable improvements across various resolutions.

\textit{For the semi-dense matchers} within A2PM framework, precision declines are observed at the training resolution, same as the point matching experiments, except for MESA+LoFTR. This possibly results from the excessive area size adjustment at the resolution of $640\!\times\!480$ (which is not square, cf. \cref{sec:a2pm-c}), which diminishes the advantages of A2PM.
However, both MESA and DMESA do significantly improve the pose precision at other square resolutions. This difference can be attributed to the overfitting issue inherent in Transformer~\cite{NaViT} of these baselines.
To improve precision, fine-tuning of these matchers on square resolutions can be conducted, which sets a new SOTA on this benchmark~\cite{mesa} and will be discussed later in \cref{sec:ab-irmf}.

\textit{For the dense matcher DKM}, although the best results are achieved at the training size, the disparity in accuracy across resolutions is relatively minor compared to semi-dense matchers. This indicates the better resolution robustness of DKM. Hence, our methods consistently deliver prominent precision enhancement for DKM across all resolutions. 
\textit{The results of RoMa} are reported in \cref{tab:roma}. Its square training resolution is compatible with A2PM framework. Our methods remarkably improve its performance. 

Overall, MESA+DKM yields the best results, which is consistent with the point matching experiments. SP+SG combined with our methods can surpass its semi-dense counterparts in the two square resolutions, \eg, AUC@20 of MESA+SP+SG: $61.49$ vs. MESA+LoFTR: $57.68$, with less computational cost. This proves the efficacy of our methods.
MESA performs better than DMESA by finding more area matches (higher ACR in \cref{tab:AM} and see \cref{fig:qbmd}). The completeness of area distribution in images is important for pose estimation. 
However, at $480\! \times\! 480$, this performance gap diminishes, particularly with DMESA+DKM surpassing MESA+DKM, demonstrating the practical value of DMESA under limited computational overhead. In our previous version~\cite{mesa}, we achieve advanced results by fine-tuning of PM model and RANSAC parameters, albeit at the expense of time and computational resources. Hence, we provide easily reproducible results here, with original model and MAGSAC++. These outcomes further validate the efficacy of our methods.

\begin{table*}[!t]
\caption{\textbf{Visual Odometry on KITTI360.} Relative gains are highlighted as subscripts. The \colorbox{colorFst}{\bf best}, \colorbox{colorSnd}{second} and \colorbox{colorTrd}{third} results are highlighted.}\label{tab:KITTI}
\resizebox{\linewidth}{!}{
\begin{tabular}{lllllllllllllll}
\toprule
 \multicolumn{3}{c}{\multirow{2}{*}{Visual Odometry}} & \multicolumn{3}{c}{Seq. 00} & \multicolumn{3}{c}{Seq.02} & \multicolumn{3}{c}{Seq. 05} & \multicolumn{3}{c}{Seq. 06} \\ \cmidrule(l){4-6} \cmidrule(l){7-9} \cmidrule(l){10-12} \cmidrule(l){13-15} 
\multicolumn{3}{c}{}  & R$_{err}\downarrow$      & t$_{err}\downarrow$       & AUC@5$\uparrow$      & R$_{err}\downarrow$      & t$_{err}\downarrow$       & AUC@5$\uparrow$ & R$_{err}\downarrow$      & t$_{err}\downarrow$       & AUC@5$\uparrow$  & R$_{err}\downarrow$      & t$_{err}\downarrow$       & AUC@5$\uparrow$    \\ \midrule
\multirow{4}{*}{\rotatebox{90}{Sparse}} & \multicolumn{2}{l}{SP~\cite{superpoint}+SG~\cite{superglue}}                    &0.053&0.99&80.41&0.064&1.08&79.63&0.056&1.11&79.22&0.061&0.95&81.42 \\
&\multicolumn{2}{l}{SGAM~\cite{sgam}+SP+SG}               & \rd 0.036 & \rd0.89 & \rd82.98{\good$_{+3.20\%}$} & \rd0.050 & \rd0.92 & \rd81.76{\good$_{+2.67\%}$} & \rd0.042 & \rd0.96 & \rd81.57{\good$_{+2.97\%}$} & \rd0.054 & \rd0.78 & \rd84.62{\good$_{+3.93\%}$} \\ \dashlineours{2}{15}
&\multicolumn{2}{l}{\ours MESA+SP+SG} & \fs 0.027 & \fs 0.58 & \fs88.89{\good$_{+10.55\%}$} & \fs0.041 & \fs0.82 & \fs87.33{\good$_{+9.67\%}$} & \fs0.034 & \fs0.78 & \fs87.62{\good$_{+10.60\%}$} & \fs0.037 & \fs0.62 & \fs88.24{\good$_{+8.38\%}$} \\
&\multicolumn{2}{l}{D\ours MESA+SP+SG} & \nd 0.034 & \nd0.84 & \nd83.53{\good$_{+3.88\%}$} & \nd0.046 & \nd0.89 & \nd82.80{\good$_{+3.98\%}$} & \nd0.039 & \nd0.92 & \nd82.10{\good$_{+3.64\%}$} & \nd0.041 & \nd0.75 & \nd85.28{\good$_{+4.74\%}$} \\ \midrule
\multirow{12}{*}{\rotatebox{90}{Semi-Dense}} &\multicolumn{2}{l}{ASpan~\cite{aspanformer}}                    &0.087&1.47&71.91&0.173&2.34&61.33&0.112&1.67&67.83&0.114&1.32&74.19   \\
& \multicolumn{2}{l}{SGAM+ASpan}  & \nd0.054 & \rd1.32 & \rd76.22{\good$_{+5.99\%}$} & \nd0.131 & \nd2.12 & \nd66.35{\good$_{+8.19\%}$} & \rd0.083 & \rd1.43 & \rd73.46{\good$_{+8.30\%}$} & \nd0.073 & \rd1.17 & \rd78.52{\good$_{+5.84\%}$} \\	\dashlineours{2}{15}
& \multicolumn{2}{l}{\ours MESA+ASpan} & \rd0.068 & \fs1.17 & \fs78.25{\good$_{+8.82\%}$} &\fs 0.121 & \rd2.17 & \fs66.42{\good$_{+8.30\%}$} & \nd0.078 & \nd1.39 & \fs76.72{\good$_{+13.11\%}$} & \rd0.078 & \fs0.78 & \fs81.11{\good$_{+9.33\%}$} \\
& \multicolumn{2}{l}{D\ours MESA+ASpan} & \fs0.051 & \nd1.23 & \nd76.45{\good$_{+6.31\%}$} & \rd0.134 & \fs2.10 & \rd66.08{\good$_{+7.74\%}$} & \fs0.064 & \fs1.26 & \nd75.21{\good$_{+10.88\%}$} & \fs0.062 & \nd1.01 & \nd80.36{\good$_{+8.32\%}$} \\ \cmidrule(l){2-15} 
& \multicolumn{2}{l}{QT~\cite{quadtree}} &0.131&2.97&58.10&0.164&3.84&55.26&0.152&3.60&53.42&0.153&2.87&60.06 \\
& \multicolumn{2}{l}{SGAM+QT}
& \rd0.104 & \rd2.76 & \nd61.32{\good$_{+5.54\%}$} & \rd0.131 & \rd3.42 & \rd58.23{\good$_{+5.37\%}$} & \fs0.144 & \rd3.02 & \rd56.71{\good$_{+6.16\%}$} & \nd0.110 & \rd2.41 & \rd63.21{\good$_{+5.24\%}$} \\ \dashlineours{2}{15}
& \multicolumn{2}{l}{\ours MESA+QT} & \nd0.094 & \fs2.43 & \fs66.71{\good$_{+14.82\%}$} & \nd0.113 & \fs3.14 & \fs66.21{\good$_{+19.82\%}$} & \rd0.162 & \nd2.87 & \fs64.52{\good$_{+20.78\%}$} & \rd0.123 & \nd2.33 & \fs71.22{\good$_{+18.58\%}$} \\
& \multicolumn{2}{l}{\ours DMESA+QT}& \fs0.080 & \nd2.49 & \rd61.10{\good$_{+5.16\%}$} & \fs0.102 & \nd3.17 & \nd61.26{\good$_{+10.86\%}$} & \nd0.151 & \fs2.59 & \nd60.50{\good$_{+13.25\%}$} & \fs0.094 & \fs2.04 & \nd69.43{\good$_{+15.60\%}$} \\\cmidrule(l){2-15} 
 & \multicolumn{2}{l}{LoFTR~\cite{loftr}}
& 0.112&1.55&72.80&0.110&1.49&74.16&0.112&1.49&71.59&0.114&1.28&75.66 \\
& \multicolumn{2}{l}{SGAM+LoFTR}
& \rd0.092 & \rd1.41 & \rd74.21{\good$_{+1.94\%}$} & \rd0.093 & \nd1.40 & \nd76.22{\good$_{+2.78\%}$} & \rd0.083 & \rd1.42 & \rd73.25{\good$_{+2.32\%}$} & \rd0.095 & \rd1.22 & \rd77.26{\good$_{+2.11\%}$} \\ \dashlineours{2}{15}
& \multicolumn{2}{l}{\ours MESA+LoFTR}
& \nd0.083 & \nd1.32 & \nd75.33{\good$_{+3.48\%}$} & \nd0.087 & \rd1.44 & \nd75.63{\good$_{+1.98\%}$} & \nd0.076 & \nd1.35 & \nd75.24{\good$_{+5.10\%}$} & \nd0.088 & \nd1.21 & \nd79.44{\good$_{+5.00\%}$} \\
& \multicolumn{2}{l}{\ours DMESA+LoFTR}
& \fs0.057 & \fs1.22 & \fs78.50{\good$_{+7.83\%}$} & \fs0.070 & \fs1.18 & \fs78.66{\good$_{+6.07\%}$} & \fs0.064 & \fs1.09 & \fs78.46{\good$_{+9.60\%}$} & \fs0.061 & \fs0.95 & \fs81.92{\good$_{+8.27\%}$} \\ \midrule
\multirow{4}{*}{\rotatebox{90}{Dense}} & \multicolumn{2}{l}{DKM~\cite{dkm}} 
&0.027&0.30&94.08&0.099&0.49&91.52&0.039&0.43&91.40&0.034&0.38&92.31 \\
&  \multicolumn{2}{l}{SGAM+DKM}
& \nd 0.022 & \nd 0.25 & \nd 95.32{\good$_{+1.32\%}$} & \rd0.046 & \nd0.41 & \nd 92.34{\good$_{+0.90\%}$} & \nd0.026 & \nd0.31 & \nd 92.68{\good$_{+1.40\%}$} & \rd0.027 & \nd0.34 & \nd93.67{\good$_{+1.47\%}$} \\ \dashlineours{2}{15}
&  \multicolumn{2}{l}{\ours MESA+DKM}
& \fs0.018 & \fs0.20 & \fs96.13{\good$_{+2.18\%}$} & \fs0.027 & \fs0.33 & \fs94.32{\good$_{+3.06\%}$} & \fs0.022 & \fs0.25 & \fs95.18{\good$_{+4.14\%}$} & \fs0.022 & \fs0.26 & \fs95.31{\good$_{+3.25\%}$} \\ 
&  \multicolumn{2}{l}{\ours DMESA+DKM}
& \nd0.022 & \rd0.29 & \rd94.13{\good$_{+0.05\%}$} & \nd0.034 &\rd 0.44 & \rd91.94{\good$_{+0.46\%}$} & \rd0.028 & \rd0.41 & \rd92.34{\good$_{+1.03\%}$} & \nd0.026 & \rd0.35 & \rd93.05{\good$_{+0.80\%}$} \\ \bottomrule

\end{tabular}
}
%
\end{table*}

\subsubsection{Results on ETH3D} 
The image pairs from ETH3D possess severe motion blur, light variation and textureless regions, leading to hard pose estimation. The results are summarized in \cref{tab:ETH3D}. According to the resolution range of images in ETH3D~\cite{nam2023diffmatch}, we choose $640\times 640$ as the large size and $480\times 480$ as the small one. To remove the resize distortion, we keep the original aspect ratio and use zero-padding to achieve square size, following \cite{loftr,pats}.

\textit{For the sparse matcher}, our approaches lead to a substantial increase in accuracy. Particularly noteworthy is the performance of MESA+SPSG, which achieves a superior AUC@20 of $48.85$ at $480\!\times\!480$ compared to $48.51$ at $640\!\times\!640$. This proves that our method enhances the resolution robustness of the point matcher.

\textit{For the three semi-dense matchers}, our approach consistently improve the accuracy of point matching at two different resolutions. With the smaller resolution, the original point matchers experience significant performance degradation, which our method effectively mitigate by restoring much of the lost performance, \eg, attaining a relative improvement up to $80.84\%$.

\textit{For the dense matcher}, the accuracy improvements brought by MESA remain impressive. DMESA relies on the coarse matching of ASpan for area matching. Due to the generalization issue of ASpan, its area matching accuracy falls short of that of MESA in this dataset. Moreover, dense DKM is more sensitive to the accuracy of area matching. Hence, a slight decrease in accuracy for DMESA+DKM at $640\!\times\!640$ is observed. However, at $480\!\times\!480$, DMESA still delivers performance improvements.

In summary, our approach considerably enhance the pose accuracy and resolution robustness of all point matchers on this dataset, significantly surpassing SGAM. Furthermore, DKM outperforms all other point matchers. Building upon this, MESA further enhance the accuracy of DKM, achieving performance comparable to that of in-domain dataset (ScanNet1500).

\subsubsection{Results on MegaDepth1500} 
MegaDepth serves as the outdoor training dataset for our baselines. In the MegaDepth1500 benchmark, we select a size of $832\!\times\!832$ as the large resolution, which is also used in training~\cite{loftr,aspanformer,quadtree}. A small resolution ($480\!\times\!480$) is also adopted for comparison. The resize distortion is avoided by shorter-side padding~\cite{superglue}. The results are reported in \cref{tab:YFCC+MD}.

\textit{For the sparse matcher}, our approaches yield stable and remarkable improvements in accuracy, overcoming OETR significantly. Moreover, the precision gap between resolutions is notably reduced by MESA/DMESA, demonstrating the effectiveness of our method and their robustness to resolution variations.

\textit{For the semi-dense matchers}, without the overfitting issue on the indoor training dataset (ScanNet), our methods consistently and considerably improve pose accuracy on the training size. This can be interpreted as the training resolution on this dataset is square, aligning well with the A2PM framework (cf. \cref{sec:a2pm-c}). Thus, the precise AM of our methods effectively reduce the matching redundancy and ultimately increase the performance of these semi-dense point matchers. The performance gap between resolutions is notably narrowed by our methods as well.

\textit{For the dense matcher DKM}, our methods improve pose accuracy at both resolutions, setting a new SOTA on this benchmark. Considering the AM sensitivity of DKM due to its dense computation, this proves the efficacy of MESA and DMESA. \textit{The results of RoMa} are reported in \cref{tab:roma}. It can be seen that the precision improvement from our methods is prominent as well, setting a new SOTA in this benchmark. 

Overall, our methods significantly surpass the previous SOTA, OETR. 
It is noteworthy that DMESA+DKM achieves superior results with the small resolution compared to the large one (AUC@5: $66.29$ of $480\!\times\!480$ vs. $65.48$ of $832\!\times\!832$), impressively leading to numerous reductions in computational costs.
This demonstrates the superior of DMESA in terms of both accuracy and efficiency.

\subsubsection{Results on YFCC} 
As per~\cite{adamatcher,mkpc}, the longer sides of YFCC images are typically shorter than $640$. Therefore, in this experiment, we choose $640\!\times\!640$ as the larger resolution and $480\!\times\!480$ as the smaller one. To prevent aspect ratio distortions, we apply shorter-side padding during resizing~\cite{superglue}. The results are reported in \cref{tab:YFCC+MD}.

\textit{For the sparse point matcher}, MESA brings about the most noteworthy improvement in accuracy at both resolutions. The performance improvement of DMESA is not as strong as OETR at $640\!\times\!640$, which may be attributed to the generalization issues of the coarse matcher on which DMESA relies. However, at the smaller resolution, DMESA outperforms OETR, highlighting the robustness of our method to resolution variations.

\textit{For the semi-dense matchers}, MESA also leads to prominent improvement for all matchers at both resolutions. DMESA sacrifices some accuracy but in return gains greater speed and improved flexibility. It also performs better at the smaller resolution, surpassing OETR.

\textit{For the dense matcher}, our approaches demonstrate consistent accuracy improvements, albeit relatively limited. This can be attributed to the increased difficulty in AM caused by the abundance of repetitiveness in YFCC, as DKM is sensitive to AM accuracy.

In sum, on this dataset, MESA shows the best performance and demonstrates superior generalization. Both of our approaches generally outperform OETR, proving their effectiveness.

\begin{figure*}[!t]
\centering
\includegraphics[width=\linewidth]{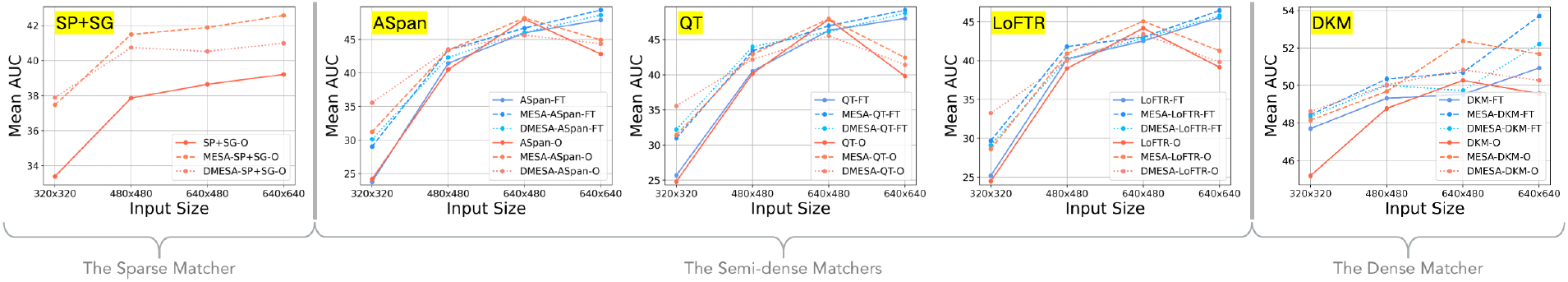}
\caption{{Experiment Results about Input Size and Model Fine-Tuning.} The figure illustrates pose estimation experiments with five point matchers on ScanNet1500, displaying a line graph of pose accuracy concerning input sizes. Two types of dashed lines represent the results of MESA and DMESA. The {\color{orange}orange} lines indicate the outcomes of the original models trained on the resolution of $640\times480$, while the {\color{SkyBlue}blue} lines represent the results of models fine-tuned on the resolution of $640\times640$.}
\vspace{-1.2em}
\label{fig:ab-size-ft}
\end{figure*}

\subsubsection{Discussion}\label{sec:pe-diss}
Here, we provide a summary and analysis of the pose estimation results obtained from the above four datasets. 
Generally, MESA and DMESA each have their own merits.

First, for the indoor datasets, MESA demonstrates superior accuracy compared to DMESA (cf. \cref{tab:SN} and \cref{tab:ETH3D}). This superiority can be attributed to the fact that MESA achieves AM results by fusing SAM segments in AG.
Consequently, area matches produced by MESA have better ACR in \cref{tab:AM} and contains rich semantic information, usually encompassing complete semantic entities. 
This proves advantageous for inside-area PM, especially in intricate indoor scenes.

Conversely, DMESA, which relies on patch matching, achieves area matches with constrained sizes but excels in AM accuracy (cf. \cref{tab:AM}). 
Hence, it performs better in scenes with repetitive patterns and coarser semantic granularity, like MegaDepth (cf. \cref{tab:YFCC+MD} left).
On the other hand, MESA possesses better cross-dataset generalization than DMESA (cf. \cref{tab:YFCC+MD} right), because of the limitation of off-the-shelf patch matching in DMESA. More generalization experiments can be found in \cref{sec:ab-cd} of the appendix.
However, DMESA requires no additional training and is faster than MESA, making it a compelling choice in practice.
Overall, both the proposed methods significantly enhance matching accuracy for all PM baselines.

\subsection{Visual Odometry}\label{sec:exp-vo}
To further evaluate the performance of our methods in downstream tasks, we conducted experiments on visual odometry, which densely estimates the camera motion in the driving scene, using the KITTI360 dataset.

\subsubsection{Experimental setup} 
According to the static scene assumption~\cite{orbslam} of our baselines, we select four sequences from the dataset that contain a small number of moving objects, each comprising 3000 images.
The parameter $T_{E_{max}}$ of MESA is set as $0.25$. The input image size is $640 \times 640$. 
We utilize the same baselines as in other experiments, which are trained on ScanNet~\cite{scannet}.
MAGSAC++ is used to estimate poses.
Following~\cite{S2LD}, we report the relative pose errors (RPE), including the rotational error ($R_{err}$) and translation error ($t_{err}$), along with the pose estimation AUC@$5$ for better comparisons.

\subsubsection{Results} 
The results are presented in \cref{tab:KITTI}.
\textit{For the sparse matcher}, MESA and DMESA both obtain notable and consistent improvement across all sequences, surpassing SGAM, validating the effectiveness of our methods.

\textit{For the semi-dense matchers}, the performance improvement brought by our methods are remarkable. MESA and DMESA respectively yield performance boosts of up to $20.78\%$ and $15.60\%$. It is worth noting that when employing LoFTR, DMESA outperforms MESA across all sequences with increased efficiency, making it a more practical choice in this situation.

\textit{For the dense matcher}, our methods further enhance its accuracy, achieving the best results on this dataset. While the performance of DMESA is inferior to SGAM, it provides an enhancement for DKM. This possibly is caused by the AM sensitivity of DKM and generalization challenge of DMESA.

In this experiment, our method has demonstrated the ability to enhance the accuracy of visual odometry for all point matchers. Notably, when integrated with our approach, SP+SG exhibits substantial improvements. It surpasses semi-dense matchers by a considerable margin, even approaching the performance level of dense matching. Given the efficiency advantage of sparse matching over dense/semi-dense matching, this evidently emphasizes the practical value of our methods.

\subsection{Study of Input Resolution and Model Fine-tuning}\label{sec:ab-irmf}
In the experiments conducted in ScanNet1500, the resolution overfitting of PM baselines is observed.
Especially for the Transformer-based methods, our methods lead to performance decline at the training resolution (cf. \cref{tab:SNMMA} and \cref{tab:SN}).
In this section, we show that this issue can be migrated by model fine-tuning tailored to the square resolution.
Moreover, we explore a broader range of resolutions to investigate the resolution robustness of our methods with or without model fine-tuning.

\subsubsection{Experimental setup} 
We select four sets of resolutions, including three square sizes ranging from small to large ($[320^2,480^2,640^2]$) and the training resolution of $640\!\times\!480$. 
The overfitting issue is absent in the outdoor training dataset, which uses the square resolution in training.
Thus, we fine-tune the point matchers at $640\!\times\!640$, obtaining fine-tuned models identified by a `-FT' suffix.
Except for SP+SG, it does not encounter any overfitting issue in previous experiments. 
The original models are labeled with an `-O' suffix. We evaluate the performance of different methods using the average of pose estimation AUC@5/10/20, referred to as \textit{Mean AUC}.

\subsubsection{Results} 
The results are depicted in \cref{fig:ab-size-ft}. \textit{For the sparse matcher}, our methods not only significantly improve performance but also reduce the accuracy gap across resolutions. The results of DMESA are particularly strong at $320\!\times\!320$, surpassing those of MESA.

\textit{For the semi-dense matchers}, we observe that the original models exhibit a consistent performance peak at the training size, indicating overfitting to this training resolution and a high sensitivity to resolution variations. Consequently, our methods show limited advantages over the original model at the training size, but they do improve accuracy at other sizes. For the fine-tuned models, our approaches enhance accuracy across all resolutions, surpassing the performance of original models at the training size. We also note a decrease in performance at $640\!\times\!480$ for the fine-tuned models, indicating that fine-tuning may not be the optimal solution for resolution sensitivity. DMESA demonstrates outstanding performance at the small resolution of $320\!\times\!320$.

\textit{For the dense matcher}, although the performance peak still exists, the original model exhibits much less sensitivity to resolution compared to the semi-dense matchers (note the scale of the vertical axis). Therefore, our methods significantly improve accuracy for the original model across all resolutions. After fine-tuning, the performance of MESA and DMESA is further enhanced at $640\!\times\!640$. DMESA continues to excel at the small resolution, consistent with its emphasis on efficiency.

Overall, our methods boost the accuracy of all point matchers and enhance their robustness to resolution variations, irrespective of whether the point matcher is fine-tuned or not. However, model fine-tuning at square resolution is effective to address the overfitting issue for Transformer-based matchers. Notably, MESA demonstrates superior accuracy at high resolutions, whereas DMESA stands out at low resolutions. Both approaches are viable for practice based on specific computational resource constraints.

\subsection{Ablation Study}\label{sec:exp-ab}

\begin{figure}[!t]
\centering
\includegraphics[width=\linewidth]{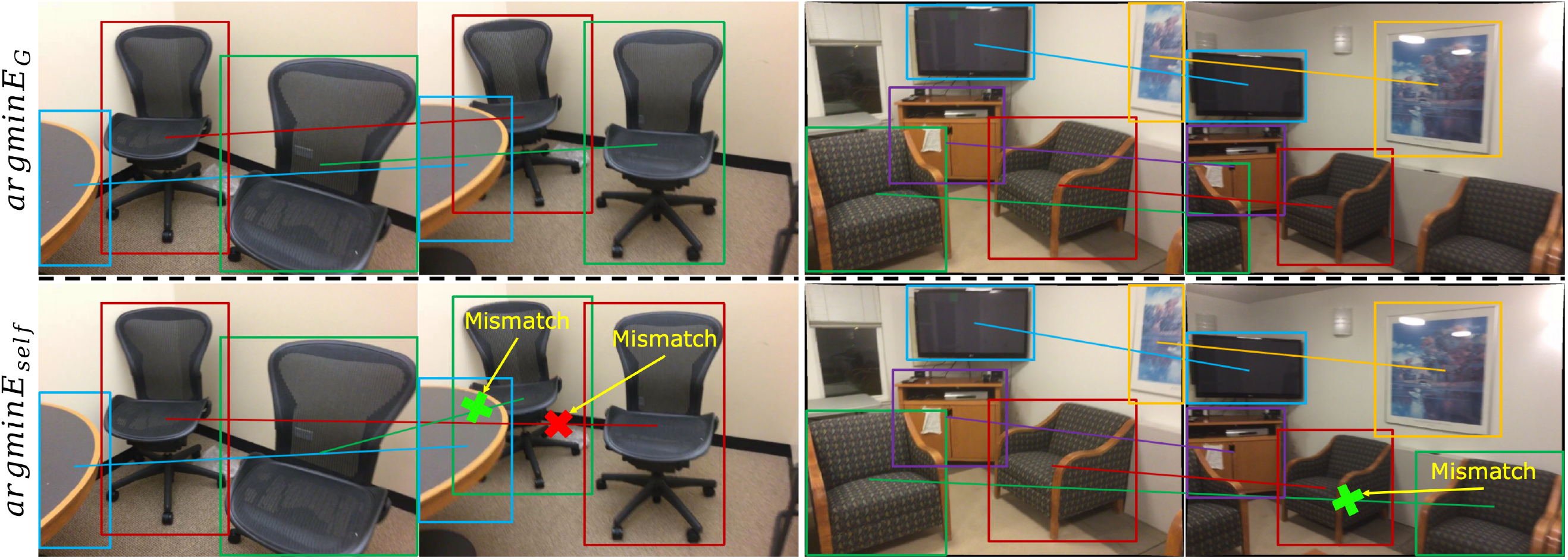}
\caption{\textbf{The qualitative comparison of Global Energy Refinement.} As AG structures of both images are considered by $E_G$, objects with the same apparent can be distinguished according to their neighbors, which are mismatched by $\arg\min E_{self}$, revealing the robustness of $\arg\min E_G$ under repetitive patterns.}
\label{fig:e_qs}
\end{figure}

\subsubsection{Understanding MESA}\label{sec:exp-ab-mesa}
To evaluate the effectiveness of our design in MESA, we conduct a comprehensive ablation study here.
We use MESA+ASpan as the baseline. The input resolution of PM is $640\times 640$ and the model of ASpanFormer is fine-tuned at $640\times 640$ as well. 
\begin{table}[t]
\caption{\textbf{Ablation study of MESA.} Four variants of MESA+ASpan are evaluated for area matching and pose estimation on the ScanNet1500 to demonstrate the importance of various components.} \label{tab:ASR}
\resizebox{\linewidth}{!}{
\begin{tabular}{ccccccc}
\toprule
\multicolumn{3}{l}{Method}               & AOR~$\uparrow$   & AMP@0.6~$\uparrow$ & PoseAUC@5~$\uparrow$ & ACR~$\uparrow$ \\ \midrule
\multicolumn{3}{l}{MESA+ASpan (\textbf{Ours})}          & 72.75 & 89.09   & 27.50       & 95.80      \\ \hdashline \noalign{\vskip 1pt}
\multicolumn{3}{l}{ w/~\textit{CSD}}           & 69.23 & 84.21   & 26.78      & 87.33    \\
\multicolumn{3}{l}{ w/~\textit{DesSim.}~\cite{pats}}           & 63.71 & 62.91   & 26.05      & 80.11     \\
\multicolumn{3}{l}{ w/~\textit{SEEMSeg.}~\cite{SEEM}}          & 70.58 & 85.52   & 26.18      & 72.51     \\
\multicolumn{3}{l}{ w/~$\arg\min E_{self}$} & 70.98     & 87.56       & 26.96          & 91.64        \\ \bottomrule
\end{tabular}
}
\vspace{-1.2em}
\end{table}

\boldparagraph{Area Graph Construction.} To justify the AG of MESA, we adopt a naive approach to match areas, which is comparing area similarity densely (CSD). In particular, we first select areas with proper size from all SAM areas of two images. The similarity of each area to all areas in the other images is then calculated and area matches with the greatest similarity is obtained. The comparison results are summarized in \cref{tab:ASR}. As AG can generate more proper areas for matching, MESA w/ CSD gets less area matches.
Thus, the area and point matching performance is also decreased by CSD. Moreover, CSD results in a significant increase in time of area matching (nearly $\times 10$ slower than MESA), due to its inefficient dense comparison.

\boldparagraph{Area Similarity Calculation.}
In contrast to our classification formulation for area similarity calculation, another straightforward method~\cite{pats} involves calculating the distance between learning descriptors of areas.
Thus, we replace our learning similarity with descriptor similarity in~\cite{pats} (\textit{DesSim}) and conduct experiments in ScanNet to investigate the impact.  
The results are summarized in \cref{tab:ASR}, including the area number per image, area matching and pose estimation performance.
Overall, the performance of \textit{DesSim} experiences a noticeable decline, due to poor area matching precision, indicating the effectiveness and importance of proposed learning similarity calculation.

\boldparagraph{Image Segmentation Source.}
We relay on SAM to achieve areas with implicit semantic, whose outstanding segmentation precision and versatility contribute to our leading matching performance.
However, areas can also be obtained from other segmentation methods.
Therefore, to measure the impact of different segmentation 
sources, we exchange the segmentation input from SAM~\cite{sam} with that from SEEM~\cite{SEEM} (\textit{SEEMSeg.}) and evaluate the performances. In \cref{tab:ASR}, MESA with \textit{SEEMSeg.} gets a slight precision decline and fewer areas compared with SAM, leading to decreased pose estimation results. These results indicates that the advanced segmentation favors our methods. Notably, MESA with \textit{SEEMSeg.} also achieves a slight improvement for ASpan, proving the effectiveness of MESA.

\boldparagraph{Global Energy Refinement.}
After \textit{Graph Cut}, the proposed global matching energy for the final area matching refinement considers structures of both AGs of the input image pair. To show the importance of this dual-consideration, we replace the global energy with naive $E_{self}$ in \cref{eq:e_self} ($\arg\min E_{self}$) and evaluate the performance. In \cref{tab:ASR}, the refinement relying on $E_{self}$ produces decreased area matching precision and a subsequent decline in pose estimation performance, due to inaccurate area matches especially under repetitiveness. The qualitative results shown in \cref{fig:e_qs} further indicate the better robustness of global energy under repetitiveness due to dual graph structure capture.

\begin{table}[!t]
\centering
\caption{\textbf{Ablation study of DMESA.} We show area matching and pose estimation performance of SP+SG+DMESA on ScanNet1500 w.r.t two parameters of DMESA. The parameters we applied are \textbf{highlighted}.} \label{tab:dmesa-ab}
\resizebox{\linewidth}{!}{
\begin{threeparttable}
\begin{tabular}{cccccc}
\toprule
\multicolumn{1}{l}{} &  & \multicolumn{1}{c}{AOR~$\uparrow$} & \multicolumn{1}{c}{AMP@0.6~$\uparrow$} & \multicolumn{1}{c}{Pose AUC@5~$\uparrow$} & \multicolumn{1}{c}{ACR~$\uparrow$} \\ \midrule
\multirow{5}{*}{\rotatebox{90}{$T_c$}} 
 & $f(1/8)$\tnote{$\dagger$} & 79.11 & 88.02 & 22.13 & 77.12 \\
 & $f(1/2)$\tnote{$\dagger$} & 79.26 & 87.54 & 21.68  & 78.02\\
 & $\bm{f(1)}$\tnote{$\dagger$} & \textbf{78.13} & \textbf{86.45} & \textbf{22.19} & \textbf{79.44} \\
 & $f(2)$\tnote{$\dagger$} & 77.41 & 84.13 & 21.56 & 80.22 \\
 & $f(4)$\tnote{$\dagger$} & 75.26 & 80.55 & 21.28 & 81.39 \\ \midrule
\multirow{5}{*}{\rotatebox{90}{$S_{EM}$}} 
 & 0 & 75.82 & 85.97 & 21.68 & 78.84 \\
 & 1 & 79.38 & 86.52 & 21.37 & 76.53 \\
 & \textbf{3} & \textbf{78.13} & \textbf{86.45} & \textbf{22.19} & \textbf{79.44} \\
 & 5 & 77.30 & 86.59 & 21.57 & 81.13 \\
 & 7 & 76.14 & 85.74 & 21.69 & 83.66 \\ \bottomrule 
\end{tabular}
\begin{tablenotes}
     \item[$\dagger$] {\small $f(x)=\frac{1}{2\pi}e^{-x}$}
\end{tablenotes}
\end{threeparttable}
}
 \vspace{-1.2em}
\end{table}
\subsubsection{Understanding DMESA}
In contrast to the complex process of MESA, DMESA involves just two parameters from two components necessitating configuration. One parameter is the confidence threshold $T_c$ for area extraction from the matching distribution; while the other, the EM algorithm step number $S_{EM}$, regulates the fusion degree of the results from the two matching directions. In this section, we perform ablation study on the two parameters using SP+SG as the point matcher on ScanNet1500, assessing both area matching and pose estimation precision. Given the orthogonal nature of two parameters, during experiments focusing on one parameter, we retain the other at its default setting ($T_c=e^{-1}/(2\pi), S_{EM}=3$).

\boldparagraph{Confidence Threshold.}
On the AM distribution, only locations with confidence exceeding the $T_c$ contribute to area matching. These confidences are determined by a GMM from the patch matches. Hence, we use the standard Gaussian distribution, $\mathcal{N}({\bm x}|0,I)\!=\!{1}/(2\pi)\! \cdot \! \exp(-{\Vert {\bm x}\Vert}^2/2)$, as a reference to set this threshold $T_c$.
Specifically, we choose {\small$\Vert {\bm x} \Vert\!=\![1/2,1,\sqrt{2},2,2\sqrt{2}]$}, resulting in corresponding confidence thresholds of: {\small$[e^{-1/8}/(2\pi), e^{-1/2}/(2\pi), e^{-1}/(2\pi), e^{-2}/(2\pi), e^{-4}/(2\pi)]$}. The results are reported in \cref{tab:dmesa-ab}. We observe that AM accuracy increases with $T_c$. However, as the coverage of areas (\textit{ACR}) in the image decreases at the same time, there is a risk of missing valid point matches, harmful to pose estimation accuracy. Therefore, we set $T_c=e^{-1}/(2\pi)$ to strike a balance between AM accuracy and coverage, achieving the best pose accuracy.

\begin{figure*}[!t]
\centering
\includegraphics[width=\linewidth]{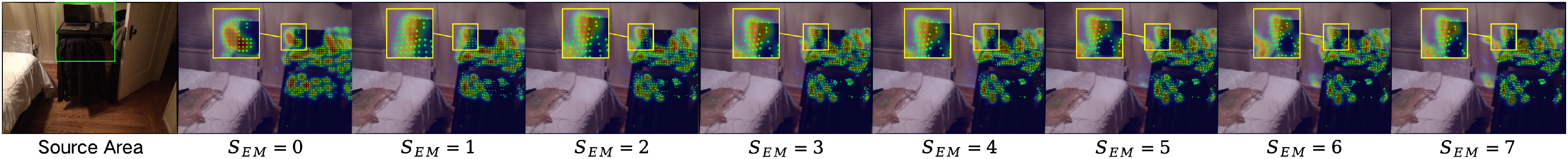}
\vspace{-1.2em}
\caption{\textbf{The qualitative results of finite-step EM refinement of DMESA.} We present the source area along with its corresponding patch matches in the target image across various EM step numbers ($S_{EM}$). The green dots are patch centers and the distributions of GMM are visualized as well. With the increase of $S_{EM}$, the patches become clustered into image regions with distinct features (\eg, the upper-left corner of the laptop in the bottom). These regions exhibit high confidence in both matching directions, thereby enhancing overall accuracy. Additionally, it is evident that an excessively large $S_{EM}$ does not contribute to patch refinement, as the patches are already stabilized in the initial stages.}
\label{fig:qua-EM}
\end{figure*}

\boldparagraph{Step number of EM.}
DMESA merges the results from two different matching directions using a finite-step EM algorithm, thus improving the area matching through cycle consistency. Hence, the number of EM algorithm steps should be kept moderate; excessively low or high step counts may bias the refined results towards one matching direction rather than achieving a consistent outcome between the two. We experimented with values of $S_{EM}=[0,1,3,5,7]$, and the results are presented in \cref{tab:dmesa-ab}. It is evident that the step count influences both the area matching accuracy (\textit{AOR}) and coverage (\textit{ACR}), thus resulting nonlinear variations in pose estimation precision. A moderate setting of $S_{EM}=3$ can yield the best pose estimation accuracy. We also provide qualitative results about the $S_{EM}$ in \cref{fig:qua-EM}, further justifying the efficacy of the moderate setting.

\subsubsection{SAM vs. SAM2}

\begin{table}[!t]
    \centering
    \caption{\textbf{SAM vs. SAM2.} We investigate the impact of recent SAM2 on our methods. The experiments are constructed on ScanNet1500. We report the performance of area matching, point matching and pose estimation.}\label{tab:sam2}
    \resizebox{\linewidth}{!}{
    \begin{tabular}{ccccc}
    \toprule
        ~ & AOR $\uparrow$ & ACR$\uparrow$ & MMA@3$\uparrow$ & AUC@5$\uparrow$  \\ \midrule
        SP~\cite{superpoint}+SG~\cite{superglue} & - & - & 20.53 & 19.27   \\ \midrule
        SAM~\cite{sam}+MESA+SP+SG & 68.44 & \textbf{94.57} & 25.34 & \textbf{22.72}   \\ 
        SAM2~\cite{sam2}+MESA+SP+SG & \textbf{70.67} & 85.43 & \textbf{25.57} & 22.48  \\ \midrule
        SAM+DMESA+SP+SG & 78.13 & \textbf{79.44} & 23.46 & \textbf{22.19}   \\ 
        SAM2+DMESA+SP+SG & \textbf{82.15} & 67.23 & \textbf{24.77} & 22.11 \\ \bottomrule
    \end{tabular}}
    \vspace{-1.2em}
\end{table}

The recent SAM2~\cite{sam2} enhances segmentation consistency across various video frames, which is particularly beneficial for area matching. Thus, we conduct experiments to assess the impact of substituting SAM with SAM2 in our methods. We opt for the ScanNet1500 dataset, given that its image pairs are sourced from indoor videos. We employ SP+SG as our point matcher with an input resolution of $480\times 480$. The performance metrics for area matching (\textit{AOR}, \textit{ACR}), point matching (\textit{MMA@3}), and pose estimation (\textit{AUC@5}) are presented in Table \ref{tab:sam2}. The results indicate a notable enhancement in area and point matching accuracy when using SAM2, attributed to its improved segmentation consistency. However, there is a substantial decrease in area coverage with SAM2, likely due to its reduced number of masks compared to SAM, which is a trade-off for achieving segmentation coherence. Consequently, the pose estimation accuracy diminishes when employing SAM2. The degradation is minor though, implying the stability of our methods with various segmentation sources.

\begin{table}[!t]
\centering
\caption{\textbf{Time Consumption Comparison.} The average time cost of area matching per image utilizing different methods is summarised. The experiment is conducted on 500 image pairs sampled from YFCC. }\label{tab:time}
\resizebox{0.6\linewidth}{!}{
\begin{tabular}{lll}
\toprule
Method & Step & Time(ms) \\ \midrule
\multirow{5}{*}{MESA} & AG Construction & 384.22 \\
 & Similarity Calculation & 2953.17 \\
 & Graph Cut & 3.24 \\
 & Energy Minimization & 6.15 \\ \dashlineours{2}{3}
 & \textbf{Total} & \cellcolor{green!20}3346.78 \\ \midrule
\multirow{5}{*}{DMESA} & AG Construction & 378.13 \\
 & Coarse Matching & 118.35 \\
 & Patch Confidence Rendering & 84.56 \\
 & EM Refinement & 125.39 \\ \dashlineours{2}{3}
 & \textbf{Total} & \cellcolor{green!10}706.43 \\ \midrule
SGAM~\cite{sgam} & \textbf{Total} & \cellcolor{green!10}693.87 \\ \midrule
OETR~\cite{OETR} & \textbf{Total} & \cellcolor{green!10}653.74 \\ \bottomrule
\end{tabular}}
\vspace{-1.2em}
\end{table}
\subsection{Running Time Comparison}
In this section, the average time consumption of AM methods on each image pair is recorded, to demonstrate the efficiency. 

\subsubsection{Experimental setup}
To facilitate the comparison with other methods~\cite{sgam,OETR}, we choose the YFCC dataset and randomly sample 500 images ($480\!\times\!480$) from it for constructing the experiment.
This experiment is conducted on an Intel Xeon Silver 4314 CPU and a GeForce RTX 4090 GPU.
Our comparative methods include SGAM, which establishes area matches grounded on explicit semantic, and OETR, which focuses on establishing matches of co-visible areas.
In addition, we record the time consumption of the every individual modules of both MESA and DMESA.

\subsubsection{Results}
The results are presented in \cref{tab:time}.
From the table, it is evident that MESA incurs the longest time consumption, primarily due to the intensive computation involved in assessing area similarities. This can be further attributed to the sparse AM framework of MESA, which leads to repetitive computation as described in \cref{sec:dmesa}. Therefore, DMESA adopts a dense AM framework, effectively reducing the repetitive computation. By incorporating a coarse matching stage of an off-the-shelf point matcher, the cost of single area matching is also reduced. Ultimately, DMESA achieves a speed approximately $5$ times faster than MESA while maintaining competitive accuracy. Furthermore, the speed of DMESA aligns closely with that of other two SOTA methods.

\section{Discussion and Conclusion}
The results presented in \cref{sec:exp} prove the effectiveness of the proposed MESA and DMESA. Both of them consistently and significantly increase precision for \textbf{six} PM baselines on \textbf{three} tasks across \textbf{five} various datasets. DMESA demonstrates a nearly fivefold speed improvement over MESA while maintaining competitive accuracy, offering a superior accuracy/speed trade-off. Besides, our methods substantially improve the matching robustness against variations in data domain and input resolution, benefiting the downstream tasks. 
However, our methods still suffers from challenges like severe repetitiveness. The utilization of SAM features is also inadequate.
Overcoming these limitations is a primary objective for our future work.
A detailed analysis of these limitations can be found in \cref{sec:lfw} of the appendix.

For effective matching redundancy reduction, we propose MESA and DMESA to leverage the general image understanding capability of SAM in this work. 
While both methods focus on area matching from SAM results, MESA follows a sparse framework, whereas DMESA adopts a dense fashion. 
Specifically, we first propose a novel graph, named AG, to model the global context of SAM segments and identify areas with prominent semantics. Then, MESA minimizes energy on the graph to match these areas, leveraging graphical models.  
To overcome the efficiency limitation from the sparse nature of MESA, DMESA is proposed as a dense counterpart.
It deduces area matches from off-the-shelf patch matches, by utilizing GMM to generate dense matching distributions for areas. To further refine accuracy, DMESA employs a finite-step EM algorithm to pursuit cycle-consistency. Our methods enable integration with PM baselines belonging to sparse, semi-dense and dense frameworks. 
In extensive experiments, consistent and prominent precision improvements from our methods for various PM baselines are observed across different datasets, confirming their efficacy.

{
    \bibliographystyle{IEEEtran}
    \bibliography{IEEEabrv,main}
}


%

%

\begin{IEEEbiography}[{\includegraphics[width=1in,height=1.25in,clip,keepaspectratio]{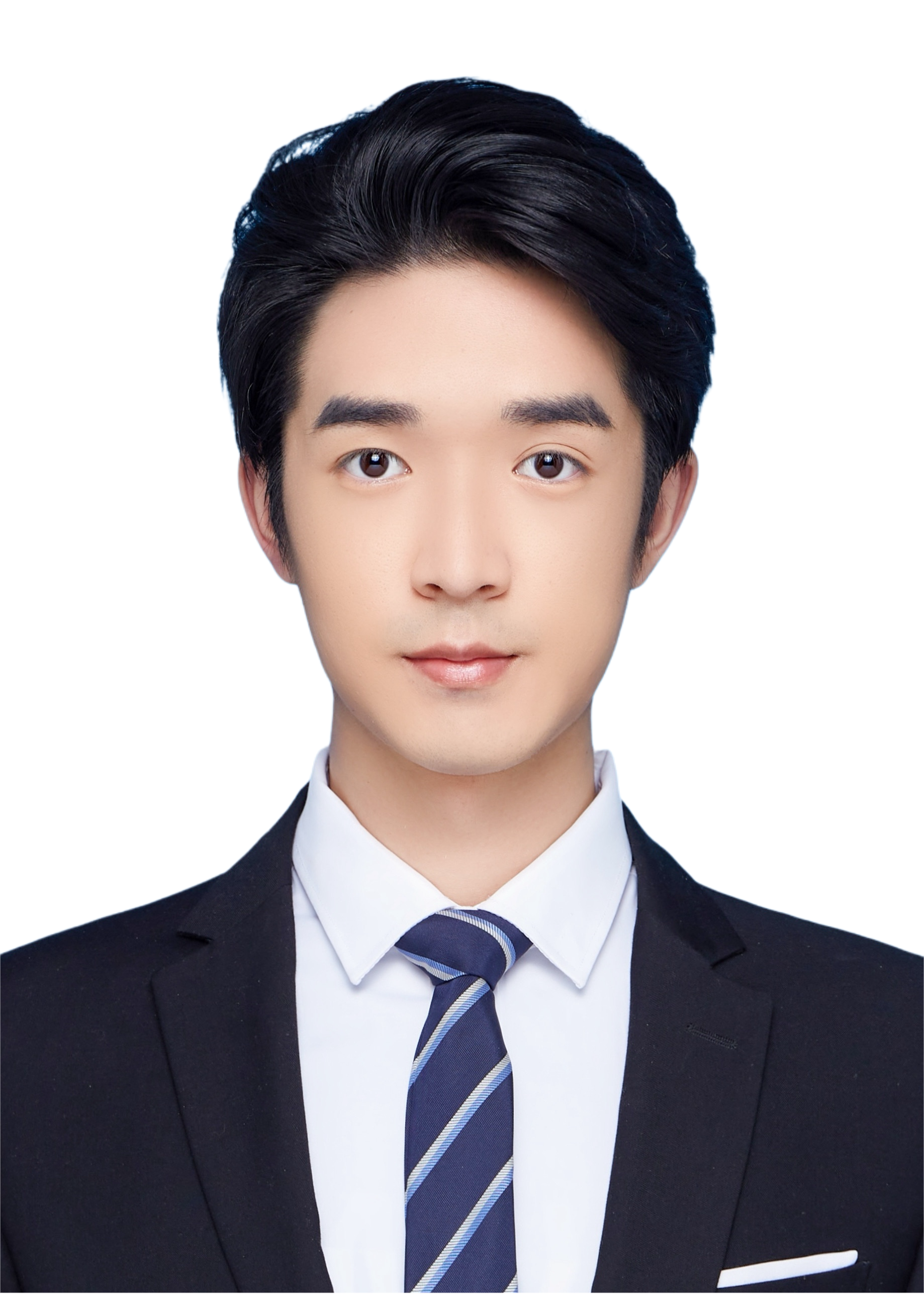}}]{Yesheng Zhang} (Student Member, IEEE)
received the B.S. and M.S. degrees in biomedical engineering from Shanghai Jiao Tong University (SJTU), Shanghai, China, in 2016 and 2022. He is currently working toward the Ph.D. degree with the Department of Automation, School of Electronic Information and Electrical Engineering, SJTU. His research interests include feature matching, visual SLAM and 3D reconstruction.
\end{IEEEbiography}

\begin{IEEEbiography}[{\includegraphics[width=1in,height=1.25in,clip,keepaspectratio]{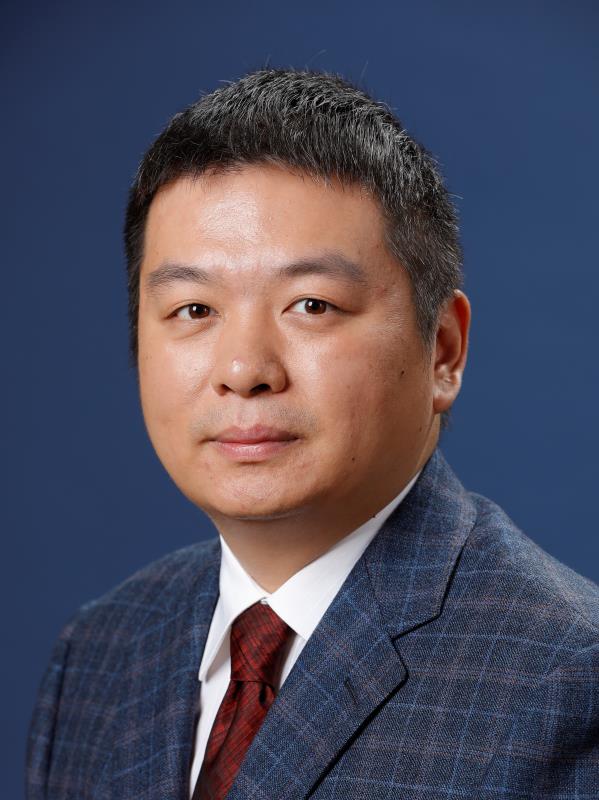}}]{Shuhan Shen} (Senior Member, IEEE) received the B.S. and M.S. degrees from Southwest Jiaotong University, Chengdu, China, in 2003 and 2006, respectively, and the Ph.D. degree from Shanghai Jiaotong University, Shanghai, China, in 2010. He is currently a Professor with the Institute of Automation, Chinese Academy of Sciences, and an Adjunct Professor with the School of Artificial Intelligence, University of Chinese Academy of Sciences. His research interests include 3D computer vision, in particular 3D reconstruction of large-scale scenes, 3D perception for intelligent robot, and 3D semantic reconstruction.
\end{IEEEbiography}

\begin{IEEEbiography}[{\includegraphics[width=1in,height=1.25in,clip,keepaspectratio]{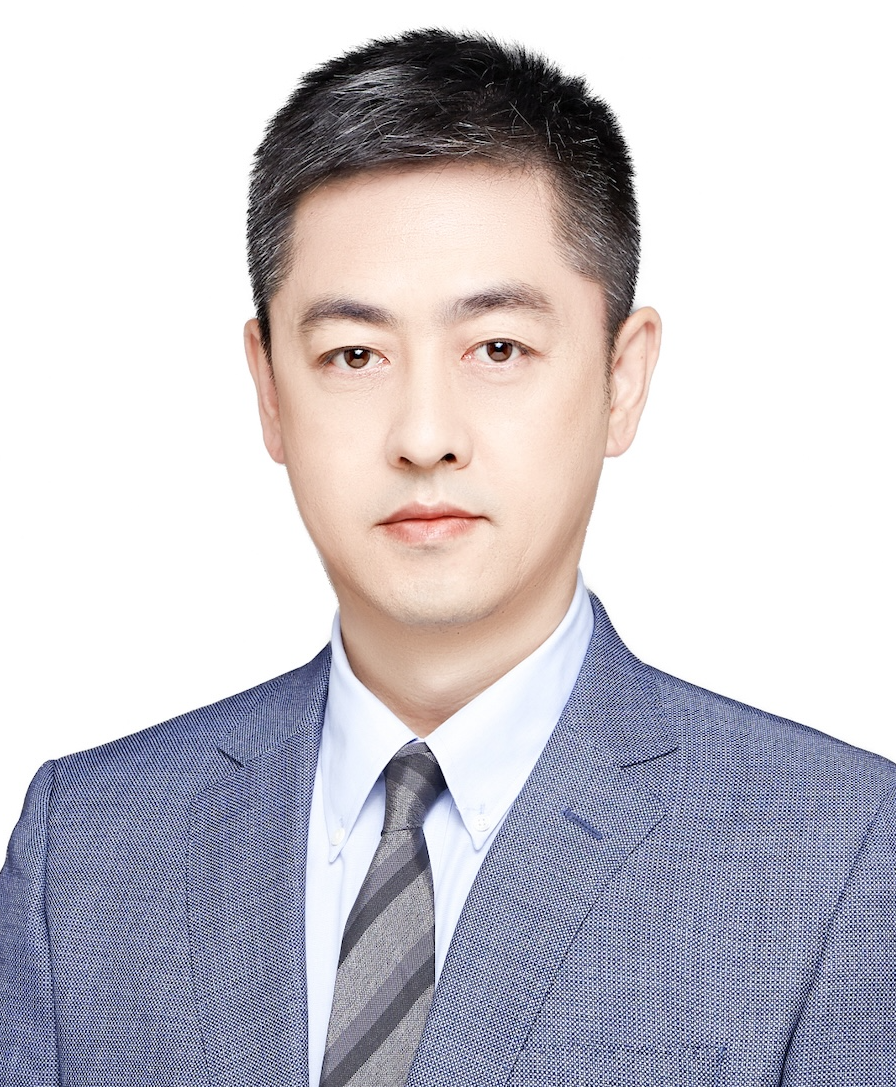}}]{Xu Zhao}
(Member, IEEE) received the Ph.D. degree in pattern recognition and intelligent system from Shanghai Jiao Tong University (SJTU), Shanghai, China, in 2011. He is currently a Full Professor with the Department of Automation, School of Electronic Information and Electrical Engineering, SJTU. He was a Visiting Scholar with the Beckman Institute, University of Illinois Urbana-Champaign, Urbana, IL, USA, from November 2007 to December 2008, and a Postdoc Research Fellow with Northeastern University, Boston, MA, USA, from August 2012 to August 2013. His research interests include visual analysis of human motion, machine learning and image/video processing.
\end{IEEEbiography}




\clearpage
\appendices

\section{Experiments of resolution overfitting issue}\label{sec:roi}
In this section, we provide the additional experiments to investigate resolution overfitting issue in Transformer-based methods. Specially, we choose the famous LoFTR~\cite{loftr} and its improved variants QT~\cite{quadtree} as the baselines. In ScanNet, their training size is $640\times 480$ with the aspect ratio $4/3$. Intuitively, the resolution overfitting is highly related to the aspect ratio, as which leads to the distortion of image context. Thus, we conduct comparison experiments on another size $896\times 672$, which maintains the same aspect ratio and improves the resolution. Considering the larger resolution brings more details and no distortion with the same aspect ratio, the performance between two sizes should be comparable. However, as we reported in \cref{tab:ro}, the performance of the baselines showcase significant descend ($28.56$ vs. $2.88$ for QT on AUC@$5$). This imply the hard resolution overfitting issue of Transformer-based point matchers, which is possibly caused by the positional encoding~\cite{NaViT}. On the other hand, our methods can increase the performance of the point matchers at the resolution of $896\times 672$ ($31.28$ of DMESA+QT vs. $11.23$ of QT). Nevertheless, although we set the area image size as the training resolution of $640\times 480$, our methods bring limited improvement (MESA) or even decrease the performance (DMESA). This can be attributed to that the excessive area size adjustment hinders the matching redundancy reduction achieved by our methods, as the training resolution is not square and we have to excessively expand some areas to fit the aspect ratio.


\section{Benefits of A2PM}
\begin{figure}[!t]
\centering
\includegraphics[width=\linewidth]{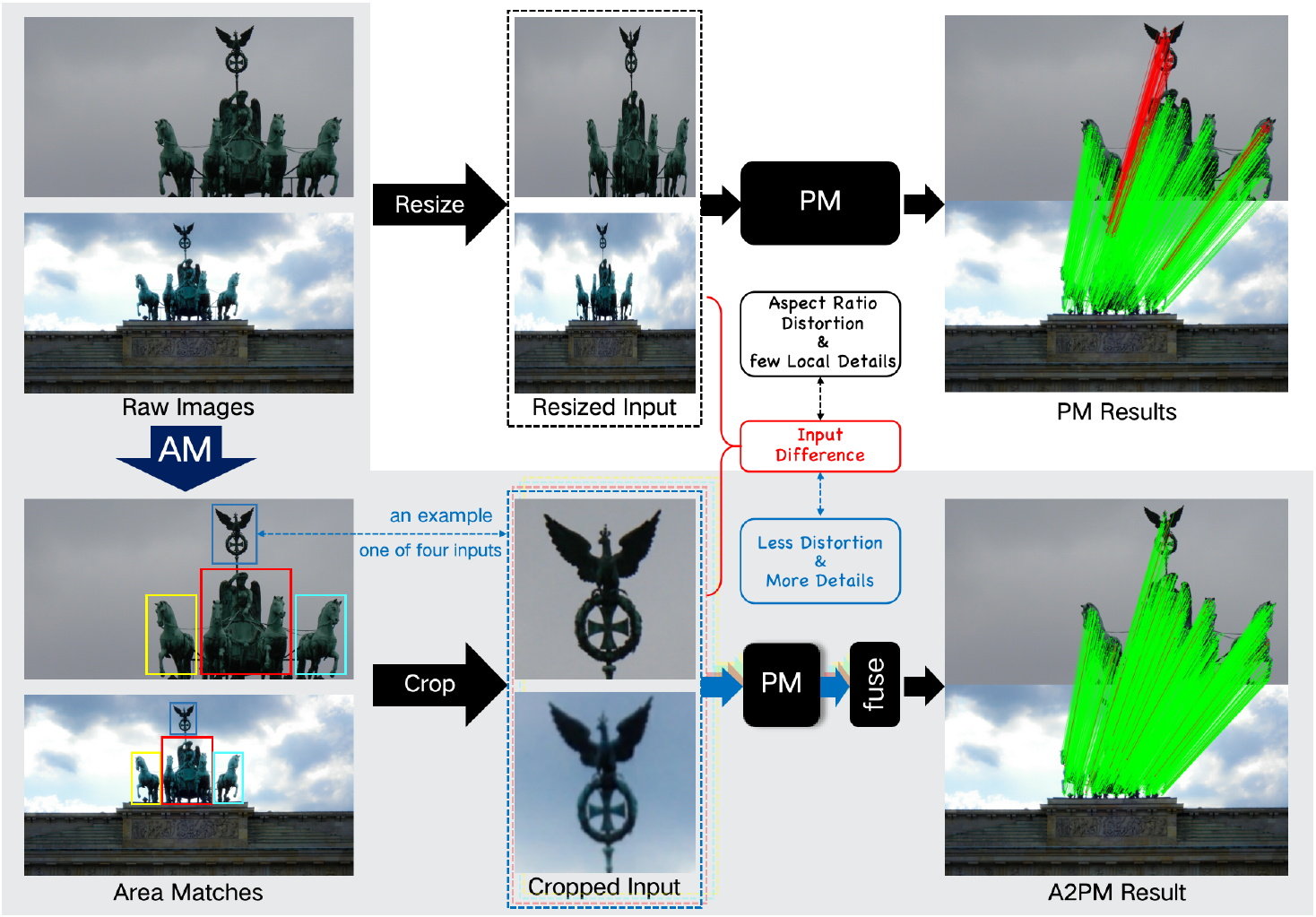}
\caption{\textbf{The benefit of the {\colorbox{Gray!30}{A2PM} framework}.} Essentially, the A2PM framework changes the input of PM. Based on the accurate AM, the matching redundancy is reduced, thus facilitating the inside-area PM with sufficient local details. Also, the cropping operation can avoid distortion from aspect ratio modification, which can be severe in resize operation (top).}
\label{fig:b-a2pm}
\end{figure}

In this section, we provide more detailed description about the benefit of the A2PM framework. See \cref{fig:b-a2pm}. The A2PM framework leverage the AM to split the original matching task into multiple easier inside-area matching tasks. Due to the reduced matching redundancy, area pair images contain substantial local details benefiting PM, which can be omitted by the original PM during resize operation. Another benefit comes from the cropping operation of A2PM, which can get the PM input with required resolution while maintaining the aspect ratio inherent to the raw image. Conversely, the resize operation widely applied in PM can result in severe distortion due to the aspect ratio variation (see the ``resized input'' in the \cref{fig:b-a2pm}). However, the premise of the above advantages is accurate area matching, which is the pursuit of the proposed MESA and DMESA.

\begin{table}[!t]
\centering
\caption{\textbf{Experiments of resolution overfitting in Transformer-based methods.} The experiments are conducted on ScanNet1500, measuring pose estimation accuracy. We select the training resolution $640\times480$ for PM, along with another resolution $896\times672$ which maintains the aspect ratio but increases the resolution.}\label{tab:ro}
\resizebox{\linewidth}{!}{
\begin{threeparttable}
\begin{tabular}{@{}cllllll@{}}
\toprule
\multirow{2}{*}{Pose estimation AUC} & \multicolumn{3}{c}{\textbf{$640\times480~(4/3)$}\tnote{$\dagger$}} & \multicolumn{3}{c}{\textbf{$896\times672~(4/3)$}\tnote{$\dagger$}} \\ \cmidrule(l){2-4} \cmidrule(l){5-7}
 & AUC@5$\uparrow$ & AUC@10$\uparrow$ & AUC@20$\uparrow$ & AUC@5$\uparrow$ & AUC@10$\uparrow$ & AUC@20$\uparrow$ \\ \cmidrule(l){1-1} \cmidrule(l){2-4} \cmidrule(l){5-7}
LoFTR~\cite{loftr} & 25.68 & 45.86 & 62.60 & 15.48 & 30.60 & 45.29 \\
MESA+LoFTR & 26.23$\goodd _{+2.14\%}$ & 46.06$\goodd_{+0.44\%}$ & 62.90$\goodd_{+0.48\%}$ & 18.17$\goodd_{+17.37\%}$ & 34.02$\goodd_{+11.18\%}$ & 49.13$\goodd_{+8.48\%}$ \\
DMESA+LoFTR & 24.37$\bad_{-5.10\%}$ & 44.42$\bad_{-3.14\%}$ & 61.34$\bad_{-2.10\%}$ & 17.58$\goodd_{+13.57\%}$ & 33.38$\goodd_{+9.08\%}$ & 48.35$\goodd_{+6.76\%}$ \\ \midrule
QT~\cite{quadtree} & 28.56 & 49.30 & 65.78 & 2.88 & 6.27 & 11.23 \\
MESA+QT & 28.74$\goodd_{+0.63\%}$ & 49.12$\bad_{-0.37\%}$ & 66.03$\goodd_{+0.38\%}$ & 5.53$\goodd_{+92.01\%}$ & 11.43$\goodd_{+82.30\% }$& 18.62$\goodd_{+65.81\%}$ \\
DMESA+QT & 26.51$\bad_{-7.18\%}$ & 46.71$\bad_{-5.25\% }$& 63.41$\bad_{-3.60\%}$ & 9.55$\goodd_{+231.60\%}$ & 20.20$\goodd_{+221.17\%}$ & 31.28$\goodd_{+178.54\%}$ \\ \bottomrule
\end{tabular}
\begin{tablenotes}
     \item[$\dagger$] {\small \textit{Input Resolution (Aspect Ratio)}.}
\end{tablenotes}
\end{threeparttable}
}
\end{table}

\begin{table*}[!t]
\caption{\textbf{Cross-Domain Evaluation of Pose Estimation.} We apply the learning models (including point matching models in baselines and area matching models in MESA and DMESA) trained on the outdoor scene (MegaDepth) to estimate camera poses in the indoor scene (ScanNet1500).
Relative gains are highlighted as subscripts. The \colorbox{colorFst}{\bf best}, \colorbox{colorSnd}{second} and \colorbox{colorTrd}{third} results are highlighted.}\label{tab:2CD}
\resizebox{\linewidth}{!}{
\begin{tabular}{llllllllllll}
\toprule
 \multicolumn{3}{c}{\multirow{2}{*}{Pose estimation AUC}} & \multicolumn{3}{c}{$640 \times 640$} & \multicolumn{3}{c}{$640 \times 480$} & \multicolumn{3}{c}{$480 \times 480$}  \\ \cmidrule(l){4-6} \cmidrule(l){7-9} \cmidrule(l){10-12} 
\multicolumn{3}{c}{}  & AUC@5$\uparrow$       & AUC@10$\uparrow$      & AUC@20$\uparrow$      & AUC@5$\uparrow$       & AUC@10$\uparrow$      & AUC@20$\uparrow$  & AUC@5$\uparrow$       & AUC@10$\uparrow$      & AUC@20$\uparrow$       \\ \midrule
\multirow{3}{*}{\rotatebox{90}{Sparse}} & \multicolumn{2}{l}{SP~\cite{superpoint}+SG~\cite{superglue}}                    & \rd \rd20.46 & \nd38.27 & \nd 54.92 & \rd 20.08 & \rd38.03 & \rd55.02 & \rd18.84 & \rd36.61 & \rd53.49   \\\dashlineours{2}{12}
 &\multicolumn{2}{l}{\ours MESA+SP+SG}               & \fs 22.34{\good $_{+9.19\%}$} & \fs 39.95{\good $_{+4.39\%}$} & \fs 56.88{\good $_{+3.57\%}$} & \fs 22.43{\good $_{+11.70\%}$} & \fs 40.12{\good $_{+5.50\%}$} & \fs 57.04{\good $_{+3.67\%}$} & \fs 22.47{\good $_{+19.27\%}$} & \fs 41.23{\good $_{+12.62\%}$} & \fs 57.89{\good $_{+8.23\%}$}  \\
 &  \multicolumn{2}{l}{\ours DMESA+SP+SG}              & \nd 20.67{\good $_{+1.03\%}$} & \nd 38.27{\zero $_{+0.00\%}$} & \rd 54.56{\bad $_{-0.66\%}$} & \nd 20.60{\good $_{+2.59\%}$} & \nd 38.37{\good $_{+0.89\%}$} & \nd 55.28{\good $_{+0.47\%}$} & \nd 20.79{\good $_{+10.35\%}$} & \nd 38.63{\good $_{+5.52\%}$} & \nd 55.25{\good $_{+3.29\%}$}   \\ \cmidrule(l){1-12} 
\multirow{9}{*}{\rotatebox{90}{Semi-Dense}} &\multicolumn{2}{l}{ASpan~\cite{aspanformer}}                    & \rd 21.99       & \rd 40.21      & \rd 56.87        & \nd 24.11       & \nd 43.61       & \rd 60.22  & \rd 22.82 & \rd 41.78 & \rd 58.32   \\\dashlineours{2}{12}
 &  \multicolumn{2}{l}{\ours MESA+ASpan}               & \fs 24.21{\good $_{+10.10\%}$} & \fs 43.77{\good $_{+8.85\%}$} & \fs 60.08{\good $_{+5.64\%}$} & \fs 25.31{\good $_{+4.98\%}$} & \fs 46.18{\good $_{+5.89\%}$} & \fs 62.04{\good $_{+3.02\%}$} & \fs 24.22{\good $_{+6.13\%}$} & \fs 44.16{\good $_{+5.70\%}$} & \fs 60.98{\good $_{+4.56\%}$}  \\
 &  \multicolumn{2}{l}{\ours DMESA+ASpan}              & \nd 22.91{\good $_{+4.18\%}$} & \nd 41.40{\good $_{+2.96\%}$} & \nd 57.44{\good $_{+1.00\%}$} & \rd 23.81{\bad $_{-1.24\%}$} & \rd 43.46{\bad $_{-0.34\%}$} & \nd 60.50{\good $_{+0.46\%}$} & \nd 23.96{\good $_{+5.00\%}$} & \nd 43.06{\good $_{+3.06\%}$} & \nd 59.43{\good $_{+1.90\%}$}   \\ \cmidrule(l){2-12} 
 & \multicolumn{2}{l}{QT~\cite{quadtree}}                       & \rd 22.40       & \rd 40.10       & \rd 56.90       & \rd 22.25       & \rd 41.51       & \rd 58.51  & \rd 21.77 & \rd 40.31 & \rd 57.02  \\\dashlineours{2}{12}
&  \multicolumn{2}{l}{\ours MESA+QT}                  & \fs 24.64{\good $_{+10.00\%}$} & \fs 43.91{\good $_{+9.50\%}$} & \fs 61.45{\good $_{+8.00\%}$} & \fs 24.32{\good $_{+9.30\%}$} & \fs 43.98{\good $_{+5.95\%}$} & \fs 60.99{\good $_{+4.24\%}$} & \fs 24.73{\good $_{+13.60\%}$} & \fs 44.15{\good $_{+9.53\%}$} & \fs 60.84{\good $_{+6.70\%}$}   \\
 &  \multicolumn{2}{l}{\ours DMESA+QT}                 & \nd 23.46{\good $_{+4.73\%}$} & \nd 41.98{\good $_{+4.69\%}$} & \nd 58.51{\good $_{+2.83\%}$} &\nd 23.06{\good $_{+3.64\%}$} & \nd 42.05{\good $_{+1.30\%}$} & \nd 59.08{\good $_{+0.97\%}$} & \nd 22.68{\good $_{+4.18\%}$} & \nd 41.82{\good $_{+3.75\%}$} & \nd 58.65{\good $_{+2.86\%}$}  \\ \cmidrule(l){2-12} 
 & \multicolumn{2}{l}{LoFTR~\cite{loftr}}                    & \rd 19.79       & \rd 36.91       & \rd 52.63       & \rd 20.94       & \rd 38.68       & \rd 54.61 & \rd 19.82 & \rd 37.59 & \rd 53.28   \\\dashlineours{2}{12}
 &  \multicolumn{2}{l}{\ours MESA+LoFTR}               & \fs 21.34{\good $_{+7.83\%}$} & \fs 39.23{\good $_{+6.29\%}$} & \fs 55.12{\good $_{+4.73\%}$} & \fs 22.31{\good $_{+6.54\%}$} & \fs 40.34{\good $_{+4.29\%}$} & \fs 57.12{\good $_{+4.60\%}$} & \fs 21.56{\good $_{+8.78\%}$} & \fs 40.07{\good $_{+6.60\%}$} & \fs 57.33{\good $_{+7.60\%}$}   \\
 &  \multicolumn{2}{l}{\ours DMESA+LoFTR}              & \nd 20.99{\good $_{+6.06\%}$} & \nd 38.51{\good $_{+4.33\%}$} & \nd 54.23{\good $_{+3.04\%}$} & \nd 21.11{\good $_{+0.81\%}$} & \nd 39.14{\good $_{+1.19\%}$} & \nd 55.29{\good $_{+1.25\%}$} & \nd 21.17{\good $_{+6.81\%}$} & \nd 39.49{\good $_{+5.05\%}$} & \nd 55.50{\good $_{+4.17\%}$}   \\ \cmidrule(l){1-12} 
\multirow{3}{*}{\rotatebox{90}{Dense}} & \multicolumn{2}{l}{DKM~\cite{dkm}}                    & \rd 25.67       & \rd46.01       & \nd 63.05       & \rd27.00       & \nd 47.42       & \rd64.59 & \rd25.75 &	\rd45.71 &	\rd62.96  \\\dashlineours{2}{12}
&  \multicolumn{2}{l}{\ours MESA+DKM}               & \fs 28.91{\good $_{+12.62\%}$} & \fs 48.77{\good $_{+6.00\%}$} & \fs 65.18{\good $_{+3.38\%}$} & \fs 28.89{\good $_{+7.00\%}$} & \fs 49.34{\good $_{+4.05\%}$} & \fs 66.32{\good $_{+2.68\%}$} & \fs 28.12{\good $_{+9.23\%}$} & \fs 48.31{\good $_{+5.00\%}$} & \fs 65.49{\good $_{+3.87\%}$}   \\
&  \multicolumn{2}{l}{\ours DMESA+DKM}              & \nd 25.92{\good $_{+0.97\%}$} & \nd 46.33{\good $_{+0.70\%}$} & \rd 62.84{\bad $_{-0.33\%}$} & \nd 27.14{\good $_{+0.52\%}$} & \rd 47.39{\bad $_{-0.06\%}$} & \nd 64.70{\good $_{+0.17\%}$} & \nd 26.11{\good $_{+1.40\%}$} & \nd 46.09{\good $_{+0.83\%}$} & \nd 63.33{\good $_{+0.59\%}$}    \\  \bottomrule
\end{tabular}
}

\end{table*}
\section{Study of Cross-Domain Generalization}\label{sec:ab-cd}
Given the broad range of application scenarios of feature matching, the ability to generalize across domains is crucial for matching methods. Therefore, in this section, we construct experiments to evaluate the cross-domain generalization of our methods.

\subsubsection{Experimental setup}
To establish the cross-domain matching task, we employ models trained in the outdoor dataset (MegaDepth), including \textbf{both} the point matching models and learning models in MESA (learning area similarity) and DMESA (patching matching), to perform feature matching on indoor images (ScanNet1500). The baseline selection, resolution range, and method parameter settings in this experiment are kept consistent with other experiments on the ScanNet1500 dataset (cf. \cref{sec:exp-pe}).

\subsubsection{Results}
We present the results in \cref{tab:2CD}. \textit{For the sparse matcher}, our methods result in an overall increase in accuracy, showcasing the prominent generalization. Both MESA and DMESA achieve the best results at the smallest resolution of $480\!\times\!480$, proving the resolution robustness of our methods. This also means our methods can attain better accuracy with less computational cost, which matters in applications with limited computation budget.

\textit{For the semi-dense point matchers}, the accuracy drop of our methods at the training resolution observed in in-domain experiments is eliminated, replaced by an overall performance improvement. This indicates that our approaches can significantly enhance the generalization of the semi-dense matchers. Moreover, our methods reduce the accuracy gap between different sizes, showcasing resolution robustness.

\textit{For the dense matcher}, our methods lead to a remarkable increase in accuracy. Particularly at the small size of $480\!\times\!480$, MESA+DKM reaches the precision level of in-domain performance (cross-domain MESA+DKM on AUC@5: $28.89$ vs.$29.76$ of in-domain DKM), demonstrating the enhancement of our method on the cross-domain generalization.

In the experiments, MESA showcases better generalization in contrast to DMESA, as the pre-trained coarse matcher in DMESA suffers from the domain gap. Nonetheless, leveraging the benefits of A2PM and the remarkable versatility of SAM, DMESA still contributes to improving the generalization of point matchers.

\section{Details of AG Completion}\label{sec:d-agc}
In this section, we provide additional details of completing the Area Graph (AG). The initial AG contains few directed edges, due to the splitting nature of SAM, which hinders robust and efficient matching. Thus, we propose to generate more graph nodes to form a tree structure for AG, \ie the graph completion algorithm.
The detailed process for the graph completion is depicted in \cref{alg:gc}, which takes initial AG ($\mathcal{G}_{ini}$) as input and outputs the final AG ($\mathcal{G}$) with scale hierarchy. Furthermore, we describe the area \texttt{clustering} and two main area operations adopted to \texttt{generate} higher level nodes in the algorithm as follows.

\begin{figure}[!t]
\centering
\includegraphics[width=\linewidth]{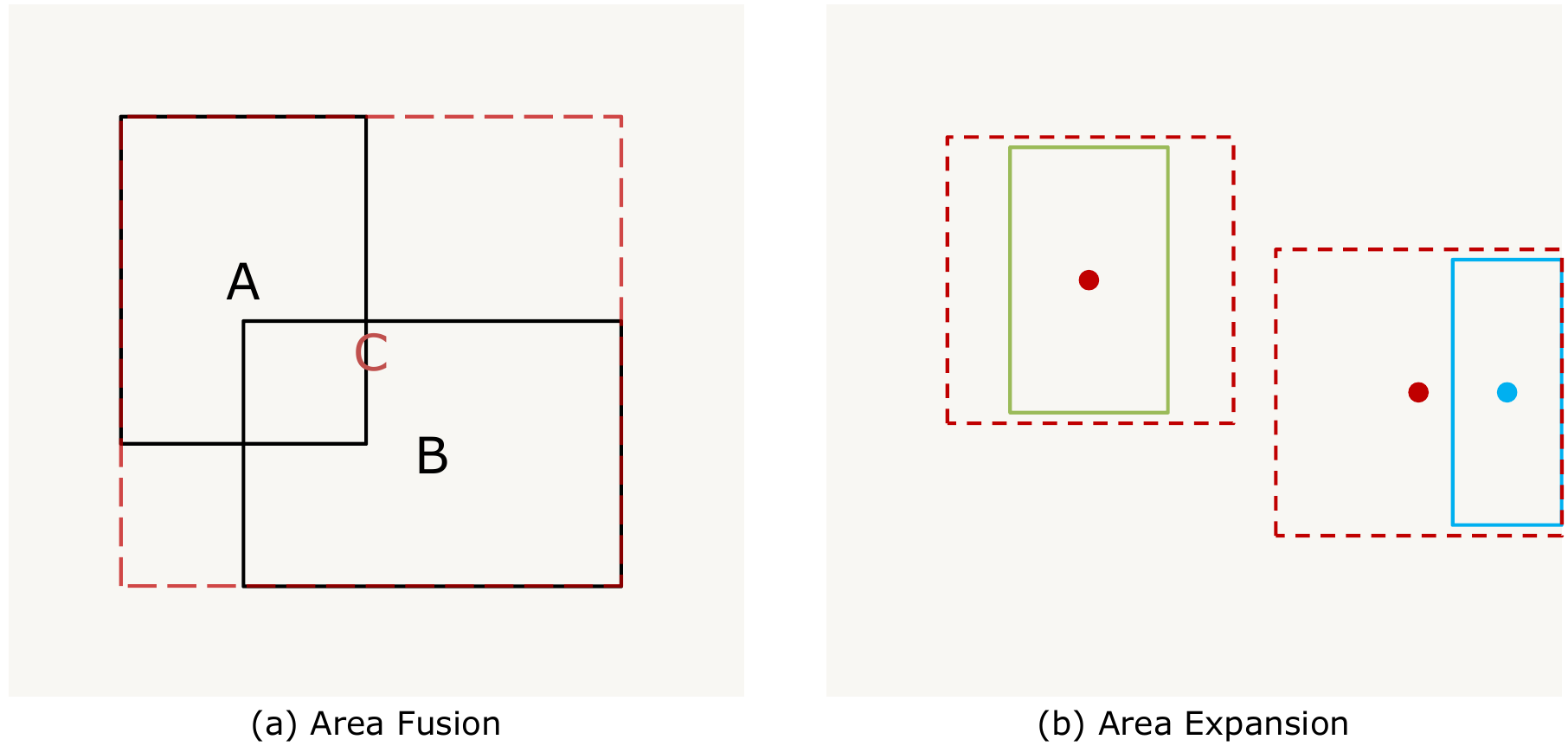}
\caption{\textbf{The Area Fusion and Area Expansion.} \textbf{(a)} Area fusion is to achieve the smallest area (\textit{C}) containing the input areas (\textit{A} and \textit{B}). \textbf{(b)} Generally, area expansion is to fix the \textcolor[rgb]{.61,.73,.35}{original area} center and expand its size to the smallest size of the next level. When the \textcolor[rgb]{.0,.69,.94}{original area} is too close to the image boundary, we will move the area center to keep the expanded area inside the image.}
\vspace{-1.2em}
\label{fig:ao}
\end{figure}

\begin{algorithm2e}[t]
    \normalem
    \caption{Graph Completion}\label{alg:gc}
    \KwIn{$\mathcal{G}_{ini}=\langle \mathcal{V}_{ini}, \mathcal{E}_{ini}\rangle$}
    \KwOut{$\mathcal{G} = \langle \mathcal{V}, \mathcal{E}\rangle$}

    \For{$l$ in $[0, L-1]$}
    {
        initial orphan node set $\mathcal{O} = \varnothing$\;
        \For{$v_i \in \{v_i|l_{a_i} = l\}$}
        {
        \uIf{$v_i$ has no parent}
        {
            add $v_i$ into $\mathcal{O}$\;
        }
        }
        \texttt{cluster} the nodes in $\mathcal{O}$ based on their area centers\;
        \For{each node cluster $\mathcal{C}_h=\{v_k\}_{k=0}^C$}
        {
            \eIf{$C \ge 2$}
            {
                \For{each $v_k \in \mathcal{C}_h$}
                {
                    \uIf{$v_k$ has not been fused}
                    {
                        fuse area $a_k$ with its nearest neighbor $a^n | v^n \in \mathcal{C}_h$:
                        $a^f = F({a}_k, a^n)$\;
                        \texttt{generate} higher level node $v^f$ for $a^f$\;
                        add $v^f$ into $\mathcal{V}_{ini}$\;
                        form edges by Link Prediction: $\{e_h\}_h = LP(v^f, \mathcal{V}_{ini})$\;
                        add $\{e_h\}_h$ into $\mathcal{E}_{ini}$\;
                    }
                }
            }
            {
            Update the single node $v_0$: $v^u_0 = Up(v_0)$\;
            construct edges: $\{e_j\}_j = LP(v^u_0, \mathcal{V}_{ini})$\;
            add $\{e_j\}_j$ into $\mathcal{E}_{ini}$\;
            }
        }
    }
    $\mathcal{E} = \mathcal{E}_{ini}$\ \;
    $\mathcal{V} = \mathcal{V}_{ini}$\ \;
    output the updated AG: $\mathcal{G} = \langle \mathcal{V}, \mathcal{E}\rangle$\;
\end{algorithm2e}

\subsection{Area Clustering.}
For orphan nodes in each level, we cluster them based on their area centers to decide which operation will be performed on them.
We use the k-means algorithm with elbow method~\cite{kmeans} to determine the cluster number. The candidate cluster number is set as $\{1,\dots,n\}$, where $n$ is the number of orphan nodes in the current level. This algorithm is fed with area centers and outputs labeled ones.

\subsection{Area Fusion and Expansion.}
Area fusion and expansion are key operations in our graph completion algorithm.
Specifically, area fusion is to find the largest outer rectangle of the two areas as the new area, as depicted in \cref{fig:ao} (a). Due to the careful threshold settings of our area level, the fused area size will exceed current level and be awaited for subsequent operations.
On the other hand, the expansion operation is to expand the area to the next level size (\cref{fig:ao} (b)). In particular, suppose the lower bound of size for the next level is $s^2$, if both of the area width and height are smaller than $s$, we expand the height and width of the area to $s$, keeping the area center fixed. Otherwise if area width $w \geq s$, we let the area height $h = s^2/w$, keeping the area center fixed, and vice versa.
The above operations are performed when the expanded area is inside the image. On the other hand, if the expanded area is outside the image, the area center will be moved as shown in \cref{fig:ao} (b).

\section{Computation Complexity Analyze of MESA}\label{sec:cca}
Here, we analyze the computation complexity of proposed graphical area matching, demonstrating the main source of the efficiency issue in MESA.

\subsection{Area Similarity Calculation}
Firstly, area similarity calculation is performed to achieve the required node energies in the graph, serving as the prerequisite of our graphical area matching.
Suppose we have two AGs, $\mathcal{G}^0 $ and $ \mathcal{G}^1$, for the input image pair, $\mathcal{G}^0$ gets $N$ nodes ($|\mathcal{V}^0|=N$) and $\mathcal{G}^1$ gets $M$ nodes ($|\mathcal{V}^1|=M$). Therefore, the dense graph energy calculation needs $M \times N$ times similarity calculation. However, owing to the similarity conditional independence of ABN (Sec.~\ref{sec:abn}), the actual number ($M' \times N'$) of similarity calculation is smaller than $M \times N$, as $N' < N$. Nevertheless, directly setting children pair similarities as $0$ is too rough (\cref{eq:abn}), as large scale differences also leads to near-zero similarity between areas. In practise, we only set the related similarities of \textit{next level children} as $0$ for area matching accuracy and the efficiency from ABN is still helpful to our approach.  Moreover, we only care about the similarities between source nodes in $\mathcal{G}^0$ and other nodes in $\mathcal{G}^1$, because we collect source nodes with specific level from $\mathcal{G}^0$ to match, \eg, usually $3\sim 4$ areas in indoor scene and less in outdoor scene. Therefore, we have $M' < M$. Similarly, in the case of duality, \ie, collecting source nodes from $\mathcal{G}^1$ to match, we only need to perform a few supplementary calculations, as similarities are symmetric and reusable.
Thus, the real computation complexity of area similarity computation is $O(M' \times N')$, where $M' \times N' < M \times N$.

\subsection{Edge Energy Calculation}
Except the node energy calculation, the edge energy is also needed to be determined for \textit{Graph Cut}. The computation complexity of edge energy calculation is related to edge number of $\mathcal{G}^0$ and $\mathcal{G}^1$. Assume $|\mathcal{E}_0|=E$ and $|\mathcal{E}_1|=K$, the specific computation complexity is $O(E+K)$.

\subsection{Global Energy Minimization}
In our global energy minimization for area matching refinement, the matching energy of parent, children and neighbour pairs all need to be calculated. Taking parent matching energy for example, we derive its computation complexity as follows. Suppose $n$ nodes are achieved as match candidates through \textit{Graph Cut} and each node gets $Q_i,\;i\in(0,n]$ parent nodes, there are $Q_i \times V$ node similarities need to be accessed (as the similarity calculation is finished), where $V$ is the parent node number of the source node. Hence, the total computation complexity for parent matching energy in global energy minimization is $O(\sum_i^n Q_i \times V)$. The children matching energy and neighbour matching energy are similar.
As $n$ is the number of node after \textit{Graph Cut}, it is small in most cases, \eg, usually $<3$ area nodes. Moreover, the number of parent nodes (or children, neighbour nodes) is also limited. Therefore, the computation complexity for global energy minimization is acceptable in practise.

In sum, the efficiency issue of MESA mainly lies in the Area Similarity Calculation part, which contains quadratic computational complexity. This issue comes from the sparse area matching framework in MESA, thus motivating us to design the dense counterpart, DMESA.

\begin{table}[!t]
\caption{\textbf{Ablation study of resolution settings.} Two different image cropping methods are compared for the proposed MESA. Both semi-dense and dense point matchers are combined for evaluation. We report the pose estimation AUC$@5^{\circ}/10^{\circ}/20^{\circ}$ and the \textbf{best} results of two series are highlighted respectively.}\label{tab:ABAC}
\resizebox{\linewidth}{10mm}{
\begin{tabular}{ccccc}
\toprule
Method                       & Cropping Approach     & AUC@5~$\uparrow$ & AUC@10~$\uparrow$ & AUC@20~$\uparrow$ \\ \midrule
\multirow{2}{*}{MESA+ASpan} & \textit{C$\rightarrow$R}             &  24.67 & 43.72  & 61.29  \\
                             & \textit{E$\rightarrow$C} & \bf 27.51 & \bf 47.47  & \bf 65.04  \\ \hdashline \noalign{\vskip 1pt}
\multirow{2}{*}{MESA+DKM}   & \textit{C$\rightarrow$R}            & 30.19 & 51.49  & 68.79  \\
                             & \textit{E$\rightarrow$C} & \bf 33.42 & \bf 55.04  & \bf 71.98 
                             \\ \bottomrule
\end{tabular}
}

\end{table}

\section{Additional Ablation Study}\label{sec:aas}

\subsection{Ablation Study on the Resolution Setting}\label{sec:ab-ic}
The resolution setting is non-trivial and important in the A2PM framework, as different settings lead to different image quality and distortions. Here, we construct experiments to investigate the impact of different resolution settings. In practice, different resolution settings correspond to different area image cropping operations.  Thus, we compare two different cropping methods: \textbf{1)} the straightforward cropping method (\textit{C$\rightarrow$R}), which \textbf{crops} areas with original aspect ratios and then \textbf{resizes} them to input resolution. It means using arbitrary area image resolution.
\textbf{2)} the \textit{E$\rightarrow$C} cropping method, which first \textbf{expands} the area to correspond with the aspect ratio of the point matcher input and then \textbf{crops} these ares. This corresponds to our setting described in \cref{sec:a2pm-c}.

The experiment is conducted on ScanNet1500~\cite{scannet} benchmark. We combine MESA with both semi-dense (ASpan) and dense (DKM) point matchers for complete comparison. Results are summarized in Tab.~\ref{tab:ABAC}. As we can seen that the \textit{E$\rightarrow$C} cropping approach outperforms the \textit{C$\rightarrow$R} approach with a large margin for both MESA+ASpan and MESA+DKM, proving its superiority due to high resolution and less distortion. Therefore, we adopt the \textit{E$\rightarrow$C} approach for area image cropping, which means the area image resolution shares the same aspect ratio with the input PM resolution.

\begin{table}[!t]
\centering
\caption{\textbf{Ablation study of global energy parameters.} We compare different parameter settings for global energy refinement in MESA\_ASpan and report the area matching performance, area number per image (AreaNum), and the pose estimation performance. Results are highlighted as \colorbox{colorFst}{\bf first}, \colorbox{colorSnd}{second} and \colorbox{colorTrd}{third}.}\label{tab:ABEP}
\resizebox{\linewidth}{!}{
\begin{tabular}{cccccc}
\toprule
$E_{G}$ Parameters & $T_{E_{max}}$ & AOR~$\uparrow$   & AMP@$0.6$~$\uparrow$ & Pose AUC@$5^\circ$~$\uparrow$ & AreaNum~$\uparrow$ \\ \midrule
\multirow{3}{*}{\makecell[c]{$\mu=5,\;\alpha=2,$\\$\beta=2,\;\gamma=1$}}
& 0.35          & 61.76 &  65.54     &  \rd 23.57              & \nd 4.69    \\              
& 0.25          & 63.91 &  71.13     & 22.41              & 3.47    \\
& 0.15          & 60.44 & 62.57     & 21.46              & 3.27    \\ \hdashline \noalign{\vskip 1pt}
\multirow{3}{*}{\makecell[c]{$\mu=4,\;\alpha=2,$\\$\beta=2,\;\gamma=2$}}                     
& 0.35          & \fs 67.98 & \fs 80.09     & \nd 23.74              & \fs 5.76    \\
& 0.25          & \rd 64.94 & \rd 72.24     & \fs 24.01              & \rd 4.62    \\
& 0.15          & 61.74 & 65.50     & 23.55              & 3.86    \\ \hdashline \noalign{\vskip 1pt}
\multirow{3}{*}{\makecell[c]{$\mu=7,\;\alpha=1,$\\$\beta=1,\;\gamma=1$}}                     
& 0.35          & \nd 65.98 & \nd 78.10     & 22.71             & 3.27    \\
& 0.25          & 62.32 & 66.54     & 23.56              & 2.92    \\ 
& 0.15          & 60.32 & 64.38     & 22.37                 & 2.77      \\ \bottomrule
\end{tabular}
}
\vspace{-1.2em}
\end{table}

\subsection{Ablation Study on Global Energy Parameters}\label{sec:ab-ep}
The parameters for our global energy refinement in MESA mainly consists of global energy balance parameters ($E_G$ Parameters) in \cref{eq:eg} and the threshold parameter $T_{E_{max}}$.
The four $E_G$ Parameters reflect the importance of four energy terms, \ie, self matching energy, parent, children and neighbour matching energy.
The $T_{E_{max}}$ controls the maximum energy of the final match, the smaller it is the stricter the refinement.
Here, we construct experiments on ScanNet1500 to investigate the performance impact of these parameters. In particular, we compare three groups of $E_G$ Parameters and three groups of $T_{E_{max}}$ to evaluate their impact on MESA\_ASpan. The input size of ASpan is $480\times 480$. The area matching performance, pose estimation performance and area number per image are summarised in \cref{tab:ABEP}.
Generally, if two areas are matched, their parent, children and neighbour nodes should have high similarities due to spatial relationships between them.
At the same time, the self matching energy should still be an important reference in matching refinement.
Thus we choose three parameter settings including different weights on three kinds of node matching energies and different emphasis on self-matching energy.
The experiment results in Tab.~\ref{tab:ABEP} show that the weights of three parameter settings set to the same is better for area matching performance ($\alpha\!=\!\beta\!=\!\gamma~vs.~\alpha\!=\!\beta\!\not=\!\gamma$).
Giving sufficient consideration on global matching leads to accurate area matching along with best point matching performance ($\mu\!=\!4~vs.~\mu\!=\!7$).
Despite the semi-dense matcher is not sensitive to area matching accuracy, better area matching leads to higher pose estimation precision.
Therefore, we choose $[\mathsf{4,2,2,2}]$ as our energy setting.
On the other, the $T_{E_{max}}$ is a critical parameter as well. 
The smaller $T_{E_{max}}$ means stricter global matching energy request, but it may also mistake some accurate area matches when too small.
Different $E_G$ Parameter settings prefer different values of $T_{E_{max}}$ and $0.35$ suits the best for ours.

\begin{figure}[!t]
\centering
\includegraphics[width=\linewidth]{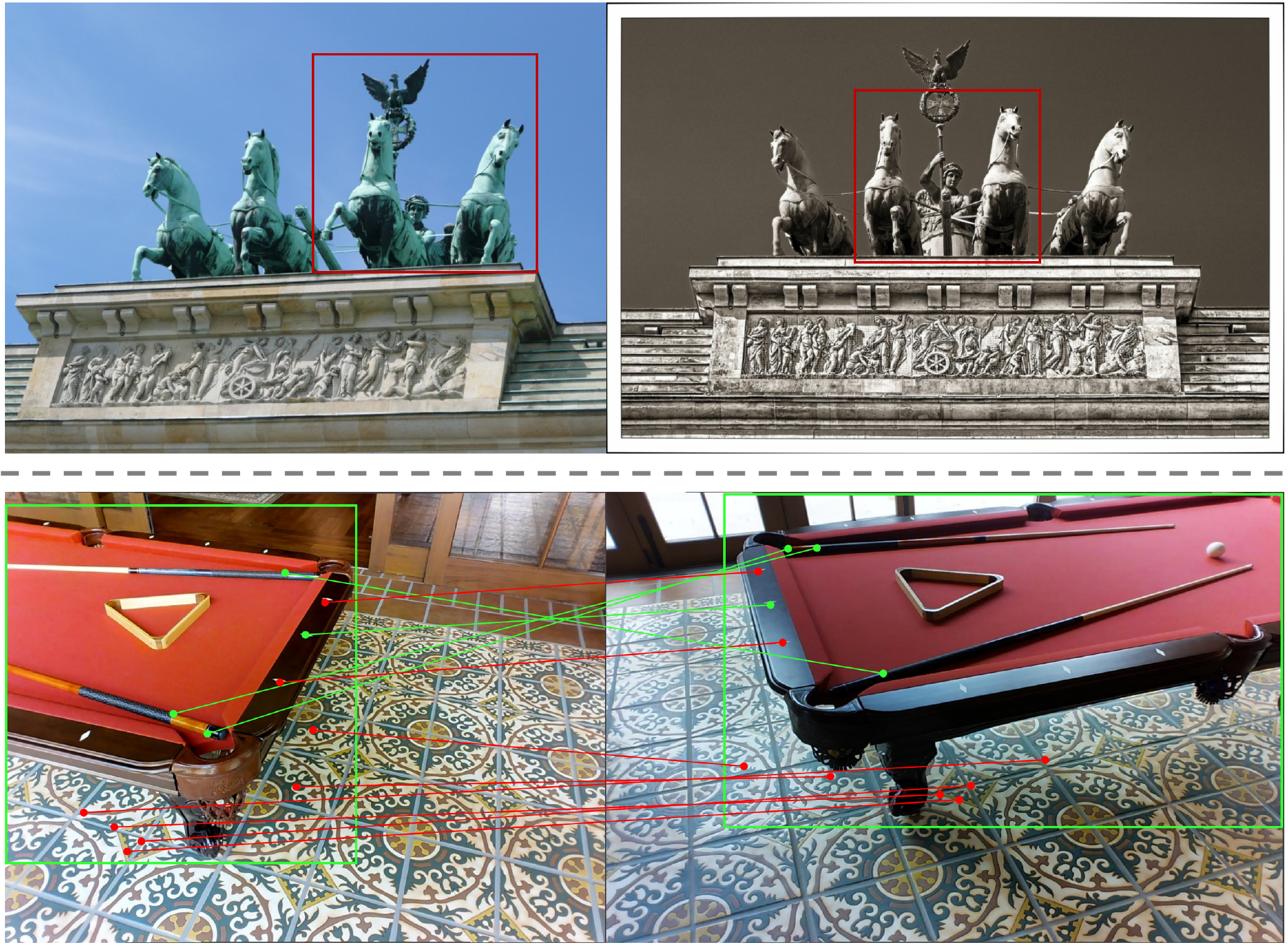}
\caption{\textbf{The Failure case of MESA and DMESA.} \textbf{Top}: MESA may produce false area matches when repeated objects and large viewpoint variance occur at the same time. The impact of this kind of erroneous match can be alleviated by post-processing like GAM~\cite{sgam}. \textbf{Bottom}: Area matching (DMESA+SPSG) cannot completely resolve the repetitiveness issues in matching. While detailed feature comparisons within area matches can differentiate some repetitive points, challenges arising from highly repetitive textures and symmetrical objects persist unresolved.}
\vspace{-1.2em}
\label{fig:fc}
\end{figure}

\section{Limitation and Future Work}\label{sec:lfw}
One common limitation of MESA and DMESA is the under-utilisation of SAM features. As we mentioned before, SAM possesses the high-level image understanding across a wide range of domains due to the massive training dataset and carefully designed models. Therefore, its image embedding is an extremely strong high-level representation, which has the potential to replace our learning similarity model. Then, the computation cost can be reduced as well. However, the naive attempt to use SAM features as descriptors of areas failed, possibly because the SAM segmentation pays more attention on intra-image contexts rather than inter-image ones like feature matching. Hence, the SAM feature needs further distillation for area matching, which will be an objective of our future work. 

On the other hand, as MESA fuses image areas based on their 2D distances, which may not be lifted equivalently to 3D. Thus, some inconsistent area fusions between two images arise and lead to inaccurate point matching, \eg, shown in \cref{fig:fc} top. Although the post-processing like GAM~\cite{sgam} may help, it also introduces extra computation cost. To address this issue, feature-guided fusion can be adopted, where the SAM feature can be employed and lead to consistent area fusion. 

Moreover, our methods cannot perfectly solve the repetitiveness issues. Some repetitive patterns can be discerned by unique objects within the area matches. However, when the central object, such as the symmetrical pool table in \cref{fig:fc} bottom, does not facilitate the identification of true matches from repetitive patterns, our methods are unable to assist point matchers in managing receptiveness issues.

Finally, there is an optimization space related to efficiency for the A2PM framework. Although the area matching speed of DMESA aligns with the current SOTA, the overall matching process of the A2PM framework is still time-intensive. This can be attributed to that the original \textit{single} matching task is divided into multiple matching tasks by A2PM. This issue could be addressed by parallel computation and GPU acceleration. On the other hand, considering the significant precision improvement achieved by our methods, they are valuable for some downstream tasks that are not sensitive to time cost, such as SfM.



\end{document}